\documentclass{article}

\PassOptionsToPackage{numbers, compress}{natbib}

\usepackage[main, final]{neurips_2026}

\usepackage{subcaption}
\usepackage{xspace}
\usepackage{xparse}
\usepackage[table]{xcolor}
\usepackage[most]{tcolorbox}
\usepackage{rotating}     

\usepackage[utf8]{inputenc} 
\usepackage[T1]{fontenc}    
\usepackage{hyperref}       
\usepackage{url}            
\usepackage{booktabs}       
\usepackage{amsfonts}       
\usepackage{nicefrac}       
\usepackage{microtype}      
\usepackage{tabu}
\usepackage{soul}
\usepackage{bbm}
\usepackage{lipsum}
\usepackage{kantlipsum}
\usepackage{tabularx}
\usepackage{thmtools}

\usepackage{cancel}

\usepackage{amsmath,amssymb,amsfonts,amsthm, dsfont, color}
\usepackage{algorithm}
\usepackage{algorithmic}
\usepackage{mathtools}
\usepackage{graphicx}
\usepackage{textcomp}
\usepackage{xcolor, fancyhdr}
\usepackage{enumitem}
\usepackage{float}
\usepackage{nicefrac}

\usepackage{wrapfig}
\usepackage{mathtools}

\definecolor{MyGreen1}{RGB}{20,180,40}
\usepackage{multirow, tikz,float}
\allowdisplaybreaks

\usepackage{nicefrac,color,mathrsfs,float}
\usepackage{multirow,caption,tikz}
\usetikzlibrary{shapes.misc, bending, positioning, calc, fit}
\usetikzlibrary{decorations.pathreplacing}
\usetikzlibrary{arrows.meta, shapes,patterns.meta}
\usetikzlibrary{decorations.text}
\tikzset{gray arrow/.style={-{Triangle[length=1mm]}, gray}}

\tikzset{
  block/.style    = {draw, thick, rectangle, minimum width = 2em},
sblock/.style      = {draw, thick, rectangle, minimum height = 2em,
minimum width = 2em}, 
}

\usepackage{mathabx}

\newcommand{\calX}{\mathcal{X}}

\usepackage{comment}

\newcommand{\binary}{\{0,1\}}

\newcommand{\set}[1]{[#1]}

\DeclarePairedDelimiterX{\infdivx}[2]{(}{)}{%
  #1\;\delimsize\|\;#2%
}

\usepackage{prettyref,xspace}
\usepackage{tikz}

\newrefformat{cond}{Condition~\ref{#1}}
\newrefformat{eq}{Eq.~\eqref{#1}}
\newrefformat{thm}{Thm.~\ref{#1}}
\newrefformat{th}{Theorem~\ref{#1}}
\newrefformat{chap}{Chapter~\ref{#1}}
\newrefformat{sec}{Sec.~\ref{#1}}
\newrefformat{algo}{Algorithm~\ref{#1}}
\newrefformat{fig}{Fig.~\ref{#1}}
\newrefformat{tab}{Table~\ref{#1}}
\newrefformat{rmk}{Remark~\ref{#1}}
\newrefformat{clm}{Claim~\ref{#1}}
\newrefformat{def}{Def.~\ref{#1}}
\newrefformat{cor}{Corollary~\ref{#1}}
\newrefformat{lmm}{Lemma~\ref{#1}}
\newrefformat{prop}{Proposition~\ref{#1}}
\newrefformat{pr}{Proposition~\ref{#1}}
\newrefformat{app}{App.~\ref{#1}}
\newrefformat{prob}{Problem~\ref{#1}}
\newrefformat{ques}{Question~\ref{#1}}
\newrefformat{notee}{Note~\ref{#1}}
\newrefformat{assump}{Assumption~\ref{#1}}
\newrefformat{issuee}{Issue ~\ref{#1}}
\newrefformat{fix}{Fix ~\ref{#1}}

\newcommand{\vect}[1]{\boldsymbol{#1}}

\newcommand{\ba}{\vect{a}}

\newcommand{\bd}{\vect{d}}

\newcommand{\bh}{\vect{h}}

\newcommand{\br}{\vect{r}}
\newcommand{\bs}{\vect{s}}

\newcommand{\bx}{\vect{x}}

\newcommand{\btheta}{\vect{\theta}}

\usepackage{derivative}

\newcommand{\define}{\triangleq}

\newcommand{\pth}[1]{\left( {#1} \right)}
\newcommand{\qth}[1]{\left[ {#1} \right]}
\newcommand{\sth}[1]{\left\{ {#1} \right\}}
\newcommand{\eg}{e.g.\xspace}
\newcommand{\ie}{i.e.\xspace}

\mathchardef\mhyphen="2D

\definecolor{MyGreen1}{RGB}{20,180,40}

\usepackage{siunitx}

\let\ast\star 

\usepackage{minitoc,tocloft}
\usepackage{etoc}

\usepackage[capitalize,noabbrev]{cleveref}
\crefname{section}{Sec.}{Secs.}
\Crefname{section}{Sec.}{Secs.}
\crefname{subsection}{Sec.}{Secs.}
\Crefname{subsection}{Sec.}{Secs.}
\crefname{subsubsection}{Sec.}{Secs.}
\Crefname{subsubsection}{Sec.}{Secs.}
\crefname{figure}{Fig.}{Figs.}
\Crefname{figure}{Fig.}{Figs.}
\crefname{equation}{Eq.}{Eqs.}
\Crefname{equation}{Eq.}{Eqs.}
\crefname{appendix}{App.}{Apps.}
\Crefname{appendix}{Appendix}{Appendices}

\newcommand{\method}{\textsc{Terminator}\xspace}      
\newcommand{\rtx}[1][]{CoT#1\xspace}    

\NewDocumentCommand{\logprob}{o}{%
  \IfNoValueTF{#1}{log-probability\xspace}{log-probabilities\xspace}%
}
\NewDocumentCommand{\tokconf}{o}{%
  \IfNoValueTF{#1}{Token-Confidence\xspace}{Token-Confidences\xspace}%
}
\newcommand{\bfx}{\ensuremath{\boldsymbol{x}}\xspace}

\newcommand{\bfz}{\ensuremath{\boldsymbol{z}}\xspace}
\newcommand{\bfr}{\ensuremath{\boldsymbol{r}}\xspace}            
\newcommand{\bff}{\ensuremath{\boldsymbol{s}}\xspace}            
\newcommand{\bfs}{\ensuremath{\boldsymbol{d}}\xspace}            
\newcommand{\bfa}{\ensuremath{\boldsymbol{a}}\xspace}            
\newcommand{\ans}{\ensuremath{\hat{\boldsymbol{a}}}\xspace}      

\newcommand{\tok}{\ensuremath{i^\ast}\xspace}            %

\newcommand{\lrm}{\texttt{LRM}\xspace}
\newcommand{\ourdata}{\ensuremath{\mathsf{HORL}}-dataset\xspace}

\newlength{\commentwidth}
\newdimen\remainingheight
\newcommand*{\calcremainingheight}{%
    \ifdim\pagegoal=\maxdimen
        \remainingheight\dimexpr\textheight-0.4pt\relax
    \else
        \remainingheight\dimexpr\pagegoal-\pagetotal-\lineskip-0.4pt\relax
    \fi
}


\newtcolorbox[auto counter]{mainbox}[2][]{%
  colframe=green!10!white, 
  colback=green!10!white,  
  before upper={{\textbf{Main Contributions.} #2}},
  #1
}

\title{\method: Learning Optimal Exit Points for Early Stopping in Chain-of-Thought Reasoning}

\author{%
  Alliot Nagle\thanks{Corresponding Author: \texttt{acnagle@utexas.edu}} \\
  UT Austin \\
  \And
  Jakhongir Saydaliev \\
  EPFL \\
  \And
  Dhia Garbaya \\
  ENS Paris-Saclay \\
  \And
  Michael Gastpar \\
  EPFL \\
  \And
  Ashok Vardhan Makkuva\thanks{Co-last authors.} \\
  Télécom Paris (IP Paris) \\
  \And
  Hyeji Kim\footnotemark[2] \\
  UT Austin
}

\begin{document}

\maketitle

\begin{abstract}

Large Reasoning Models (LRMs) achieve impressive performance on complex reasoning tasks via Chain-of-Thought (CoT) reasoning, which enables them to generate intermediate thinking tokens before arriving at the final answer. However, LRMs often suffer from significant \emph{overthinking}, spending excessive compute time even after the answer is generated early on.
Prior work has identified the existence of an optimal reasoning length such that truncating reasoning at this point significantly shortens CoT outputs with virtually no change in performance.
However, determining optimal CoT lengths for practical datasets is highly non-trivial as they are fully task and model-dependent. In this paper, we precisely address this and design \method, an early-exit strategy for LRMs at inference to mitigate overthinking. The central idea underpinning \method is that the first arrival of an LRM's final answer is often predictable, and we leverage these first answer positions to create a novel dataset of optimal reasoning lengths to train \method. Powered by this approach, \method achieves significant reductions in CoT lengths of $14\% $–$55\%$ on average across four challenging practical datasets: MATH-500, AIME 2025, HumanEval, and GPQA, while outperforming current state-of-the-art methods and reducing inference latency by more than 2$\times$ compared to the original LRM. \looseness=-1

Project page: \url{https://terminator-llm.github.io}

\end{abstract}
\section{Introduction}
\label{sec:intro}

The advent of Large Reasoning Models (LRMs) has proven itself to be a critical next step for Large Language Models (LLMs) to surpass human-level performance. LRMs use test-time compute to ``think'' through a problem before answering, an approach that has led to significant performance gains across many challenging tasks \cite{openai2024learning}. However, this improvement does not come for free, as an LRM will generate thousands of additional thinking tokens to solve a single problem, compared to its non-reasoning counterparts \cite{guo2025deepseek}. Worse yet, LRMs spend a significant amount of their reasoning tokens double-checking their work and exploring different solutions when they have already generated the final answer, that they will eventually settle on, much earlier in the \rtx, a phenomenon known as \textit{overthinking} \cite{luo2025o1pruner,chen2025donotthink}. Prior work has shown that the length of a \rtx can be reduced by $50\%$ or more on average with little drop in accuracy \cite{kang2025c3ot,zhang2025adaptthink,yang2025dynamic}, demonstrating the extent to which compute is wasted during LRM inference. \looseness=-1

Given that reasoning can be wasteful, a natural question to ask is, \textit{for any given accuracy, does there exist an optimal reasoning length?} 
Previous works have shown that LRM performance, as a function of reasoning length, gradually increases, peaks, and then decreases, suggesting the existence of an optimal reasoning length \cite{wu2025when,lee2025critical}.
Additionally, some recent works propose novel RL-training algorithms to fine-tune LRMs to produce shorter \rtx[s] \cite{luo2025o1pruner,lou2025adacot,gao2025faroptimal,yi2025shorterbetter,shrivastava2025sample} and establish the Pareto frontier for those methods, showing that gaps still exist between them \cite{gao2025faroptimal}. 
While these works focus on retraining an LRM, inference-time methods such as DEER \cite{yang2025dynamic} enable early termination of reasoning without retraining it. However, for practical tasks none of these methods either determines or utilizes the optimal-length reasoning, which in fact provides the best possible reduction in \rtx length. \looseness=-1

In this paper, we precisely address this by introducing the novel notion of \emph{hindsight-optimal reasoning length} (\cref{sec:hindsight}): given a reasoning task, in hindsight, {what is the fewest number of tokens that an LRM needs to generate before providing the same answer it would have provided without shortened reasoning?}
Namely, we mark the first logical arrival, as opposed to any other occurrence, of the LRM's final answer as the hindsight-optimal exiting position. Leveraging this notion, we design a novel inference-time early-exit algorithm \method that significantly outperforms current state-of-the-art methods in reductions to \rtx lengths on challenging practical datasets (\cref{fig:our-method}). In particular, \method capitalizes on the fact that the first arrival of the final answer is (1) marked by a distinctive shift in the LRM's token-level confidence and token usage distribution, and (2) can be used as a signal to train a binary probe classifier for effective early-exiting during reasoning. \looseness=-1

\begin{figure*}[t]
    \centering
    \includegraphics[width=1.0\textwidth]{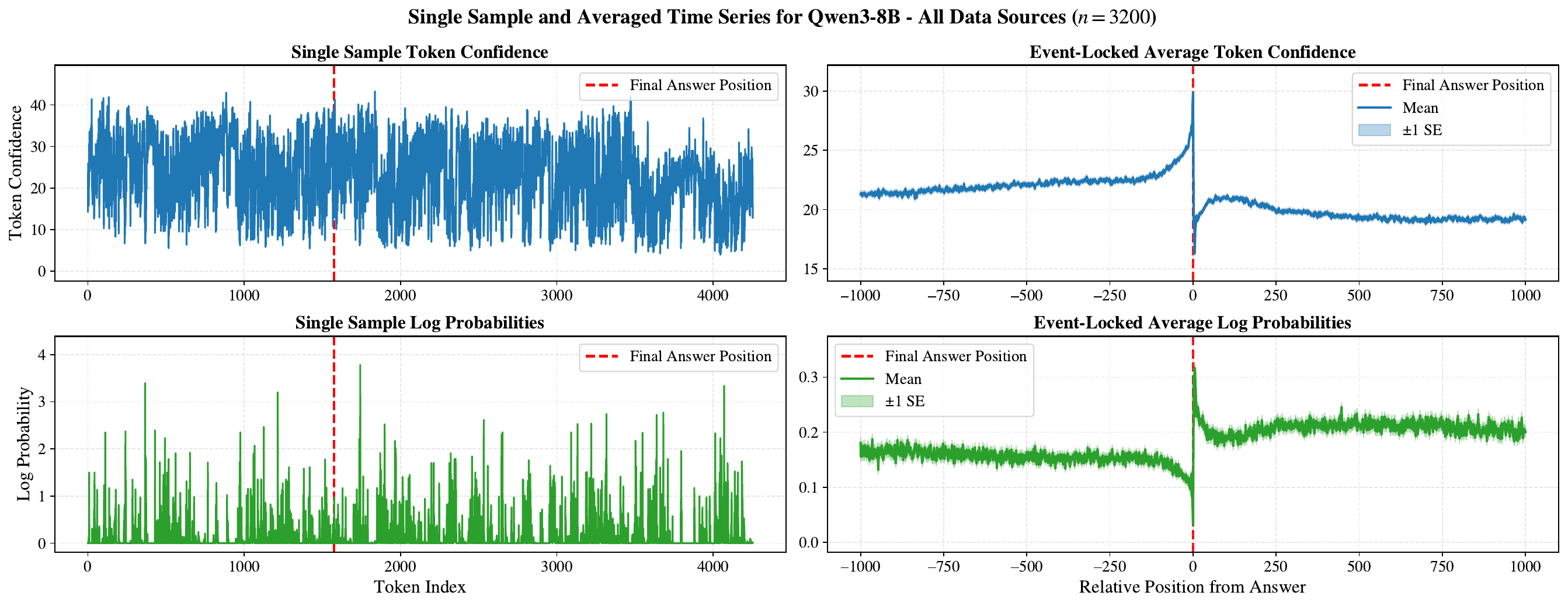}
    \caption{\textbf{Event-Locked Averaging of \tokconf.} Event-locked averaging shows a consistent agreement on spiking behavior at the answer position in each \rtx, but disagrees elsewhere. On the other hand, this phenomenon is not readily observable in the single-sample case. Figures on the \textbf{left} show the \tokconf \cite{fu2025deepthink} and \logprob trajectories throughout reasoning for a single, randomly selected sample; figures on the \textbf{right} show the effect of \textit{event-locked averaging} on the position of the first arrival of the final answer across all \rtx[s]. The 3200 \rtx[s] used are a random subset of our training set, which combines AIME (1983--2024), MATH, OpenCoder-SFT, and OpenScience. \cref{fig:aligned-timeseries-aime,fig:aligned-timeseries-math,fig:aligned-timeseries-opencoder,fig:aligned-timeseries-openscience} in \cref{app:additional_experiments} show similar trends for each dataset separately. The standard error (SE) is shown as a shaded region and becomes clearer when zoomed in.}
    \label{fig:aligned-timeseries-all}
\end{figure*}

\begin{tcolorbox}[colback=green!5!white,colframe=green!75!black]

\textbf{Main Contributions.} In summary, we make the following contributions:

\begin{itemize}[leftmargin=1.25em]

    \item We introduce the novel notion of hindsight-optimal reasoning, using which we show that the first arrival of an LRM's final answer is marked by observable and meaningful signals (\cref{fig:aligned-timeseries-all}). To the best of our knowledge, this is the first such analysis of its kind.
    
    \item We design \method, a novel, lightweight, early-exit algorithm for LRMs that is trained on optimal-length \rtx[s] (\cref{sec:binary-classifer}), resulting in more than a $2\times$ reduction in inference latency compared to the original LRM (\cref{sec:main-results}).
    
    \item We introduce a robust pipeline for identifying the first arrival of the final answer in \rtx[s], using which we construct a novel optimal-length \rtx training dataset (\cref{sec:extraction}). 
    
\end{itemize}

\end{tcolorbox}

\begin{figure*}[t]
    \centering
    \includegraphics[width=1.0\textwidth]{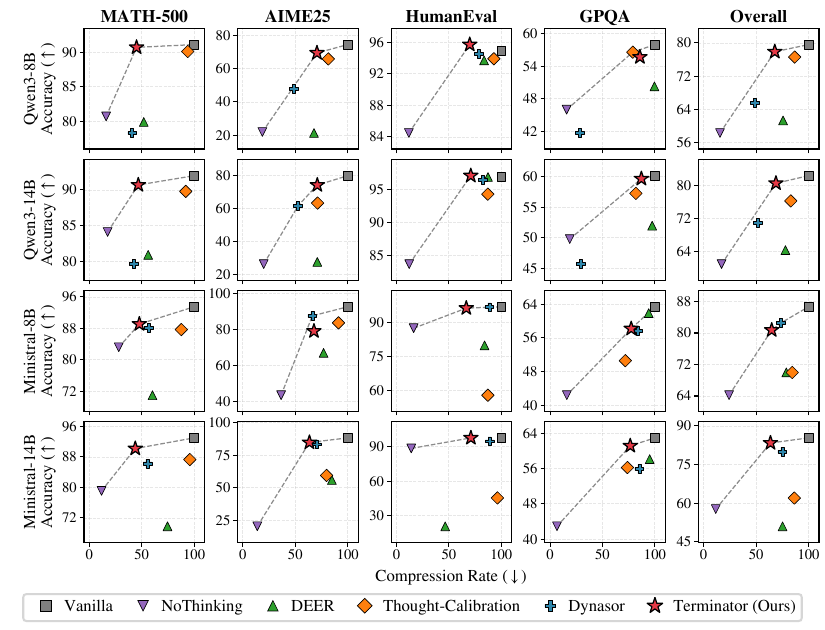}
    \caption{
    \textbf{Pareto Frontier.} \method defines the Pareto frontier on 14 of the 16 (LRM, benchmark) pairings, outperforming prior work. Each point represents a method's accuracy and compression rate, with lower compression rates indicating greater token savings and hence compute. The dashed line traces the Pareto frontier connecting non-dominated solutions. We refer to \cref{sec:main-results} and \cref{tab:performance} in \cref{app:additional_experiments} for more details.}  
    \label{fig:pareto-frontier}
\end{figure*}

\section{Preliminaries}
\label{sec:prelim}

\subsection{Notation}
\label{sec:notation}

A Large Reasoning Model \lrm takes as input the prompt sequence $\bx = \pth{x_1, x_2, x_3, \ldots, x_L} $ and produces two outputs $\br$ and $\bs$ auto-regressively (\cref{fig:our-method}). Here $ \br = \pth{ r_1, r_2, r_3, \ldots, r_M}$ is the \rtx sequence generated during the thinking stage, \ie $r_i = \lrm(\bx, \br_{<i})$ for $i \in \set{M} \define \{1, \ldots, M\}$, and $ \bs = \pth{s_1, s_2, s_3, \ldots, s_N}$ is the solution that summarizes this \rtx and contains a final answer \ans, which could be a single numerical answer, a math expression, code, a multiple-choice option, etc. Here $s_j = \lrm(\bx, \br, \bs_{<j})$ for $j \in \set{N}$. Note that the final answer \ans is separate from the ground-truth answer $\ba$; they may or may not be in agreement with each other. Throughout the paper, \ans always refers to the final answer of the full \rtx, not the final answer generated after exiting a \rtx early. Furthermore, when referring to \ans with respect to its position in a \rtx, we always mean the \textit{earliest logical arrival} of \ans unless stated otherwise explicitly. By the earliest logical arrival of \ans, we are referring to the sequence of logical steps in the \rtx that yields the final answer \ans for the first time. For any early-exit strategy, a key metric to gauge its performance is the per-sample compression rate (CR): $\frac{M_{\text{early}}}{M}$, where $M_{\text{early}} \in \set{M}$ is the token index of early exit in $\br$. 
Accuracy (Acc) measures the proportion of problems where the correct answer is produced.

\subsection{Hindsight-optimality}
\label{sec:hindsight}
We now formally define our novel notion of hindsight-optimality. Given an input prompt $\bx \in \calX^L$ of length $L$ over a vocabulary $\calX$, an LRM \lrm generates a corresponding \rtx $\br \in \calX^M$ and solution $\bs \in \calX^N$, where the solution contains a final answer \ans$\in \calX$. The hindsight-optimal reasoning length ($\mathsf{HORL}$) is defined as the earliest position in the completed \rtx at which the final answer \ans has been logically reached. Mathematically,
\begin{align}
 \mathsf{HORL}(\bx, \br, \bs, \ans)
    &\define \min\sth{i \in [M]: \br_{\leq i} \text{ contains the earliest logical arrival of } \ans}.
    \label{eq:hindsight}
\end{align}
Here, $\br_{\leq i}$ is said to contain the earliest logical arrival of \ans if, by position $i$, the sequence of reasoning steps in $\br$ has produced the first derivation of the final answer \ans. Thus, $\mathsf{HORL}$ is a retrospective property of the realized \rtx $\br$ and final answer \ans. 

\subsection{Token-Confidence}
\label{sec:token-confidence}

Our analytical experiments require a measure of LRM's confidence during the generation of a \rtx. To this end, we use the \tokconf metric, which gauges the uncertainty of a chosen token. Mathematically, for every $ i \in \set{M}$, the corresponding \tokconf $C_i$ is defined as
\begin{equation}
\label{eqn:token-confidence}
C_i \define -\frac{1}{K} \sum_{k \in \mathcal{T}_K(i)} \log \mathbb{P}_{\lrm}\pth{r_i = k \mid \bfx, \bfr_{< i} },
\end{equation}
where $\mathbb{P}_{\lrm}\pth{r_i = \cdot \mid \bfx, \bfr_{< i} }$ is the \lrm prediction probability at position $i$ and $\mathcal{T}_K(i) \define \text{Top-}K \qth{\mathbb{P}_{\lrm}\pth{r_i = \cdot \mid \bfx, \bfr_{< i}  } } $ is the set of vocabulary tokens corresponding to the Top-$K$ probabilities. In other words, \tokconf is the average (negative) \logprob across the Top-$K$ probabilities (we set $K=20$ in our experiments). The higher it is, the more confident the model is in its predictions. \looseness=-1

This measure is based on the Self-Certainty metric \cite{kang2025scalable}, computed as the KL-divergence between the uniform distribution and the token distribution, and is based on the following idea: the higher the confidence of the model, the further its predictions should be from the uniform distribution. We also note that, while Token-level \logprob[s] are commonly used as a proxy for confidence, we prefer the \tokconf measure \cite{fu2025deepthink} here, as it is principled and produces less noise. Both are used, and the same conclusions can be drawn using either. \looseness=-1

\section{Motivation}
\label{sec:motivation}

Shortly after the breakthrough of LRMs, it was observed that they exhibit an overthinking phenomenon where, despite arriving at the correct answer, they continue to consider alternative solution paths, possibly leading to other incorrect answers \cite{chen2025donotthink,luo2025o1pruner}. While LRMs achieve greatly improved performance over their non-reasoning counterparts, they do so at a much higher inference-time cost: up to thousands of additional tokens are generated to form the \rtx before arriving at the final solution \cite{guo2025deepseek}. Many follow-up works have observed the same overthinking phenomenon and developed methods to mitigate wasteful token expenditure \cite{zhang2025adaptthink,liu2025answer,wu2025thought,zhang2025reasoningmodels}. \looseness=-1

Towards designing an optimal early-exit strategy to stymie overthinking, we build upon the following key observation: once an LRM generates a \rtx \bfr and a final solution \bff, we can observe the final answer \ans. Then, \textit{in hindsight}, we can determine precisely where the LRM should have exited the \rtx to avoid wasting tokens, \ie $\mathsf{HORL}$, and instead generate the final solution. To this end, we first only need to check for \ans, not \bfa, in \bfr, since the LRM may never even have generated the correct (ground-truth) answer and thus may not exist in \bfr. Second, by choosing to terminate reasoning after the arrival of \ans in \bfr, all steps that are useful to arriving at \ans are kept, and anything after is skipped as it is usually redundant. \looseness=-1

While the above procedure requires the explicit knowledge of \ans to check for its arrival, \emph{are there meaningful markers to implicitly detect its arrival?}

\looseness=-1

\begin{figure*}[t]
    \centering
    \includegraphics[width=1.0\textwidth]{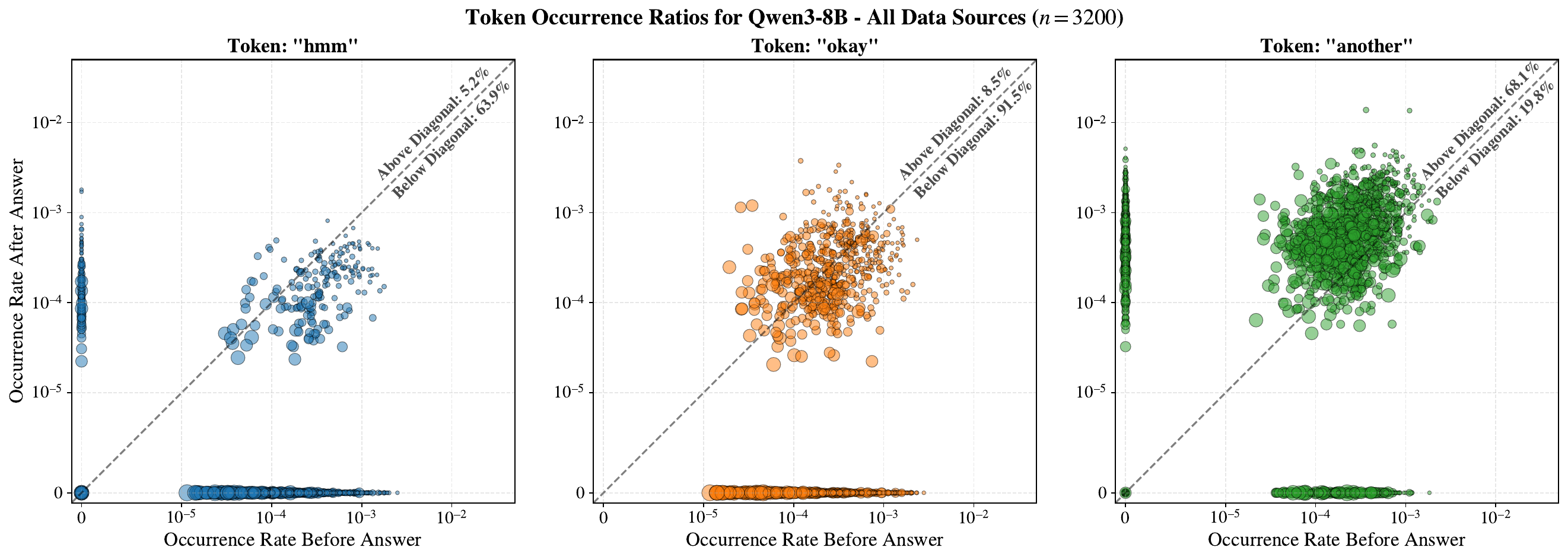}
    \caption{\textbf{Token Usage Frequency Shift.} ``Thinking token'' usage changes depending on whether the final answer has been generated in the \rtx. Rates are computed by counting the raw number of occurrences of the token before and after the answer, and then normalizing each count by the respective number of tokens in the before and after bins. The arrival of the final answer is hinted at by changes in the rates for these tokens. The relative length of a \rtx is captured by its dot size, where a longer \rtx has a larger dot. \cref{app:additional_experiments} demonstrates similar results for other ``thinking tokens'' in \cref{fig:token-scatter-all} and for each data source in \cref{fig:token-scatter-aime,fig:token-scatter-math,fig:token-scatter-opencoder,fig:token-scatter-openscience}.}
    \vspace{-8pt}
    \label{fig:token-scatter}
\end{figure*}

\textbf{Detecting the Answer Early. } We analyze trends in Token-Confidence during CoT reasoning on AIME (1983--2024), MATH, OpenCoder-SFT, and OpenScience, as shown in \cref{fig:aligned-timeseries-all,fig:token-scatter}. \cref{fig:aligned-timeseries-all} reveals a sharp transition in \tokconf around the first occurrence of \ans in event-locked averages, formed by aligning each \rtx['s] first \ans position to $0$ and averaging across samples. Because these \rtx[s] span math, science, and coding datasets, this transition suggests a consistent cross-domain signal; per-dataset versions appear in \cref{fig:aligned-timeseries-aime,fig:aligned-timeseries-math,fig:aligned-timeseries-opencoder,fig:aligned-timeseries-openscience}.

%

\cref{fig:token-scatter} complements this by comparing token frequencies before and after the first \ans occurrence for three ``thinking tokens'': \texttt{hmm}, \texttt{okay}, and \texttt{another}. These tokens are associated with ongoing reasoning, e.g., \texttt{wait}, \texttt{so}, \texttt{alternatively}, \texttt{therefore}, and we hypothesize that their usage changes once \ans has been generated \cite{wang2025wait,qian2025demystifying,ding2025thinkingtokens}. Indeed, \texttt{hmm} and \texttt{okay} occur more often before \ans, while \texttt{another} occurs more often after it. Although not every thinking token shows a clear before/after bias, these shifts indicate that token-frequency distributions can signal when \ans has appeared; additional examples are shown in \cref{fig:token-scatter-all}. In each scatter plot, the axes show before- and after-\ans token rates, computed as raw counts divided by the corresponding number of tokens, and point diameter indicates relative \rtx length.

\textbf{Moving to Online Inference and Challenges. } While these results strongly indicate the early arrival of \ans, using them during online inference remains a challenge. In the case of \cref{fig:aligned-timeseries-all}, event-locked averaging requires multiple \rtx[s] to be generated simultaneously, each with reasonable estimates of the position of the answer. Under those circumstances, the spiking behavior will emerge. But attaining a reasonable estimate of the answer position for a single \rtx during inference is the original problem we are tasked with. While applying the event-locked averaging signal to online inference is limited, it does indicate that an underlying trend can be extracted. \looseness=-1

Similarly, each dot in \cref{fig:token-scatter} requires full knowledge of each \bfr and its position of \ans so that the rates can be calculated accordingly. Again, these results show a shift in the usage frequency of certain tokens before and after the first occurrence of \ans, but translating the signal into an online inference algorithm remains challenging. \looseness=-1

\begin{figure}[t]
  \centering
  \resizebox{\linewidth}{!}{\definecolor{attn}{HTML}{FF7F0E}    
\definecolor{ff}{HTML}{2CA02C}      
\definecolor{embed}{HTML}{1F77B4}   
\definecolor{stop}{HTML}{D62728}    
\definecolor{go}{HTML}{1F77B4}      

\begin{tikzpicture}[>={Latex[length=1mm,width=1mm]}, node distance=1cm]
        
    \node (input) {What is $2+3$?};     
    \node[right=0.1cm of input]    (start_think) {\texttt{<think>}};
    \node[right=0.1cm of start_think] (lets)    {Let's};
    \node[right=0.1cm of lets, yshift=-0.8pt] (start)    {start};
    \node[right=0.1cm of start] (dots)    {$\dots$};
    \node[right=0.1cm of dots, yshift=-1.4pt] (you)    {you};
    \node[right=0.1cm of you] (get)    {get};
    \node[right=0.05cm of get] (rect_anchor)    {};
    \node[right=0.1cm of get, yshift=1pt] (five)    {$5$};
    \node[right=0.1cm of five]    (end_think) {\texttt{</think>}};
    \node[right=0.1cm of end_think] (in)    {In};
    \node[right=0.1cm of in] (dots_again)    {$\dots$};
    \node[right=0.1cm of dots_again] (is)    {is};
    \node[right=0.1cm of is] (five_again)    {$5$};

    \draw [decorate,decoration={brace,amplitude=6pt,mirror,raise=1ex}] 
    (input.south west) -- (input.south east) node[midway, yshift=-2em, draw=none, rectangle, rounded corners, fill=attn!30] (input_x) {Input $\bx$};

    \draw [decorate,decoration={brace,amplitude=6pt,mirror,raise=1ex}] 
    (lets.south west) -- (five.south east) node[midway,yshift=-2em, draw=none, rectangle, rounded corners, fill=attn!30] (cot_r) {CoT $\br$};

    \draw [decorate,decoration={brace,amplitude=6pt,mirror,raise=1ex}] 
    (in.south west) -- (five_again.south east) node[midway,yshift=-2em, draw=none, rectangle, rounded corners, fill=attn!30]{Solution $\bs$};
    
    \node[above=0.5cm of get, minimum width = 10.9cm, minimum height = 0.8cm] (lrm) [draw, rectangle, rounded corners, fill=attn!30] {Large Reasoning Model \texttt{LRM}};

   \draw[-] (input.north) -- (lrm.west -| input.north);
   \draw[->] (lrm.west -| input.north) -- (lrm.west);
    \foreach \i in {start_think, lets, start, you, get, five, end_think, in, is, five_again}{
        \draw[->] (\i.north) --  (lrm.south -| \i.north);
    }

    \node[above=1.8cm of start_think]    (cot_lets) {Let's};
    \node[above=1.8cm of lets] (cot_start)    {start};
    \node[above=1.84cm of start] (cot_with)    {with};
    \node[right=0.12cm of cot_with] (cot_dots)    {$\dots$};
    \node[above=1.82cm of you] (cot_get)    {get};
    \node[above=1.85cm of get] (cot_five)    {$5$};
    \node[above=1.8cm of five] (above_five)    {};
    \node[above=1.85cm of end_think] (cot_in)    {In};
    \node[above=1.8cm of in] (cot_summary)    {summary};
    \node[above=1.85cm of is] (cot_five_again)    {$5$};

    \path (cot_summary) -- (cot_five_again) node[midway] (cotdots_midpoint) {$\dots$};

    \foreach \i in {cot_lets, cot_start, cot_with, cot_get, cot_five,  cot_in, cot_summary, cot_five_again}{
        \draw[->] (lrm.north -| \i.south) -- (\i.south);
    }

    \def\dis{0.6cm}
    \node[above=\dis of $(cot_lets)!0.3!(cot_start)$]    (h1) {};
    \node[above=\dis of $(cot_start)!0.37!(cot_with)$]    (h2) {};
    \node[above=\dis of $(cot_with)!0.42!(cot_dots)$]    (h3) {};
    \node[above=\dis of $(cot_get)!0.35!(cot_five)$]    (hi_minus_one) {};
    \node[above=\dis of $(cot_five)!0.30!(above_five)$]    (hi) {};

    \node[above=0.6cm of h3, minimum width = 5.9cm, minimum height = 1cm] (optexit) [draw, rectangle, rounded corners, fill=ff!40] {\method};

    \draw[->, dotted, thick]
        (cot_lets.north) -- (cot_lets.north |- optexit.south)
        node[pos=0.45, fill=white, inner sep=1pt] (bh1) {$\bh_1$};

    \draw[->, dotted, thick]
        (cot_start.north) -- (cot_start.north |- optexit.south)
        node[pos=0.45, fill=white, inner sep=1pt] (bh2) {$\bh_2$};

    \draw[->, dotted, thick]
        (cot_with.north) -- (cot_with.north |- optexit.south)
        node[pos=0.45, fill=white, inner sep=1pt] (bh3) {$\bh_3$};

    \draw[->, dotted, thick]
        (cot_get.north) -- (cot_get.north |- optexit.south)
        node[pos=0.45, fill=white, inner sep=1pt] (bhi_minus_one) {$\bh_{i-1}$};

    \draw[->, dotted, thick]
        (cot_five.north) -- (cot_five.north |- optexit.south)
        node[pos=0.45, fill=white, inner sep=1pt] (bhi) {$\bh_i$};

    \path (bh3) -- (bhi_minus_one) node[midway] (bhi_midpoint) {$\dots$};

   \def\dis{2cm}
    \node[above=\dis of h1]    (binary1) {$\textcolor{stop}{\mathbf{0}}$};
    \node[above=\dis of h2]    (binary2) {$\textcolor{stop}{\mathbf 0}$};
    \node[above=\dis of h3]    (binary3) {$\textcolor{go}{\mathbf 1}$};
    \node[above=\dis of hi_minus_one]    (binaryi_minus_one) {$\textcolor{stop}{\mathbf 0}$};
    \node[above=\dis of hi]    (binaryi) {$\textcolor{go}{\mathbf 1}$};

    \path (binary3) -- (binaryi_minus_one) node[midway] (binary_midpoint) {$\dots$};

    \foreach \i in {1,2,3,i_minus_one,i}{
    \draw[->] (optexit.north -| binary\i.south) -- (binary\i.south);
    }

    \node[draw, fit=(binary3) (binary_midpoint) (binaryi_minus_one) (binaryi)] (binary_box) {};

    \node[right=1.7cm of binary_box] (majority) [draw, rectangle, fill=embed!20] {$\text{Sum}(b_{i-9 : \, i}) > 5$};

    \draw[->] (binary_box.east) -- (majority.west) node[midway,above] {\small last $10$ bits} node[midway,below] {$b_{i-9 : \, i}$};

    \node[right=1.3cm of five_again] (curve_anchor2) {};
    \node[above=3cm of curve_anchor2] (curve_anchor1) {};

    \draw[->] (majority.east) to [out=0,in=90] (curve_anchor1.north) to node[pos=0.3, below,sloped] {\textcolor{go}{Inject \texttt{</think>}}} node[pos=0.3, above,sloped] (early_exit) {Early exit}  (curve_anchor2.north) to [out=-90, in=-60] (end_think.south);

\end{tikzpicture}}
  \caption{\textbf{Early stopping via \method.} \method is a binary probe classifier that predicts whether to exit or not at every CoT token. Once the majority of prediction bits within a window ($10$ here) are $1$, \texttt{</think>} is injected into the LRM's token stream to stop thinking.}
  \label{fig:our-method}
\end{figure}

\textbf{Our Approach. } Under the hood of the LRM, there are clearly meaningful signals to indicate the earliest arrival of \ans. However, as outlined above there are unique challenges in leveraging them in a hand-guided way to design an early-exit online inference algorithm. To address this, we approach this through the lens of \emph{prediction}: \ie predicting whether an LRM's final answer \ans has been generated or not. To this end, the core idea behind our method is to train a probe classifier---the \method---on the hidden states of the final layer, thereby utilizing as many of the LRM's underlying signals as possible (\cref{fig:our-method}). Prior work has examined hidden states to assess whether LRMs know when their intermediate \rtx answers are correct \cite{zhang2025reasoningmodels}, a finding primarily aimed at understanding model internals rather than designing practical early-exit methods at inference. Our work adopts a fundamentally different, deployable approach by probing for the \textit{final answer} \ans, a signal that is fully self-contained within the LRM's reasoning process and requires no ground-truth labels at inference time, thereby enabling a principled early-exit strategy. \looseness=-1

{\bf Advantages of \method over Prior Methods.} To train the probe classifier, we process inputs at the token level, offering much finer-grained predictions than prior work. That is, our dataset is curated for a token classification task. To the best of our knowledge, all previous methods use much coarser granularity in their training dataset where they chunk each CoT \bfr according to some heuristic, such as ``thinking tokens'' or paragraph delimiters such as \texttt{\textbackslash n\textbackslash n}. Then, at inference time, they exit once the predicted probability crosses a data-calibrated threshold \cite{liu2025answer,wu2025thought,zhang2025reasoningmodels}. 
In contrast, our approach offers two-fold advantages at inference-time: (1) a probe classifier trained with our dataset has the ability to exit immediately after \ans is generated, and (2) while our approach is amenable to using a data-calibrated threshold, it is not necessary. The main drawback of data-calibrated thresholding is that it requires additional samples from the evaluation data distribution, and the resulting threshold is therefore specific to that distribution and may not transfer well to other distributions. \looseness=-1

However, obtaining the \ans positions to create our \ourdata in a scalable way is challenging and highly non-trivial, which we precisely address in the next section. \looseness=-1
\section{\method: Methodology}
\label{sec:methods}

Given a full \rtx and the corresponding final solution from an LRM, the earliest logical arrival to the LRM's final answer can be detected in the \rtx. However, reliable detection for tens of thousands of \rtx[s] is a unique challenge, which we address through our pipeline in \cref{sec:extraction}. We then present our method for training \method, which is a probe classifier, in \cref{sec:binary-classifer}. \looseness=-1

\subsection{Early Answer Extraction, Identification, and Verification}
\label{sec:extraction}

\begin{figure}[t]
    \centering
    \scalebox{0.6}{\definecolor{attn}{HTML}{FF7F0E}    
\definecolor{ff}{HTML}{2CA02C}      
\definecolor{embed}{HTML}{1F77B4}   
\definecolor{stop}{HTML}{D62728}    
\definecolor{go}{HTML}{1F77B4}      

\begin{tikzpicture}[
    >={Latex[length=1mm,width=1mm]},
    node distance=8mm,
    box/.style={
        rectangle,
        fill=go!20,
        rounded corners,
        draw,
        thick,
        minimum width=4.3cm,
        minimum height=2.6cm,
        align=center,
        inner sep=7pt
    }
]

\node[box] (b1) {
{\large{\textbf{Answer Extraction}}} \\[4pt]
{Input:} $\bs$ (final solution)\\
Ask \lrm to extract $\ans$ from $\bs$\\
{Output:} $\ans$
};

\node[
    above=0.6cm of b1,
    draw=none,
    rectangle,
    rounded corners,
    fill=attn!30
] (start) {$(\bx, \br, \bs)$};

\node[box, right=of b1] (b2) {
{\large \textbf{Answer Position Identification}}\\[4pt]
{Input:} $(\br, \ans)$ \\
Ask \lrm to find a span of text $\bd$ \\
leading to the first logical arrival of $\ans$\\
{Output:} $\bd$
};

\node[box, right=of b2] (b3) {
\large{\textbf{Answer Verification} }\\[4pt]
{Input:} ($\bd, \ans$)\\
Ask \lrm whether $\ans \in \bd$\\
{Output:} $u \in  \sth{\text{True, False}} $
};

\node[box, right=of b3] (b4) {
\large{\textbf{Token-Index Extraction} }\\[4pt]
{Input:} ($\bd, \br, \ans$)\\
Extract the token position of $\ans \in \br$ \\
{Output:} Position $i^\ast \in [M]$
};

\node[
    above=0.6cm of b4,
    draw=none,
    rectangle,
    rounded corners,
    fill=attn!30
] (end) {$(\bx, \br, i^\ast)$};

\draw[->] (start.south) -- (b1.north);
\draw[->] (b1.east) -- (b2.west);
\draw[->] (b2.east) -- (b3.west);
\draw[->] (b3.east) -- (b4.west) node[midway, above] {True};
\draw[->] (b4.north) -- (end.south);

\draw[->]
    (b3.north)
    -- ++(0,0.8cm)
    -| (b2.north)
    node[pos=0.25, above] {False};

\end{tikzpicture}}
  \caption{\textbf{Training-Dataset Curation Process.} We use an LRM to (1) extract final answer \ans from final solution \bff, (2) identify the earliest position of \ans in the \rtx \bfr, and (3) verify that the position was correct. If it was, then we can extract the exact position of \ans from the \rtx at the final token-index extraction step; otherwise, we retry the identification step with feedback.}
  \label{fig:data-curation}
\end{figure}
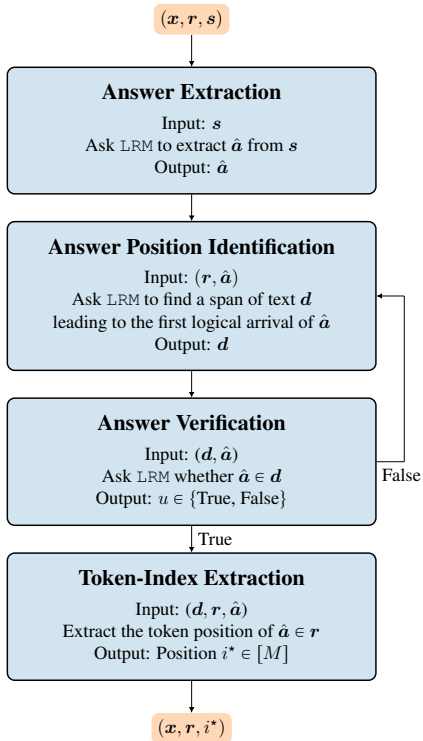

Our early answer extraction, identification, and verification pipeline (\cref{fig:data-curation}) is a critical component of our data curation process; at its core is an LRM that (1) extracts the final answer \ans from final solution \bff (answer extraction), (2) identifies the earliest logical arrival to \ans in \bfr (answer identification), (3) verifies that the extraction step was successful (answer verification). Finally, (4) we extract the exact position of \ans from the \rtx (token-index extraction). \looseness=-1

{\bf Rationale.} Extracting the position of \ans is not trivial. Human inspection and annotation of \rtx[s] is one route, but it is expensive and not scalable. Our early attempts at answer extraction, identification, and verification relied solely on fuzzy pattern matching, resulting in many false positives despite our best efforts to accommodate as many edge cases as possible. The primary challenge is that identifying the answer position within a \rtx is a semantic search problem that cannot be reliably solved with fuzzy or regex pattern matching for numerical answers, mathematical expressions, and code. Examples where pattern matching fails for these three is given in \cref{app:ans-pos-fail-modes}.
Using an LRM for all three cases confirmed that the earliest answer positions can be reliably extracted. \looseness=-1

{\bf Our Extract-Identify-Verify Pipeline.}
We first extract the final answer \ans from \bff, where it is explicitly marked, \eg with \texttt{\textbackslash boxed\{\}}, making extraction straightforward for the LRM. Next, the LRM identifies a span \bfs that both precedes and includes \ans. This ensures that \bfs is a unique substring of \bfr, allowing us to recover the exact token position of the earliest occurrence of \ans. The LRM then verifies that \bfs contains \ans. If verification fails, the LRM repeats the identification step, providing textual feedback listing all previously selected spans that did not contain \ans, reducing the chance of selecting the same span again. If no valid span is found within the retry limit, the corresponding \rtx is excluded from the training set. Otherwise, we locate \bfs in \bfr and retrieve the earliest answer token position \tok. \cref{alg:answer-extraction} in \cref{app:extract-and-verify} gives pseudocode for this procedure. These steps scale the construction of \method, which we use to train our probe classifier. 

{\bf Computational Cost.}
Our cost-benefit analysis in \cref{app:cost-analysis} shows that the inference-time benefits of \method substantially outweigh the cost of running this pipeline. \looseness=-1

\subsection{\method: Binary Probing Classifier}
\label{sec:binary-classifer}

Our approach entails training a small classification model $\btheta$ on the LRM's final-layer hidden states $\bh_i$ and making a binary prediction $b_i$ at each CoT position $i \in [M]$ (\cref{fig:our-method}). More specifically, our model reuses the same transformer block from the LRM and adds a prediction head. The weights of the transformer block are copied from the final block of the LRM, which we found performs slightly better than random initialization, and the prediction head is randomly initialized. During training, the task is to predict whether the first occurrence of the final answer has been generated (label $1$) or not (label $0$). Given the causality of the transformer block, every prediction depends on the history of the \rtx up to that point, but the predictions themselves are made independently of each other. Due to the inherent class imbalance of this early-exiting prediction task, our model is trained with class-weighted binary cross-entropy loss,  which for a single sample $(\bx, \br, \bs, \tok)$ is computed as: \looseness=-1
\begin{equation}
\label{eq:bce-loss}
L(\btheta) = -\frac{1}{M} \sum_{i=1}^{M} \Big[ w_1 \cdot y_i \cdot \log p_i + w_0 \cdot (1-y_i) \cdot \log(1-p_i)\Big],
\end{equation}
where $y_i = \mathbbm{1}(i \geq \tok) \in \binary$ denotes the ground-truth label corresponding to answer arrival and $p_i=\mathbb{P}_{\btheta}\pth{b_i = 1 \mid \bfx, \bfr_{\leq i} }$ is the predicted probability for each $i \in [M]$, with $M$ being the CoT length, and $w_0$ and $w_1$ the class weights. These weights are automatically computed from the training dataset using inverse frequency weighting as $w_0 = (n_0 + n_1)/2 n_0$ and $w_1 = (n_0 + n_1)/(2 n_1)$, where $n_0$ and $n_1$ are the total number of $0$ and $1$ labels in the training dataset, respectively. \looseness=-1

Here we note that \method is inspired by the findings of optimal-length reasoning literature; we seek to train a model on hindsight-optimal \rtx[s] to encourage \method to early-exit as soon as the final answer is generated. Unlike other methods, \method is \emph{free} of data-calibrated thresholding and is trained on several data sources (math, coding, and STEM problems) simultaneously. \looseness=-1
\section{Experiments}
\label{sec:experiments}

\subsection{Implementation Details}

\textbf{Models. } We train and evaluate our method on LRMs from two different model families: Qwen3-8B and Qwen3-14B \cite{yang2025qwen3}, and Ministral-3-8B-Reasoning-2512 and Ministral-3-14B-Reasoning-2512 \cite{liu2026ministral3}. We use Qwen3-30B-A3B-Thinking-2507 for our answer extraction, identification, and verification pipeline. Our trained models consist of a single transformer layer initialized from the final layer of the LRM and a binary prediction head. We compare \method against (1) prompt-based approaches, including Vanilla, NoThinking \citep{ma2025reasoning}, DEER \citep{yang2025dynamic}, and Dynasor \citep{fu2025efficiently}, and (2) a probe-based approach, Thought Calibration \citep{wu2025thought}. Vanilla is a direct evaluation of the LRM without any intervention. NoThinking prompts the model to skip the reasoning phase and generate the final solution \bff directly. DEER splits the reasoning into chunks, checks the average token probability after every chunk, and exits if it exceeds a threshold.
Dynasor periodically prompts the model to produce intermediate answers at fixed token intervals and triggers early exit when 8 consecutive answers are consistent.
Thought Calibration trains linear probes on the hidden representations of reasoning steps to automatically decide when to stop generation. We retrain these probes for our 4 models using their \textit{Supervised} method. For further details of the baselines' implementations, we refer to \cref{app:baseline_implementation}. \looseness=-1

\textbf{Datasets. } We form a training data mix with AIME (1983--2024) \cite{maa1983aime}, MATH \cite{lightman2024lets}, OpenCoder-SFT \cite{huang2024opencoder}, and OpenScience \cite{nvidia2025openscience}. We form our training datasets by sampling three \rtx[s] from each dataset, identifying the answer positions (see \cref{sec:extraction}), and assigning the corresponding training labels. We evaluate our method and all baselines on AIME 2025 \cite{maa1983aime}, MATH-500 \cite{lightman2024lets}, HumanEval \cite{chen2021evaluating}, and GPQA \cite{rein2024gpqa}. Additional details on our training datasets are available in  \cref{app:dataset}. \looseness=-1

\textbf{Training and Inference. } During training, we optimize for high performance on a holdout validation set for our prediction task; we choose our model based solely on how well it performs on the binary predictive task, without peeking at the evaluation dataset performance. Our validation metric of choice is the Macro-F1 score. Additional details on training \method are provided in \cref{app:training-details}. \looseness=-1

We use vLLM \cite{kwon2023efficient} with asynchronous requests to sample \rtx[s] when curating our training datasets. 
During inference with our trained model, a sliding window of the 10 most recent predictions is used, and the \texttt{</think>} token is injected when more than 50\% of the labels are 1 (majority voting). Our main results reported in this paper use a threshold of 0.7 to predict 1. Additional details on \method's selected inference parameters are provided in \cref{app:inference-details}. \looseness=-1
\vspace{-4pt}

\subsection{\method: Main Results}
\label{sec:main-results}
\cref{fig:pareto-frontier} shows the performance of \method and relevant baselines with respect to the Compression Rate (lower is better) and Accuracy (higher is better). To ensure a fair comparison, all methods are evaluated using the same \rtx[s] that are used in the vanilla baseline, except the NoThinking baseline, for which no \rtx is generated. While \cref{fig:pareto-frontier} shows that \method achieves a better overall compression-performance trade-off over prior methods, the same results---presented in table format---in \cref{tab:performance} show that \method also achieves best or second best performance on 28 out of 32 metrics. Notably, while methods such as \textit{Dynasor} achieve aggressive token reduction, they do so at the cost of significant accuracy degradation. \method consistently occupies a favorable position on the accuracy-efficiency Pareto frontier across all four evaluated LRMs, demonstrating that its advantages are robust to model architecture and scale. \looseness=-1
\vspace{-4pt}

\begin{wraptable}{r}{0.40\linewidth}
\vspace{-1em}
\centering
\caption{\textbf{Latency Analysis.} Latency and throughput benchmarks on MATH-500 problems for Qwen3-based vanilla and \method models. Values are mean $\pm$ 95\% CI.}
\label{tab:latency}
\small
\setlength{\tabcolsep}{4pt}
\renewcommand{\arraystretch}{0.95}
\begin{tabular}{@{}l c c@{}}
\toprule
\textbf{Method} & \textbf{Latency (s)} & \textbf{Throughput} \\
\textbf{} & \textbf{} & \textbf{(tok/s)} \\
\midrule
\multicolumn{3}{@{}l}{\cellcolor[rgb]{0.9,0.9,0.9}\textit{\textbf{Qwen3-8B}}} \\
\textit{Vanilla} & $32.68 \pm 9.59$ & $151.5 \pm 4.4$ \\
\textit{Terminator} & $14.10 \pm 6.27$ & $135.2 \pm 2.0$ \\
\midrule
\multicolumn{3}{@{}l}{\cellcolor[rgb]{0.9,0.9,0.9}\textit{\textbf{Qwen3-14B}}} \\
\textit{Vanilla} & $43.38 \pm 13.98$ & $98.0 \pm 2.0$ \\
\textit{Terminator} & $18.76 \pm 6.52$ & $90.6 \pm 0.8$ \\
\bottomrule
\end{tabular}
\vspace{-18pt}
\end{wraptable}


\subsection{Ablation Studies}
\label{sec:ablations}
We further run ablation studies on \method with respect to its (1) latency and throughput over the vanilla LRM, (2) performance against the truncated early-exit baseline, (3) out-of-distribution (OOD) performance, and (4) \method's recovery of our observed answer signal phenomena (\cref{fig:aligned-timeseries-all,fig:token-scatter}). Our results in this section will be reported mostly on Qwen3-8B alone, with results on Qwen3-14B, Ministral-3-8B, and Ministral-3-14B reported in \cref{fig:pareto-frontier,tab:performance}. \looseness=-1

\textbf{Latency Analysis. } \cref{tab:latency} shows the results of our latency analysis; \method halves the average latency over the vanilla LRM, despite incurring a small overhead of 10.8\% for Qwen3-8B and 7.5\% for Qwen3-14B, respectively. Note that as the base LRM size increases, \method incurs a proportionally smaller overhead since its architecture (a single transformer layer and an FFN) remains fixed. For our analysis, we develop a vLLM-compatible implementation of \method and compare with running the vanilla LRM in vLLM. Both methods are evaluated on the same subset of MATH-500 questions with a batch size of 1, disabled prefix caching, and on a single GH200. \looseness=-1

\begin{figure}[t]
    \centering
    \includegraphics[width=0.8\linewidth]{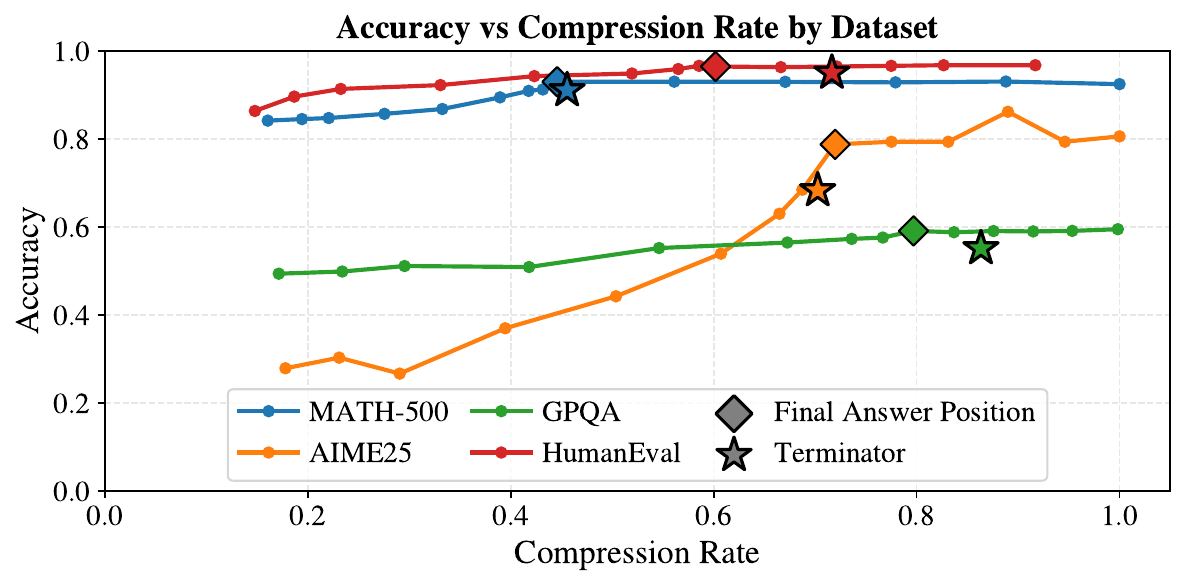}
    \caption{\textbf{Effects of Early \rtx Termination.} Test set \rtx[s] are evaluated after truncating them at various points via \texttt{</think>} and asking the LRM for a final solution and answer.
    Diamond-shaped markers show the hindsight-optimal answer position (not achievable). \method is close to optimality on all datasets. \looseness=-1 }
    \label{fig:truncation}
\end{figure}


\textbf{Hindsight-Optimal \rtx[s]. } Since \method is trained on hindsight-optimal reasoning length ($\mathsf{HORL}$) \rtx[s], it is natural to ask where \method lies on the accuracy-compression frontier relative to the ground-truth $\mathsf{HORL}$, which is shown in \cref{fig:truncation}. Each dot on the curves represents the accuracy when each \rtx was truncated early, and the LRM was forced to give a final solution and final answer. The diamond-shaped markers represent the position of the first occurrence of \ans, and therefore represent the points corresponding to hindsight-optimal reasoning. As expected, the accuracy remains constant after this point, showing that additional reasoning beyond \ans does not yield better performance. We plot \method alongside these curves to show how close it is to the hindsight-optimal \rtx length and performance. Even though the $\mathsf{HORL}$ baseline is not achievable by any method in principle, \method is notably close to it for all datasets. \looseness=-1

\textbf{OOD Evaluation. } We train separate models on each of the four tasks in our training dataset and evaluate them on the test datasets. Whereas our main results in \cref{fig:pareto-frontier} use \method trained on all four tasks, this experiment trains on one task at a time to assess OOD generalization. \cref{fig:heatmaps} reports compression rate and accuracy, with rows denoting the training dataset, columns denoting the test dataset, and cell values denoting test performance. The results show that compression is best \emph{in-distribution}, i.e., along the diagonal, but accuracy does not always follow the same pattern. For example, training on OpenScience yields the lowest GPQA accuracy despite GPQA being in-distribution, whereas training on the OOD OpenCoder-SFT dataset improves accuracy but worsens compression to 96\%. Thus, OOD evaluation can slightly improve accuracy but often delays exiting, thereby reducing token savings on data not seen during training. \looseness=-1 

\textbf{\method Recovers Early-Exit Signals.} We further highlight that using \method's predicted exit positions, we recover the \emph{same} event-locked averaging (\cref{fig:aligned-timeseries-all}) and ``thinking token'' frequency (\cref{fig:token-scatter}) phenomena. \cref{fig:timeseries-comparison} mirrors \cref{fig:aligned-timeseries-all}, but uses all test samples rather than 3,200 randomly selected training samples. Its left and center panels show event-locked average Token-Confidence using ground-truth and \method-predicted answer positions, respectively, while the right panel shows that prediction errors are concentrated near zero, with a median difference of 7. This alignment helps explain why \method recovers most of the same signal. Similarly, \cref{fig:scatter-overlay} parallels \cref{fig:token-scatter} by overlaying scatter plots computed from ground-truth and predicted answer positions. The inset axes show nearly identical above-diagonal percentages, indicating that \method preserves the same before/after ``thinking token'' usage biases. Together, \cref{fig:timeseries-comparison,fig:scatter-overlay} show that training \method on the LRM's hidden states is sufficient to independently recover the early-exit signals identified above, justifying our approach. \looseness=-1
\section{Related Work}

\textbf{Prompt Compression. } This line of work
is concerned with compressing the input prompt (or context) before passing it to an LLM.
Some methods use \textit{soft-prompts} \cite{mu2023learning,chevalier2023adapting,ge2024incontext,qin2024dodo} to compress tokenized inputs into a sequence of embeddings. These embeddings serve as the LLM's input, allowing richer expressivity, but they are not amenable to black-box LLMs and are difficult to analyze theoretically. Other methods use \textit{hard-prompts} \cite{jung2024discrete,jiang2024longllmlingua,pan2024llmlingua2,nagle2024fundamental}, keeping the final compressed input prompt fixed to the same token vocabulary as the LLM. 
\looseness=-1

\textbf{Efficient Reasoning. } 
Analogously to soft-prompt compression, \textit{latent} or \textit{continuous} reasoning is a technique where reasoning unfolds across latent hidden states rather than discrete tokens. Methods like Coconut \cite{hao2025training}, CCoT \cite{cheng2024compressed}, and Soft Thinking \cite{zhang2025softthinking} feed the LLM's output embeddings back into the input of the LLM during the reasoning stage, which significantly decreases the number of passes through the LLM before arriving at the final answer.
LightThinker \cite{zhang2025lightthinker} uses an idea similar to AutoCompressor \cite{chevalier2023adapting}, where each reasoning step is generated as discrete tokens first, compressed, and then the compressed summary of the reasoning thus far is fed back into the LLM to generate the next step. 
Other methods, like TokenSkip and C3oT \cite{xia2025tokenskip,kang2025c3ot}, are closer to hard-prompt compression, where a prompt compressor (or a summarization) model first compresses the \rtx[s] into much shorter versions, and the LLM is retrained on these shorter \rtx[s]. CoT-Valve \cite{ma2025cotvalve} extends this idea by introducing a parameter that adds explicit control over the reasoning length after fine-tuning. CALM \cite{schuster2022confident} also keeps its outputs in the token space, but saves compute by making layer-wise token-level early-exit decisions. 
\looseness=-1

\textbf{Early-Exit Reasoning. } These methods seek to make reasoning more efficient by terminating the \rtx early. All existing methods use a consistency-based approach, injecting the \texttt{</think>} token at various points to force the model to generate an answer or a useful signal. Some methods, like EAT \cite{wang2025entropy}, DEER \cite{yang2025dynamic}, ES-CoT \cite{mao2025earlystopping}, and Dynasor \cite{fu2025efficiently} are training-free; they track signals throughout the reasoning process and exit when a threshold is crossed. Other methods, like SpecExit \cite{yang2025specexit}, Learn To Stop \cite{liu2025answer}, Thought Calibration \cite{wu2025thought}, and FlashThink \cite{jiang2025flashthink} rely on training a separate probe classifier by using consistency as the main approach for gathering their training signals. By contrast, our work constructs a training signal to predict the immediate arrival of \ans, thereby training on hindsight-optimal length \rtx[s]. In addition, our work does not require threshold tuning on validation data, which is needed for Learn To Stop and Thought Calibration.
\looseness=-1

\section{Conclusion}
We present \method, an early-exit method for LRM reasoning. Training \method requires an optimal-length dataset of \rtx[s], which can be obtained through our robust answer extraction, identification, and verification pipeline. Furthermore, we provide novel analysis and insights into the behaviors of an LRM's (1) \tokconf during reasoning (\cref{fig:aligned-timeseries-all}), and (2) shift in ``thinking token'' usage. While our training data curation pipeline works well, future work can explore making training more efficient as tens of thousands of \rtx[s] are used to train \method. \looseness=-1

Finally, we liken the averaged result in \cref{fig:aligned-timeseries-all} to the field of \textit{event-related potential} (ERP) research. An ERP is a measurable brain response elicited by a sensory, cognitive, or motor event, captured by electroencephalogram (EEG) recordings \cite{luck2014erp}. However, EEG recordings are often noisy, so ERPs are estimated using time-locked statistical estimators (\eg averaging) across multiple EEG trials. While we do not claim that our findings will align exactly with ERP research, it is quite interesting that meaningful and observable signals can be extracted from LRMs using similar approaches, and we believe this warrants further exploration in future work. \looseness=-1

\begin{ack}
This work was partly supported by ARO Award W911NF2310062, ONR Award N000142412542, NSF 2443857 and the 6G$@$UT center within the Wireless Networking and Communications Group (WNCG) at the University of Texas at Austin. This research was supported by the Swiss National Science Foundation Grant No. 10004381. Our work used the Vista cluster at the Texas Advanced Computing Center (TACC) at The University of Texas at Austin. We thank the Center for Generative AI and TACC for providing the high-performance computing resources that supported this research.
\end{ack}

%
%

\bibliographystyle{plainnat}
\bibliography{main}

\appendix
\crefalias{section}{appendix}
\crefalias{subsection}{appendix}
\setcounter{topnumber}{2}

\section{Limitations and Broader Impacts}

\subsection{Limitations}
\label{app:limitations}
Our approach relies on the assumption that exiting at the earliest occurrence of the final answer is sufficient for the LRM to use that same answer in the final solution. Although rare, the LRM does not always commit to the exited answer in the \rtx, which can cause the exited response to be marked incorrect even when the vanilla LRM, using full reasoning, eventually obtains the correct answer. Furthermore, the practicality of our approach depends on the final answer being present within the \rtx, which might not always be the case, especially for open-ended generation tasks.

Each base model also requires its own probe. \method is initialized as an exact copy of the final layer of the LRM it attaches to, including its architecture and weights, and hidden-state geometry differs substantially across model families, so we do not expect probes to transfer across families without fine-tuning. This is a light requirement in practice, since \method is a one-layer transformer with a linear head, and cross-family transfer is an interesting direction for follow-up work. Finally, our evaluation covers 8B and 14B models, which enable controlled comparisons across families at a fixed scale and across scales within a family. Prior work has not found early-exit effectiveness to differ meaningfully between small- and large-scale LRMs \cite{yang2025dynamic,wu2025thought,fu2025efficiently}.

\subsection{Broader Impacts}
\label{app:broader-impacts}
This work aims to advance machine learning research by improving the efficiency of reasoning in large language models through early-exit mechanisms based on model confidence. Potential benefits include reduced energy consumption, faster inference, and lower deployment costs, thereby improving the scalability and environmental sustainability of language model systems.

At the same time, confidence-based early termination introduces the risk of over-confidence and premature conclusions, particularly in settings where incorrect outputs may have significant consequences. These risks underscore the importance of careful calibration, evaluation, and responsible deployment practices. Our work does not introduce new classes of societal harm beyond those already associated with large language models, but it highlights the need for continued research into reliable confidence estimation and safe adaptive inference.

\section{Additional Details on Our Methods}
\label{app:methods}

\subsection{Early Answer Extraction, Identification, and Verification}
\label{app:extract-and-verify}

\cref{alg:answer-extraction} contains pseudocode for our pipeline. \tok, the index of the earliest token position containing \ans, is used to construct the label set of our training data by setting all positions prior to \tok to 0 and setting all positions at or after \tok to 1. Each of the three steps (extraction, identification, and verification) requires separate calls to an LRM; please refer to our codebase for details on the exact system prompts that we used for each step.

We label every position at or after \tok as 1 rather than supervising only near the stopping boundary. Supervising only near the stopping boundary would make the training label a one-hot vector over a sequence of up to $32$k tokens, so a trivial all-zero predictor achieves near-perfect accuracy and the probe fails to learn any exit point. Our formulation instead asks ``Has the final answer arrived?'' rather than ``Is this the exact stopping point?'', which keeps the label dense enough to learn from and lets \method still find a valid exit even when the earliest arrival has already passed.

\begin{algorithm}[t]
\caption{Answer Span Extraction, Identification, and Verification With Feedback}
\label{alg:answer-extraction}
\begin{algorithmic}[1]
\STATE \textbf{Input:} \rtx \bfr, final solution \bff, \texttt{LRM}, \texttt{tokenizer}, max retries $K$
\STATE \textbf{Output:} answer position \tok (token index where answer is reached)
\STATE
\STATE \textit{// Extract final answer value from model output}
\STATE $\ans \gets \texttt{LRM}(\text{``Extract the final answer from: ''} + \bff)$
\STATE
\STATE \textit{// Iteratively extract and verify span with feedback}
\STATE $\bfz \gets \emptyset$ \quad \textit{// feedback provided to the LRM}
\FOR{$k = 1, \ldots, K$}
    \STATE \textit{// Ask LRM to identify a string span containing the first occurrence of \ans in \bfr}
    \STATE $\bfs \gets \texttt{LRM}(\text{``Find first occurrence of ''} + \ans + \text{`` in: ''} + \bfr + \bfz)$
    \STATE
    \STATE \textit{// Verify the identified span contains the answer}
    \STATE $v \gets \texttt{LRM}(\text{``Does ''} + \bfs + \text{`` contain ''} + \ans + \text{``?''})$
    \STATE
    \IF{$v == \text{true}$}
        \STATE \textbf{break} \quad \textit{// span verified, proceed}
    \ENDIF
    \STATE
    \STATE $\bfz \gets \bfz + \text{``\textbackslash n Previous span '' + \bfs + `` was incorrect, try again''}$
\ENDFOR
\STATE
\STATE \textit{// Pattern match span text to get character-wise positioning of the span}
\STATE $c \gets \texttt{FuzzyMatch}(\bfs, \bfr)$  \quad \textit{// $c$ is an integer-based character index of \bfs in \bfr}
\STATE
\STATE \textit{// Convert to token position where answer is reached}
\STATE $\tok \gets \texttt{CharToTokenPos}(c + \texttt{len}(\bfs), \bfr, \texttt{tokenizer})$
\STATE
\STATE \textbf{return} \tok
\end{algorithmic}
\end{algorithm}

\subsection{Failure Modes of Pattern Matching for Answer Position Identification}
\label{app:ans-pos-fail-modes}
The following are examples of failure modes for pattern matching while extracting the earliest arrival of \ans from an \rtx, thus motivating our LRM-based extraction pipeline:
\begin{enumerate}
    \item \textbf{Numerical answers.} A numerical value may appear frequently throughout the \rtx in intermediate calculations, problem restatements, or discarded solution attempts, making it impossible to distinguish these occurrences from the true final answer by pattern alone. For example, if the final answer is \texttt{x = 42}, the value \texttt{42} may appear dozens of times in prior reasoning steps without having any connection to a logical arrival to that answer. The \rtx could simply consider \texttt{42} without steps for arriving to it as an answer, or \texttt{42} could appear out of pure coincidence as an intermediate value during calculation.
    \item \textbf{Mathematical expressions.} The same mathematical object can be represented in many syntactically distinct forms. For instance, $\texttt{x}^{\texttt{2}}$, \texttt{x**2}, \texttt{pow(x,2)}, and \texttt{x} $\cdot$ \texttt{x} are semantically equivalent but would not be matched by any single pattern. Differences in \LaTeX{} formatting, Unicode symbols, and whitespace further compound this.
    \item \textbf{Python functions.} A Python function may not appear as a contiguous block anywhere in the \rtx; instead, it may be generated line by line, interspersed with commentary. The final reconstructed answer, therefore, does not exist verbatim in the text, making positional matching fundamentally ill-posed.
\end{enumerate}

\subsection{Training Dataset Details}
\label{app:dataset}
Our training dataset consists of \rtx[s] from AIME (1983--2024) \cite{maa1983aime}, MATH \cite{lightman2024lets}, OpenCoder-SFT \cite{huang2024opencoder}, and OpenScience \cite{nvidia2025openscience}. All 933 samples and all 12,000 samples from the AIME (1983--2024) and MATH datasets, respectively, are used. We randomly select 12,000 samples from the \texttt{educational\_instruct} subset of the OpenCoder-SFT-Stage2 dataset, which we refer to as ``OpenCoder-SFT'' in the main paper. This subset consists of generated and validated Python coding examples. Our sampling procedure for this dataset was not uniform, unlike the others. Instead, problems are grouped by their \texttt{entry\_point} field, and sampling is split into rounds, with each round randomly sampling one problem from that group without replacement. Finally, we randomly sample an additional 12,000 samples from the \texttt{OS-Q3-235B-4} subset of the OpenScience dataset. This subset consists of multiple-choice STEM question-answer pairs that were synthetically generated from Qwen3-235B-A22B \cite{yang2025qwen3}.

Three \rtx[s] are sampled per problem by the target LRM (we used Qwen and Mistral models), yielding a set of approximately 110,799 \rtx[s] per LRM. The respective answer positions for each set of \rtx[s] is obtained with our extraction method outlined in \cref{sec:extraction}. However, this procedure is not perfect, as even with our retry logic, the answer extractor, identifier, and verifier LRM (Qwen3-30B-A3B-Thinking-2507) cannot always identify the earliest final answer position. Thus, all three of these steps are successful for roughly 70\%--80\% of \rtx[s]. Finally, a training-ready dataset for each LRM is formed by preparing label vectors (based on the answer positions), loss masks (based on the positions of \texttt{<think>} and \texttt{</think>}), and tokenizing the \rtx[s].

\textbf{Retained and Dropped \rtx[s]. }
\cref{tab:pipeline-retention} breaks down pipeline retention by data source. Dropped \rtx[s] are longer than retained ones in every source: $1.6\times$ on MATH and AIME (1983--2024), $1.7\times$ on OpenCoder-SFT, and $2.5\times$ on OpenScience. Retention also varies by source, from 83.5\% (OpenScience) down to 63.9\% (OpenCoder-SFT) and 64.3\% (AIME). We attribute most of this to a single cause: the pipeline drops a \rtx when it cannot localize a unique earliest arrival, and longer \rtx[s] contain more restatements, abandoned branches, and verification passes. AIME \rtx[s] are the longest and among the least retained, consistent with this account. OpenCoder-SFT is an exception that we attribute to the answer format, since a code answer is rarely a short extractable span, and so a first occurrence is often ill-posed. Filtering applies only to label construction; evaluation uses the full unfiltered benchmarks, so no reported numbers benefit from it.

\begin{table}[t]
\centering
\caption{\textbf{Pipeline Retention by Data Source.} Number of Qwen3-8B \rtx[s] processed by our answer extraction, identification, and verification pipeline, the percentage retained for training, and the median length (in generated tokens) of retained and dropped \rtx[s].}
\label{tab:pipeline-retention}
\begin{tabular}{@{}lrrrr@{}}
\toprule
\textbf{Source} & \textbf{\rtx[s]} & \textbf{Retained} & \textbf{Median Len.\ Retained} & \textbf{Median Len.\ Dropped} \\
\midrule
MATH              & 35,693  & 77.9\% & 3,398 & 5,427 \\
OpenScience       & 35,996  & 83.5\% & 2,494 & 6,274 \\
OpenCoder-SFT     & 35,958  & 63.9\% & 3,015 & 5,190 \\
AIME (1983--2024) & 2,698   & 64.3\% & 8,817 & 13,834 \\
\midrule
\textbf{All}      & \textbf{110,345} & \textbf{74.8\%} & \textbf{3,037} & \textbf{5,691} \\
\bottomrule
\end{tabular}
\end{table}

\textbf{Label Quality. }
We offer two forms of validation that the pipeline's exit positions are accurate. First, \cref{fig:truncation} shows, for each benchmark, where the chosen exit position falls on the accuracy-compression trade-off (diamond markers), alongside accuracy from truncating \rtx[s] before and after that point. The curve is flat after the exit position, so additional reasoning past it yields no accuracy benefit. If the pipeline were latching onto superficial early answers rather than the first logical arrival, consistent accuracy gains would remain available past the labeled positions.

Second, we manually inspected 30 randomly sampled \rtx[s] from our two hardest, longest-context benchmarks, AIME25 ($n=20$) and GPQA ($n=10$), checking whether the exit position was preceded by a derivation showing how the final answer was reached. In 29 of 30 cases, it was; the single failure was a false positive, disjoint from any earliest logical arrival. We note that precise false-positive and false-negative rates are challenging to quantify, since these \rtx[s] are lengthy and often contain several sections where the final answer is deliberated without concrete steps constituting a logical arrival.

\begin{figure*}[h]
    \centering
    \includegraphics[width=0.9\linewidth]{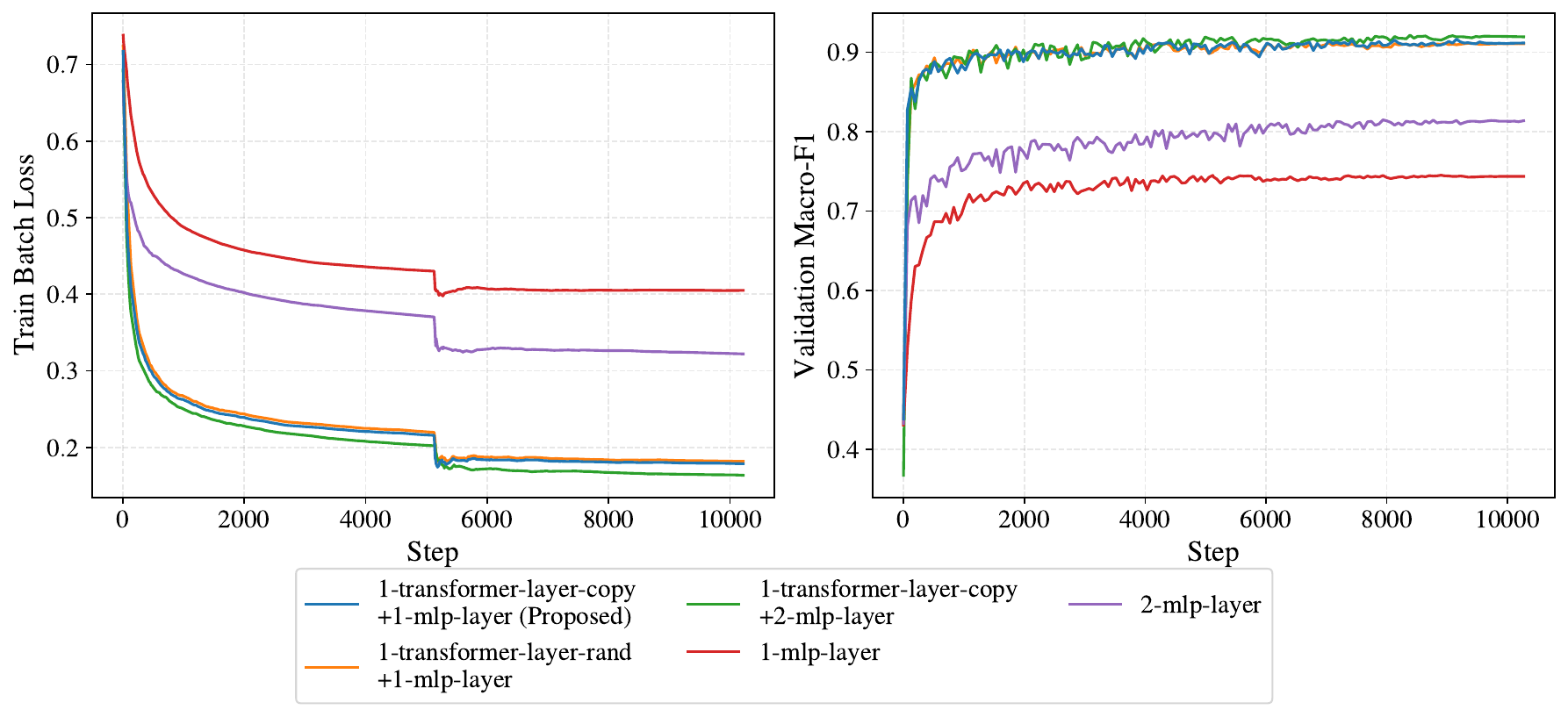}
    \caption{\textbf{Training Convergence.} During training, our proposed \method architecture achieves the best balance between the highest Macro-F1 validation scores and the lowest compute overhead.}
    \label{fig:convergence-curves}
\end{figure*}

\subsection{\method Training Details}
\label{app:training-details}
All models are trained using a consistent set of hyperparameters: batch size of 2, gradient accumulation over 8 steps, no dropout, and a weight decay of 0.01. We initialize the learning rate at $2 \times 10^{-4}$. Training follows a cosine annealing schedule with 100 steps of linear warmup, decaying to a minimum learning rate of $1 \times 10^{-6}$.

Optimization is performed with the AdamW optimizer ($\beta_1 = 0.9$, $\beta_2 = 0.999$, $\epsilon = 1 \times 10^{-8}$), and all runs use a fixed random seed of 1337. Each model consists of a single transformer layer and a single-layer binary classification head. Training proceeds for 2 epochs.

\cref{fig:convergence-curves} shows the convergence curves of the training loss and the Macro-F1 scores on a small validation set during training. The validation set is a held-out set of examples from our curated dataset, but is not seen during training. Macro-F1 is the unweighted average of per-class F1 scores, treating all classes equally regardless of their frequency, making it a robust and fair metric for evaluating performance on class-imbalanced datasets. The following architectures are studied: 
\begin{enumerate}
    \item \textbf{1-transformer-layer-copy+1-mlp-layer:} The proposed design, consisting of a single transformer layer with copied initialization followed by a one-layer MLP
    \item \textbf{1-transformer-layer-rand+1-mlp-layer:} A variant with random initialization of the transformer layer
    \item \textbf{1-transformer-layer-copy+2-mlp-layer:} A variant with a two-layer MLP head
    \item \textbf{1-mlp-layer, 2-mlp-layer:} Two MLP-only baselines with one and two layers, respectively.
\end{enumerate}
The transformer block is critical: MLP-only architectures incur a substantial $10$--$17\%$ drop in Macro-F1. When the transformer layer is included, increasing the MLP head from one to two layers yields only a marginal improvement of ${\sim}0.9\%$, despite additional computational cost. The choice of initialization for the transformer layer (copied vs.\ random) yields a tiny improvement (${\,\sim}0.1\%$); we therefore adopt copied initialization by default, as it incurs no additional cost.

\subsection{\method Inference Details}
\label{app:inference-details}
\begin{figure}[h]
    \centering
    \includegraphics[width=0.9\columnwidth]{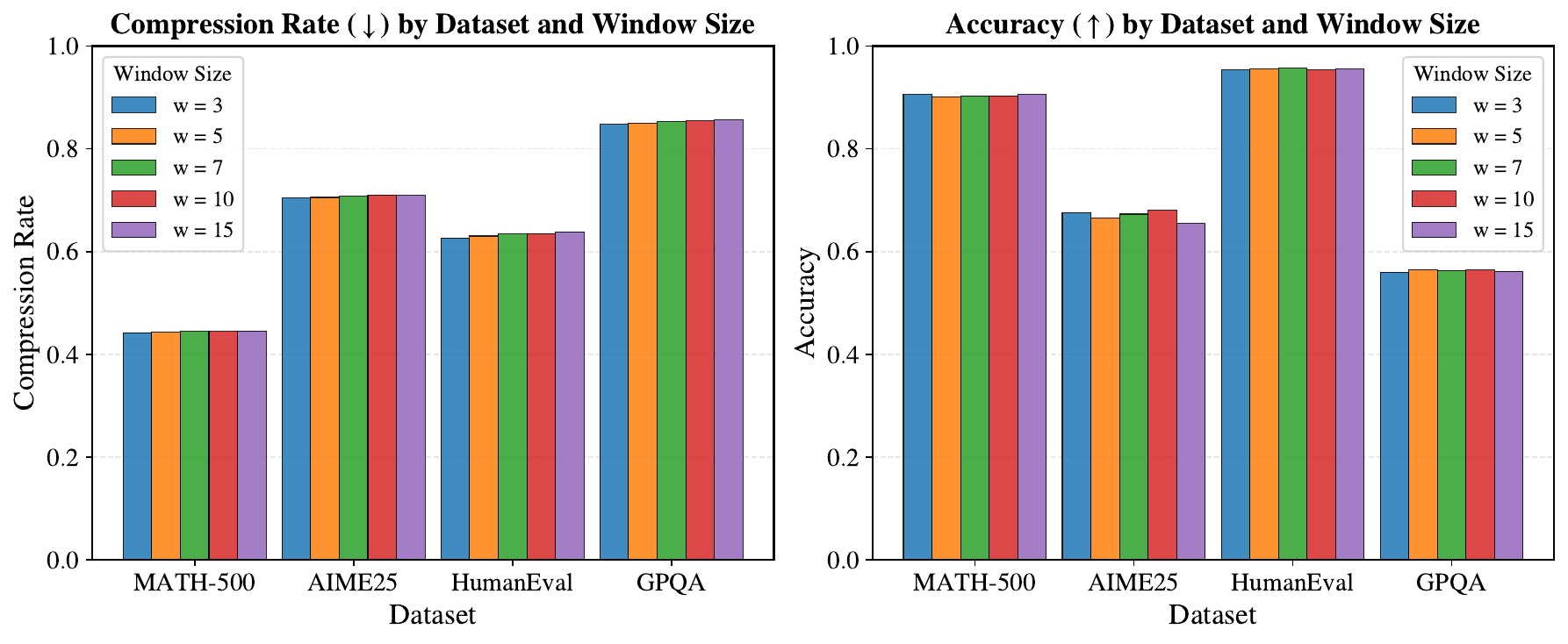}
    \caption{\textbf{Effect of Varying Window Size.} We fix the threshold to 0.7 and vary the size of the window used by \method at inference-time. \method is robust to changes in the window size; \cref{fig:aligned-predictions-all-app,fig:predictions-aime,fig:predictions-math,fig:predictions-humaneval,fig:predictions-gpqa} show that the predicted probability of \method during CoT generation tends to only increase. This means that once the predicted probability crosses the threshold, it tends not to fall back below it, and the choice of window size has little or no effect.}
    \label{fig:window-size-bars}
\end{figure}

\begin{figure*}[!t]
    \centering
    \includegraphics[width=1.0\textwidth]{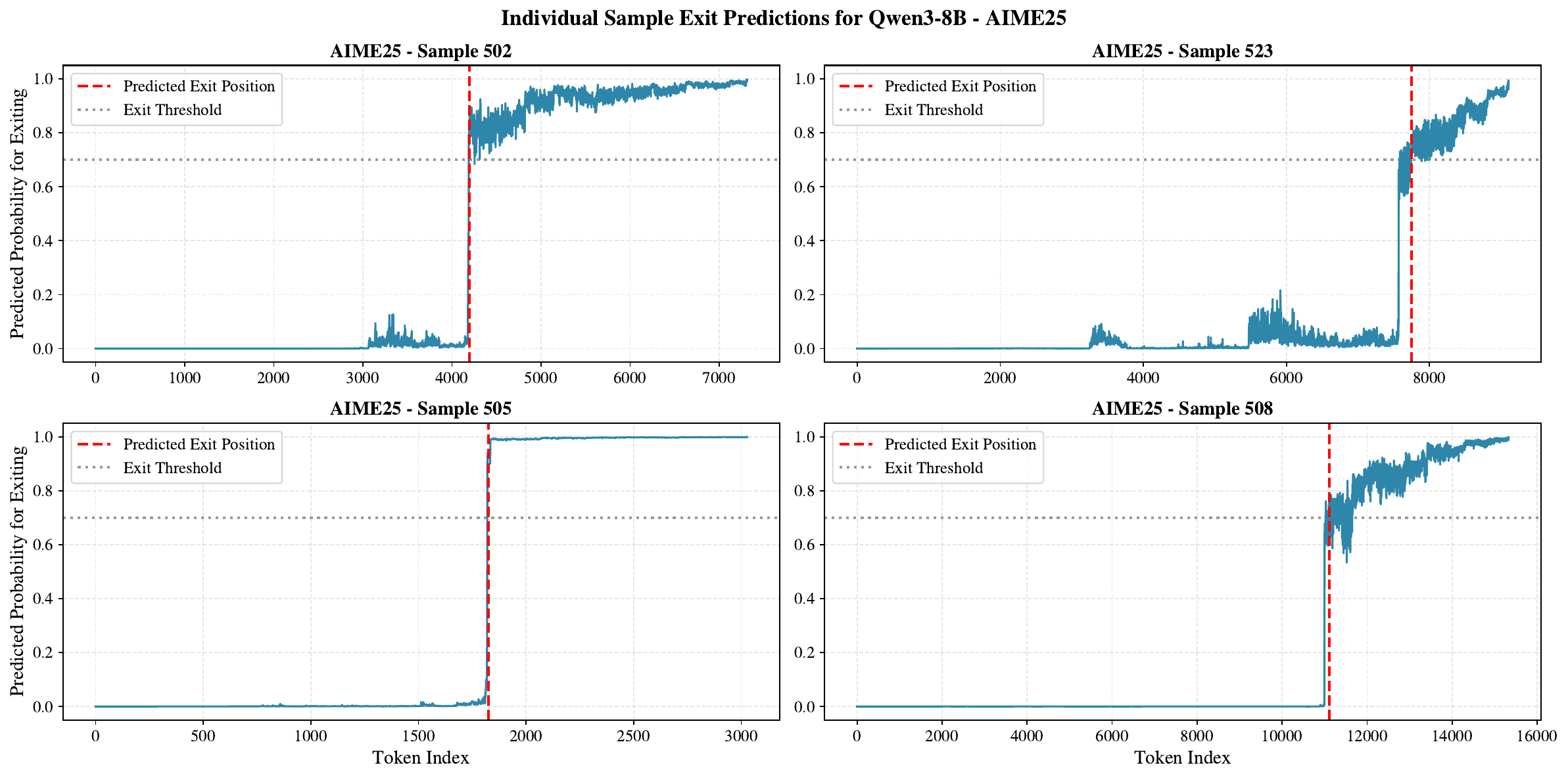}
    \caption{\textbf{Predicted Probabilities for AIME25.} \method's predicted probability stream for early-exiting on four randomly chosen samples from AIME25.}
    \label{fig:predictions-aime}
\end{figure*}

\begin{figure*}[!t]
    \centering
    \includegraphics[width=1.0\textwidth]{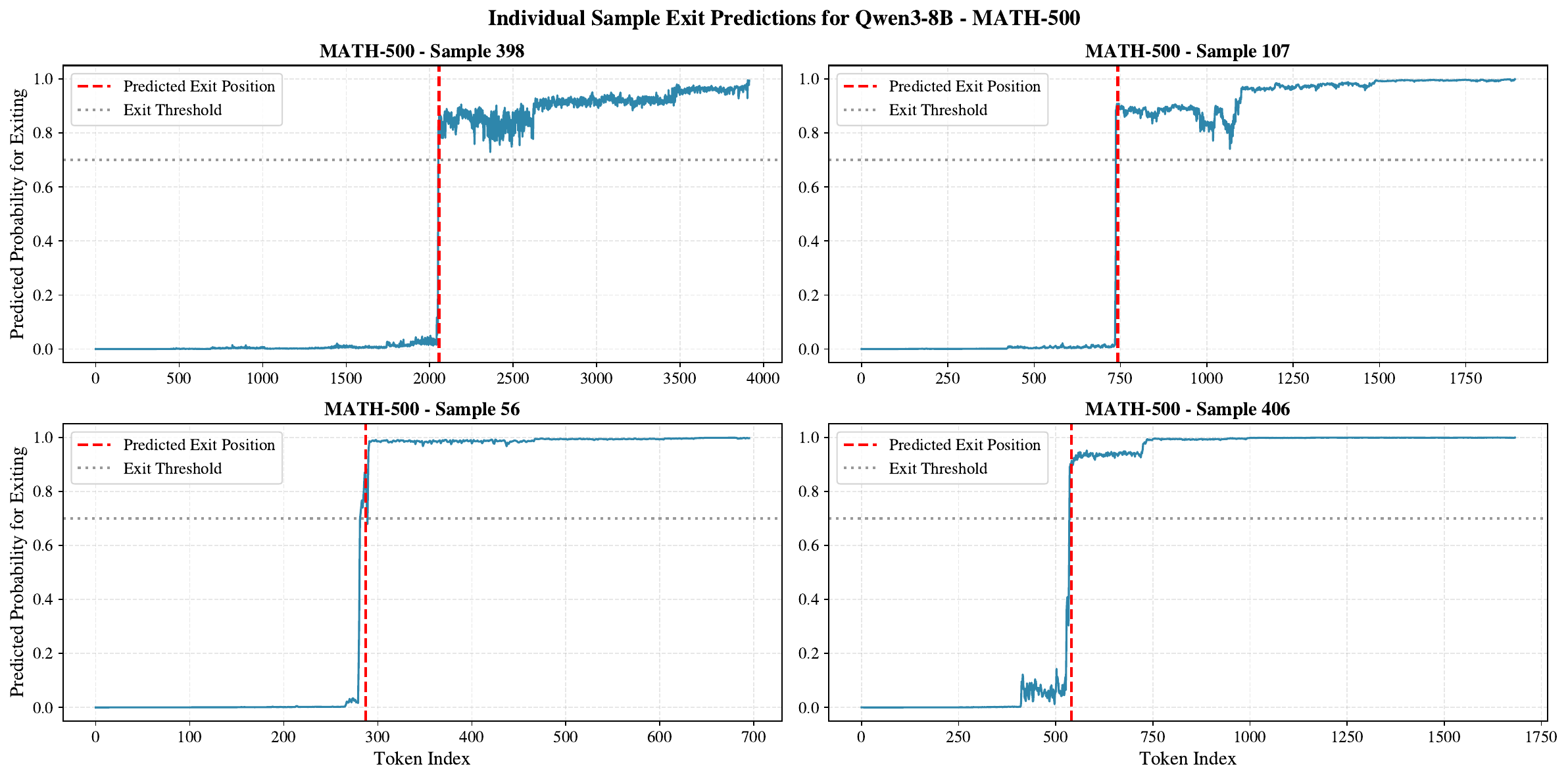}
    \caption{\textbf{Predicted Probabilities for MATH-500.} \method's predicted probability stream for early-exiting on four randomly chosen samples from MATH-500.}
    \label{fig:predictions-math}
\end{figure*}

\begin{figure*}[!t]
    \centering
    \includegraphics[width=1.0\textwidth]{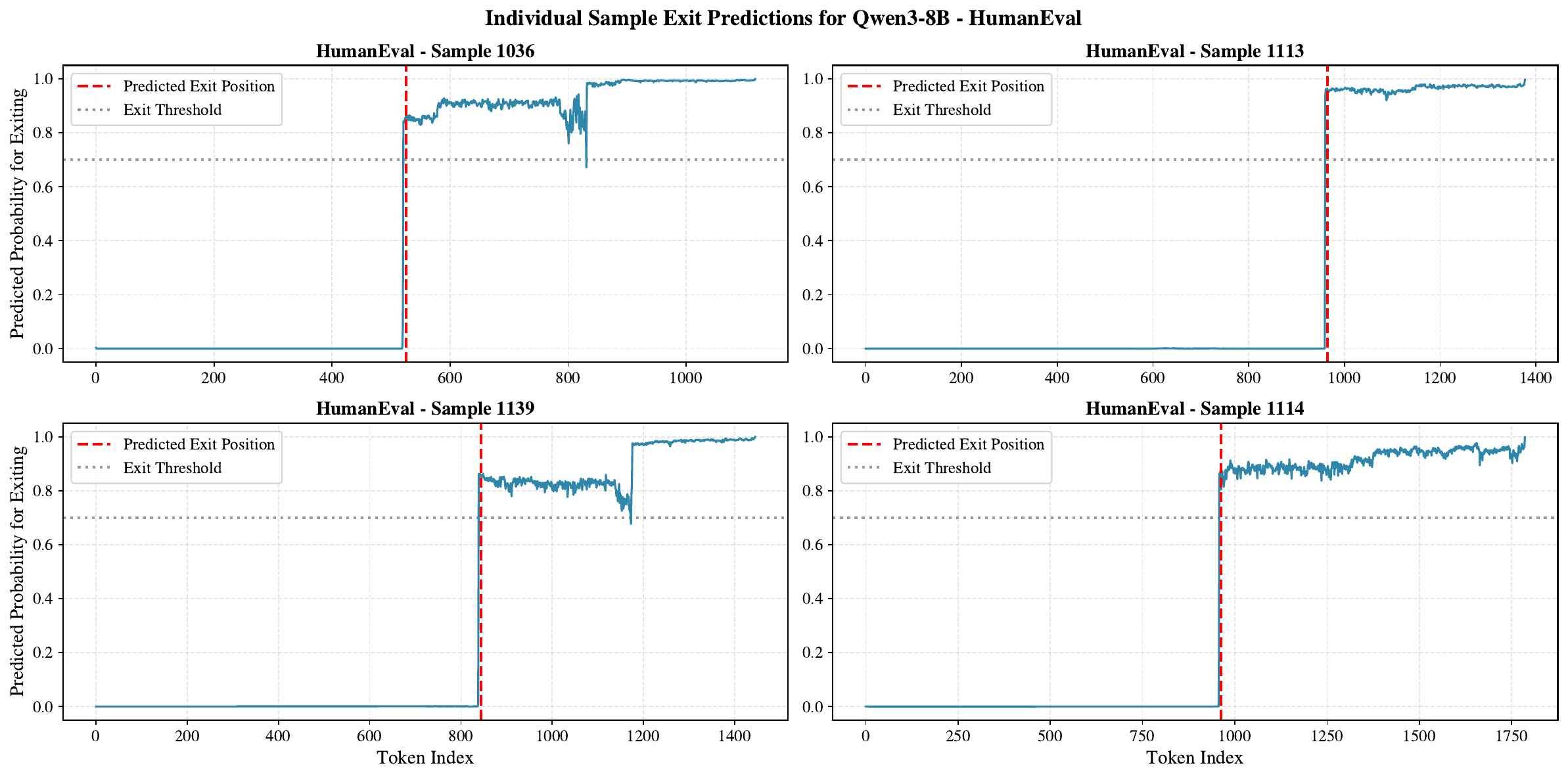}
    \caption{\textbf{Predicted Probabilities for HumanEval.} \method's predicted probability stream for early-exiting on four randomly chosen samples from HumanEval.}
    \label{fig:predictions-humaneval}
\end{figure*}

\begin{figure*}[!t]
    \centering
    \includegraphics[width=1.0\textwidth]{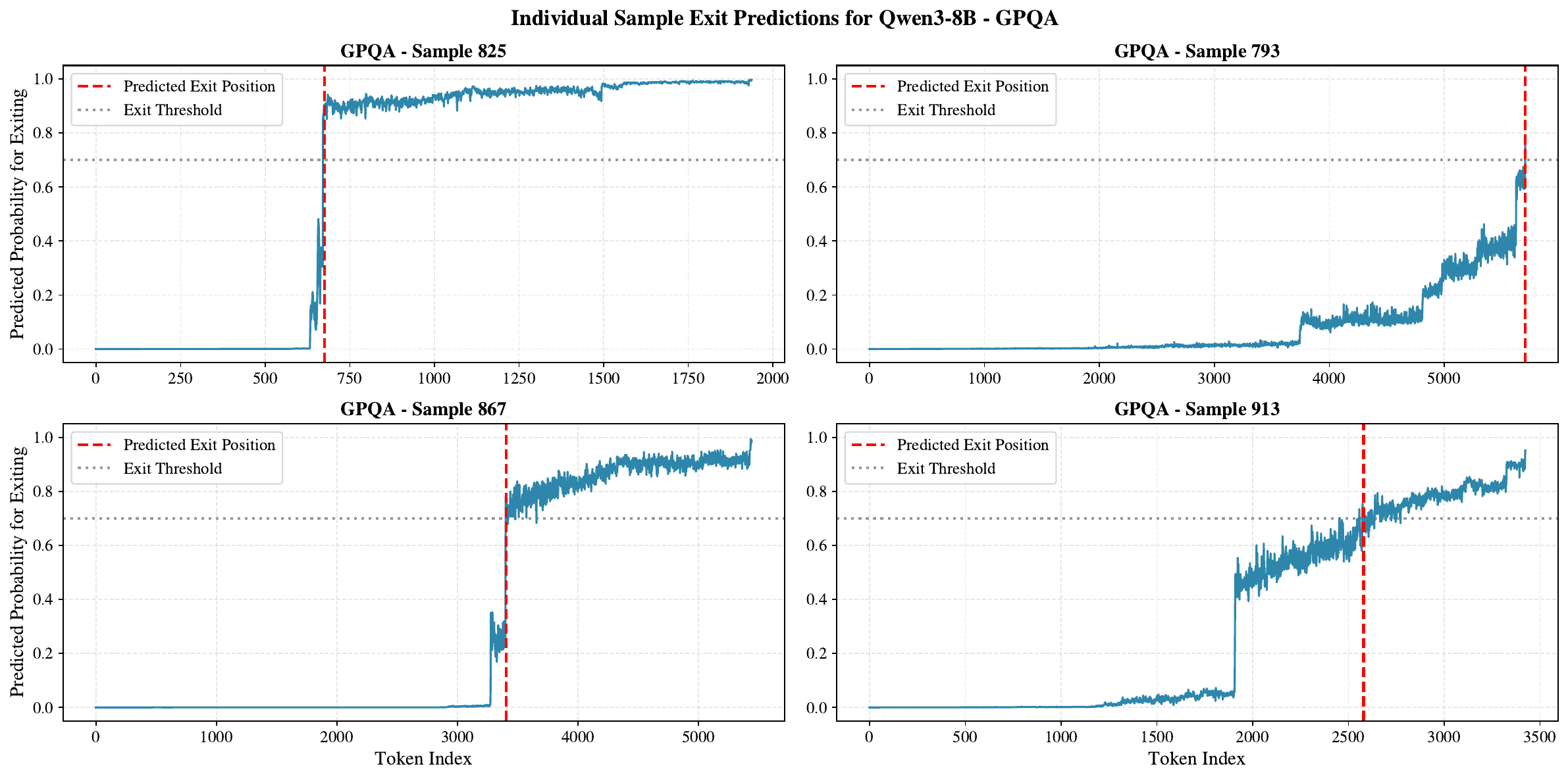}
    \caption{\textbf{Predicted Probabilities for GPQA.} \method's predicted probability stream for early-exiting on four randomly chosen samples from GPQA.}
    \label{fig:predictions-gpqa}
\end{figure*}

\textbf{Window Size and Majority Voting. } 
Our proposed strategy for early-exiting based on \method's predictions uses majority voting over a window of the $10$ most recent predictions. The window and majority vote serve as a consistency check, preventing a single spurious prediction from triggering an early exit. With Terminator scoring ${\sim}91\%$ Macro-F1 and ${\sim}93\%$ accuracy on the validation set, a window of $10$ yields fewer than one incorrect prediction on average, making the strict majority (at least $6$ of $10$) a principled default. \cref{fig:window-size-bars} explores the effect of varying the window size while fixing the threshold to $0.7$. Performance is stable across all window sizes, confirming that the method is robust to this hyperparameter. As shown in \cref{fig:aligned-predictions-all-app,fig:predictions-aime,fig:predictions-math,fig:predictions-humaneval,fig:predictions-gpqa}, once the predicted exit probability exceeds the threshold, it rarely drops below, so the window size has minimal impact on the majority vote.

\textbf{Probability-Based Aggregation. }
\cref{fig:predictions-aime,fig:predictions-math,fig:predictions-humaneval,fig:predictions-gpqa} plot the predicted exit probability over time for randomly sampled problems. The signal is step-like, staying low then rising sharply and plateauing at the exit position; where it is not a sharp step (some GPQA problems), it remains close to monotonic, and once it crosses the threshold, it rarely falls back. Averaging or EMA smoothing over a near-monotonic step function therefore adds no signal over thresholding directly; its main effect is to delay the crossing point, which is already reachable by adjusting our window size and threshold. We did not pursue calibrated sequential rules because they require held-out data representative of the evaluation distribution, which is not always available, and separate calibration parameters per benchmark, which assumes the problem domain is known at inference time.

\begin{figure}[h]
    \centering
    \includegraphics[width=0.9\columnwidth]{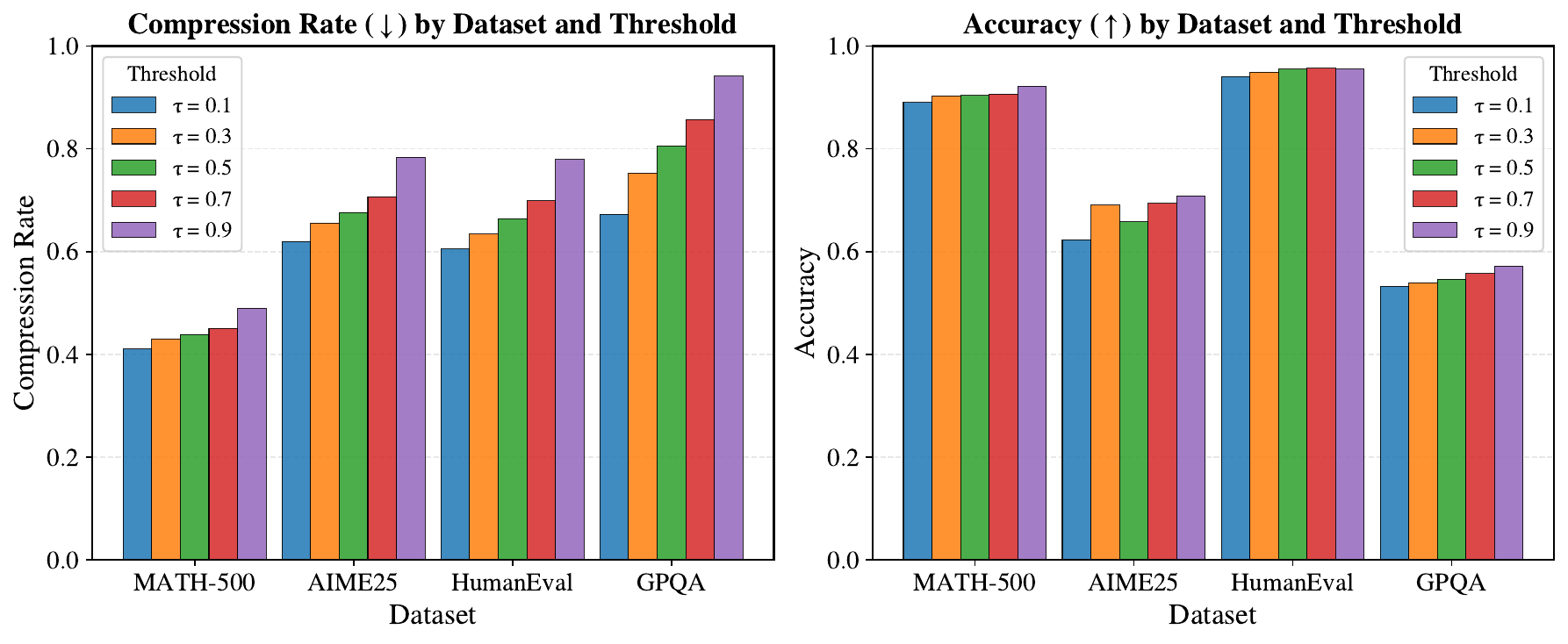}
    \caption{\textbf{Effect of Varying Predictive Threshold.} We fix the window size to $10$ and vary the threshold $\tau$ used by \method to predict whether to exit early. \method is robust to changes in the predictive threshold: lowering $\tau$ makes the method more aggressive, leading to substantially higher compression rates, while accuracy changes only modestly across datasets. In particular, datasets such as MATH-500 and HumanEval exhibit little performance degradation across thresholds from $\tau = 0.1$ to $\tau = 0.9$, suggesting that \method learns a stable confidence signal for identifying viable early-exit points.}
    \label{fig:threshold}
\end{figure}

\textbf{Thresholding. } 
\cref{fig:threshold} shows how the compression rate and the accuracy change as a result of varying the threshold used by \method to predict 0 or 1. For example, a threshold $\tau = 0.7$ requires \method to have a predictive confidence of at least 0.7 to predict 1. Interestingly, varying the threshold has a greater impact on compression rate than on accuracy; setting $\tau = 0.1$ yields 8\% and 17\% better compression rates for MATH-500 and HumanEval, respectively, compared to $\tau = 0.9$, with little change in accuracy. For example, from $\tau = 0.9$ to $\tau = 0.1$ yields a $4\%$ drop in accuracy but a $27\%$ improvement in the compression rate for GPQA. Overall, this suggests that \method learns a stable confidence signal for identifying viable early-exit points: lowering the threshold makes the method more aggressive, substantially improving compression while only modestly affecting accuracy.

\textbf{\method's Context Length.}
In general, we stick to training and evaluating \method with context lengths up to $32,768$, since those are the lengths of the training \rtx[s] anyway. However, Mistral models have unusually long \rtx[s] only for AIME problems, so we train and evaluate \method for a max length of $65,536$ tokens on AIME for Mistral models.

\subsection{Implementation of the Baseline Methods}
\label{app:baseline_implementation}
\textbf{Dynasor. } Dynasor \cite{fu2025efficiently} works by interrupting reasoning at regular token intervals (e.g. every 32, or 64 tokens) by injecting the prompt ``Oh, I suddenly got the answer to the whole problem, Final Answer: \/boxed\{'' to extract the model's current answer. The method decides to exit early when consistent answers appear across multiple probing intervals for at least $w$ times. In our experiments, we set $w$=8 and use a token interval of 64 tokens, following their so-called \textit{mild} configuration setup.

\textbf{Thought Calibration. } \citet{wu2025thought} propose training a linear probe to predict optimal stopping points during reasoning generation. The method first segments reasoning trajectories (\rtx[s]) into individual steps, delimited by ``\textbackslash n\textbackslash n'' and containing words like ``wait'' or ``but''. Three probe variants are introduced: \textit{Supervised} predicts whether the LRM is correct based on current thoughts; \textit{Consistent} predicts whether the current answer (\ans) matches the final answer (\bfa); and \textit{Novel Leaf} predicts whether the current step is a leaf node but not novel. 

The stopping decision is controlled by two hyperparameters: \textit{tolerance} $\delta$ (the maximum acceptable risk of stopping incorrectly) that implies a threshold $\lambda$ (the calibrated probe score cutoff that triggers stopping), and \textit{window size} (the number of consecutive reasoning steps averaged to smooth predictions before threshold comparison). We retrain the \textit{Supervised} and \textit{Consistent} probes on the S1-K dataset \cite{muennighoff2025s1simpletesttimescaling} for all four models in our experiments: Qwen3-8B, Qwen3-14B, Ministral-3-8B-Reasoning-2512, and Ministral-3-14B-Reasoning-2512. For inference, we use the hyperparameters specified in Table~\ref{tab:tc-hyperparams}. As shown in Table~\ref{tab:tc-hyperparams}, the Qwen3 models require lower thresholds compared to the Ministral models, as their probe outputs yield systematically lower confidence scores. Similar model-specific calibration requirements have been observed in prior work \cite{yang2025dynamic}.



\begin{table}[h]
\centering
\caption{Hyperparameters used for \textit{Supervised} and \textit{Consistent} linear probes for Thought Calibration. Ministral models' full names are Ministral-3-xxB-Reasoning-2512. Tolerance and threshold refer to $\delta$ and $\lambda$, respectively.}
\label{tab:tc-hyperparams}
\small
\begin{tabular}{l|ccc|ccc}
\hline
 & \multicolumn{3}{c|}{\textbf{Supervised}} & \multicolumn{3}{c}{\textbf{Consistent}} \\
\textbf{Model} 
& tolerance 
& threshold 
& window size 
& tolerance 
& threshold 
& window size \\
\hline
Qwen3-8B 
& 0.25 & 0.6526 & 10 
& 0.25 & 0.8790 & 10 \\

Qwen3-14B 
& 0.25 & 0.6377 & 10 
& 0.25 & 0.8679 & 10 \\

Ministral-3-8B 
& 0.10 & 0.8173 & 10 
& 0.025 & 0.9973 & 10 \\

Ministral-3-14B 
& 0.10 & 0.8306 & 10 
& 0.025 & 0.9973 & 10 \\
\hline
\end{tabular}\\[0.1cm]
\end{table}

We reported the results for the \textit{Supervised} probe in Table~\ref{tab:performance}, as it performed better across test datasets than the \textit{Consistent} probe. For comparison, we report the results for both probes in Table~\ref{tab:tc-performance}. Note that for the \textit{Consistent} probe with Ministral models, tolerance 0.025 represents the smallest feasible setting among values suggested by \citet{wu2025thought}. The smallest possible setting, being tolerance 0.01, yields threshold=1.0, resulting in no compression.
\begin{table}[h]
\centering
\caption{Performance comparison of \textit{Supervised} and \textit{Consistent} probes for Thought Calibration across models and tasks. Ministral models' full names are Ministral-3-xxB-Reasoning-2512.}
\label{tab:tc-performance}
\small
\begin{tabular}{l|cc|cc|cc|cc}
\hline
 & \multicolumn{2}{c|}{\textbf{MATH-500}} & \multicolumn{2}{c|}{\textbf{AIME 2025}} & \multicolumn{2}{c|}{\textbf{GPQA}} & \multicolumn{2}{c}{\textbf{HumanEval}} \\
\textbf{Model} 
& Acc ($\uparrow$) & CR ($\downarrow$) 
& Acc ($\uparrow$) & CR ($\downarrow$) 
& Acc ($\uparrow$) & CR ($\downarrow$) 
& Acc ($\uparrow$) & CR ($\downarrow$) \\
\hline
\multicolumn{9}{c}{\textit{Supervised}} \\
\hline
Qwen3-8B 
& 90.1\% & 93.89\% 
& 65.8\% & 81.54\% 
& 52.6\% & 78.87\% 
& 71.8\% & 92.92\% \\

Qwen3-14B 
& 89.8\% & 91.92\% 
& 63.3\% & 71.29\% 
& 54.6\% & 81.90\% 
& 74.8\% & 87.12\% \\

Ministral-3-8B 
& 87.7\% & 87.80\% 
& 83.7\% & 91.16\% 
& 47.3\% & 71.78\% 
& 47.2\% & 87.16\% \\

Ministral-3-14B 
& 87.3\% & 95.86\% 
& 59.4\% & 79.77\% 
& 49.5\% & 73.81\% 
& 27.6\% & 96.25\% \\
\hline
\multicolumn{9}{c}{\textit{Consistent}} \\
\hline
Qwen3-8B 
& 88.2\% & 72.09\% 
& 43.1\% & 54.64\% 
& 44.7\% & 45.17\% 
& 70.7\% & 73.90\% \\

Qwen3-14B 
& 85.9\% & 64.40\% 
& 41.7\% & 45.57\% 
& 48.3\% & 53.83\% 
& 70.9\% & 39.60\% \\

Ministral-3-8B
& 75.8\% & 80.53\% 
& 23.3\% & 60.50\% 
& 42.4\% & 62.69\% 
& 34.2\% & 75.24\% \\

Ministral-3-14B
& 68.7\% & 77.40\% 
& 9.4\% & 34.42\% 
& 40.6\% & 59.97\% 
& 9.6\% & 95.90\% \\
\hline
\end{tabular}\\[0.1cm]
\end{table}

\section{Additional Experimental Results}
\label{app:additional_experiments}

\subsection{Early-Exit Signal Analysis}
\label{app:early-exit-signals}

\begin{figure*}[!t]
    \centering
    \includegraphics[width=1.0\textwidth]{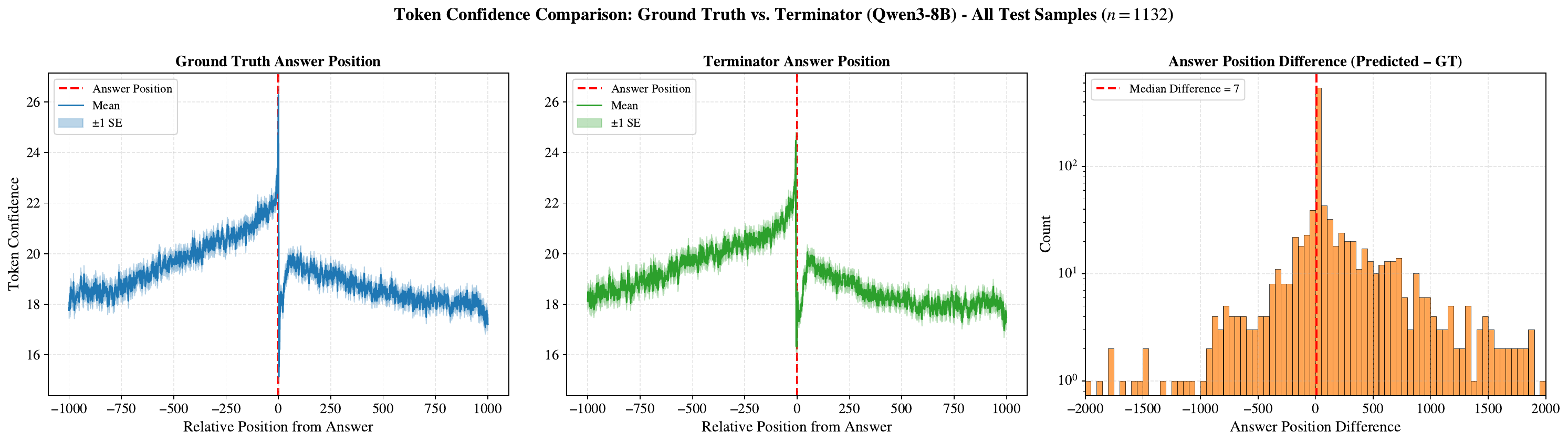}
    \caption{\textbf{\method Recovers Event-Locked Average Spiking.} The exit positions predicted by \method \textbf{(center)} recovers similar spiking behavior in the event-locked averaged Token-Confidence as the ground-truth answer positions \textbf{(left)}. The histogram of differences between the exit positions \textbf{(right)} shows that \method's predicted exit positions are close to the ground-truth. Note that the y-axis on the histogram is log-scaled.}
    \vspace{-16pt}
    \label{fig:timeseries-comparison}
\end{figure*}
\begin{figure*}[!t]
    \centering
    \includegraphics[width=1.0\textwidth]{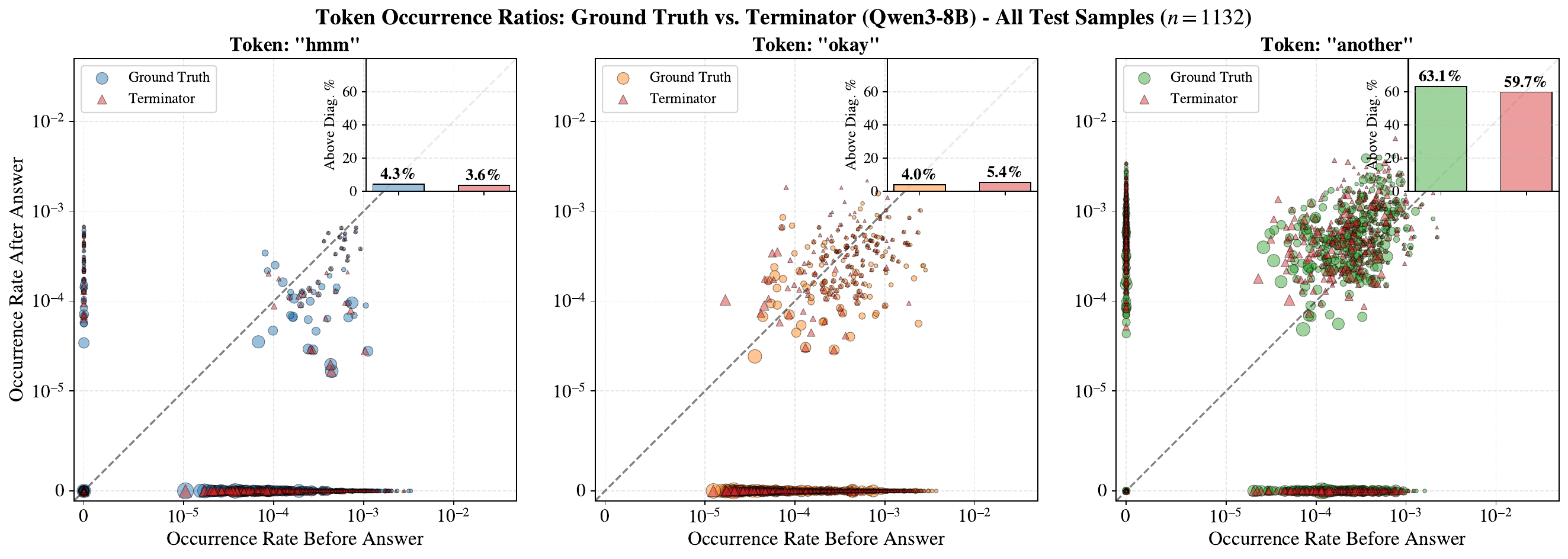}
    \caption{\textbf{\method Token Usage Biases.} The exit positions predicted by \method recover the same biases in the ``thinking token'' occurrence rates as the ground-truth answer positions. The inset axes on each panel show the percentage of dots that lie above the diagonal when the ground-truth and \method answer positions are used.}
    \vspace{-16pt}
    \label{fig:scatter-overlay}
\end{figure*}

\textbf{\method Recovers Early-Exit Signals.}
As discussed in \cref{sec:ablations}, \method's early-exit predictions can be used to recover similar early-exit signals as when the ground-truth earliest answer position is used. \cref{fig:timeseries-comparison,fig:scatter-overlay} establish this claim. Discussion on our findings are provided in \cref{sec:ablations}

\textbf{Event-Locked Averaging \tokconf.} 
The results shown in \cref{fig:aligned-timeseries-all,fig:token-scatter} motivate our approach by confirming that the first arrival of \ans is (1) marked by spiking behavior in the \tokconf, which is most easily seen in the event-locked average case, and (2) by a shift in the ``thinking token'' usage distribution. Focusing on the averaged plots in \cref{fig:aligned-timeseries-all}, we observe that the confidence of the LRM grows up to the point when \ans is first generated, where the confidence finally peaks. The confidence immediately drops after \ans is generated, which is intuitive given that the LRM immediately begins to doubt itself, often producing ``thinking tokens'' like \texttt{wait} or \texttt{but}, signaling uncertainty about the answer that was just generated. The LRM's confidence improves slightly as it continues to rethink the problem.

\begin{figure*}[!t]
    \centering
    \includegraphics[width=1.0\textwidth]{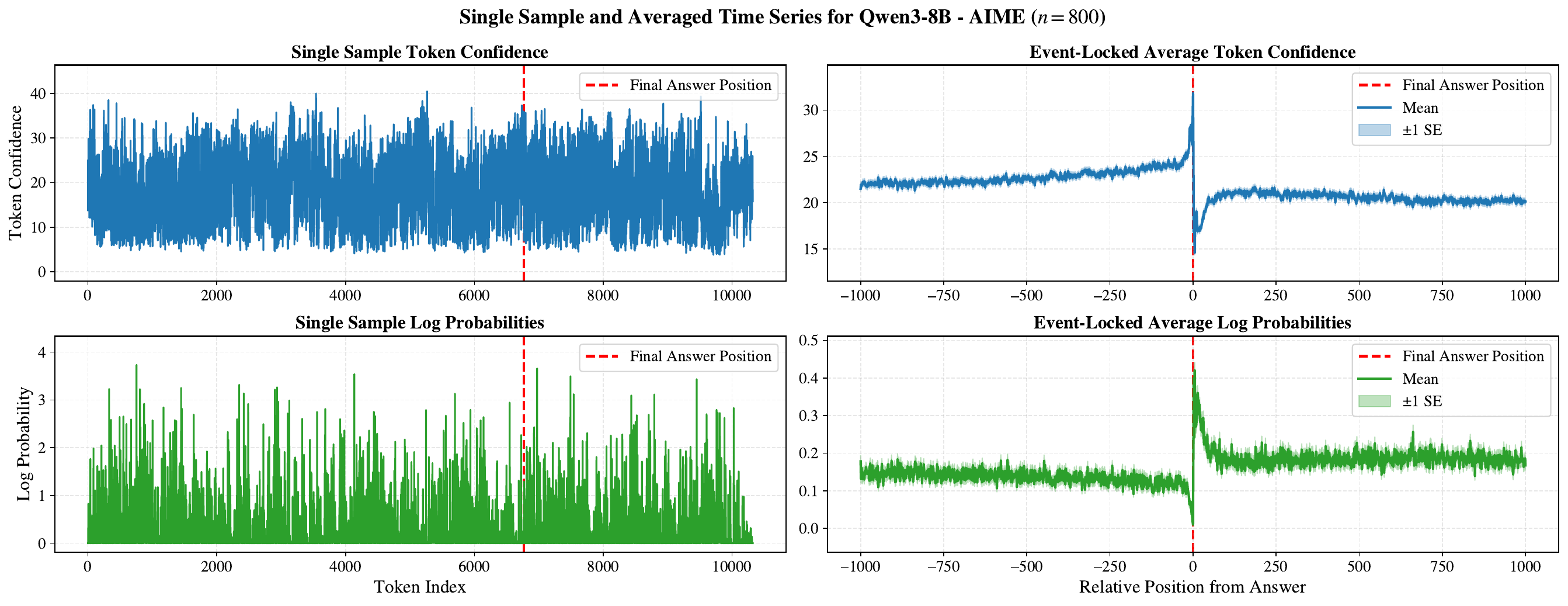}
    \caption{\textbf{Event-Locked Averaging of \tokconf for AIME (1983--2024).} A reproduction of \cref{fig:aligned-timeseries-all}, but only using \rtx[s] from 800 randomly selected AIME (1983--2024) problems.}
    \label{fig:aligned-timeseries-aime}
\end{figure*}

\begin{figure*}[!t]
    \centering
    \includegraphics[width=1.0\textwidth]{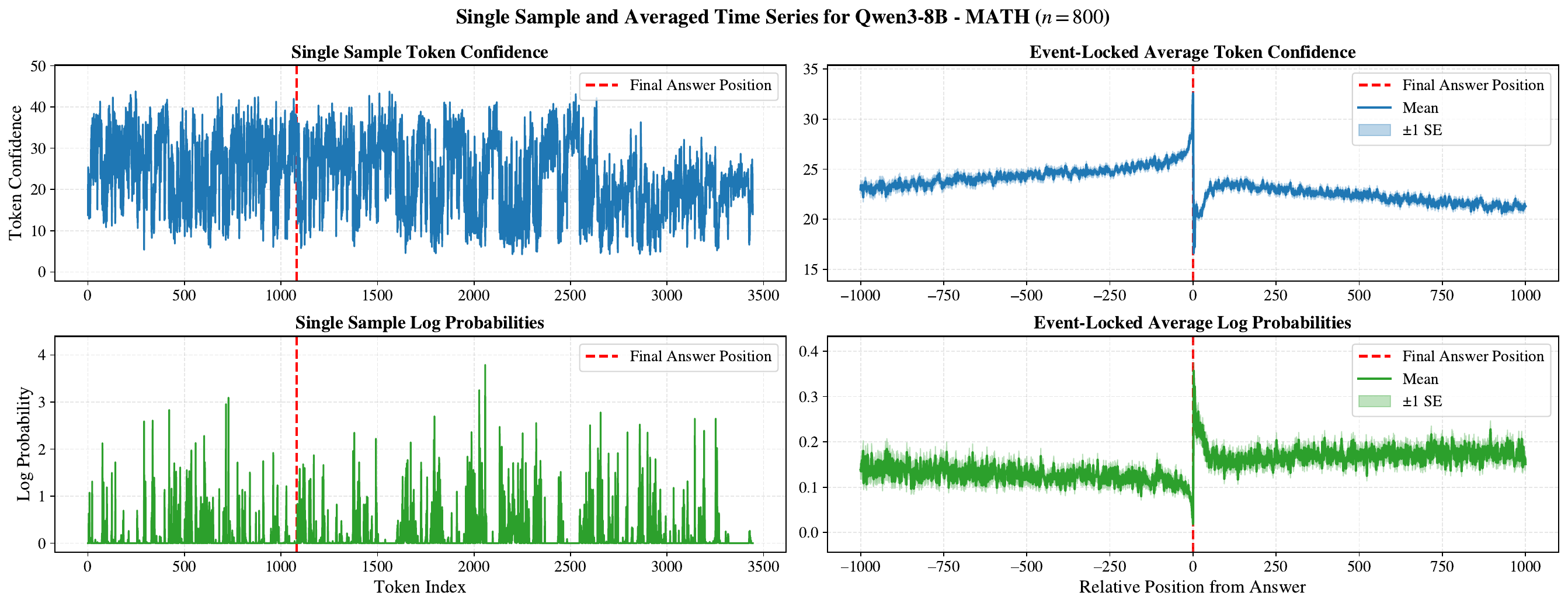}
    \caption{\textbf{Event-Locked Averaging of \tokconf for MATH.} A reproduction of \cref{fig:aligned-timeseries-all}, but only using \rtx[s] from 800 randomly selected MATH problems.}
    \label{fig:aligned-timeseries-math}
\end{figure*}

\begin{figure*}[!t]
    \centering
    \includegraphics[width=1.0\textwidth]{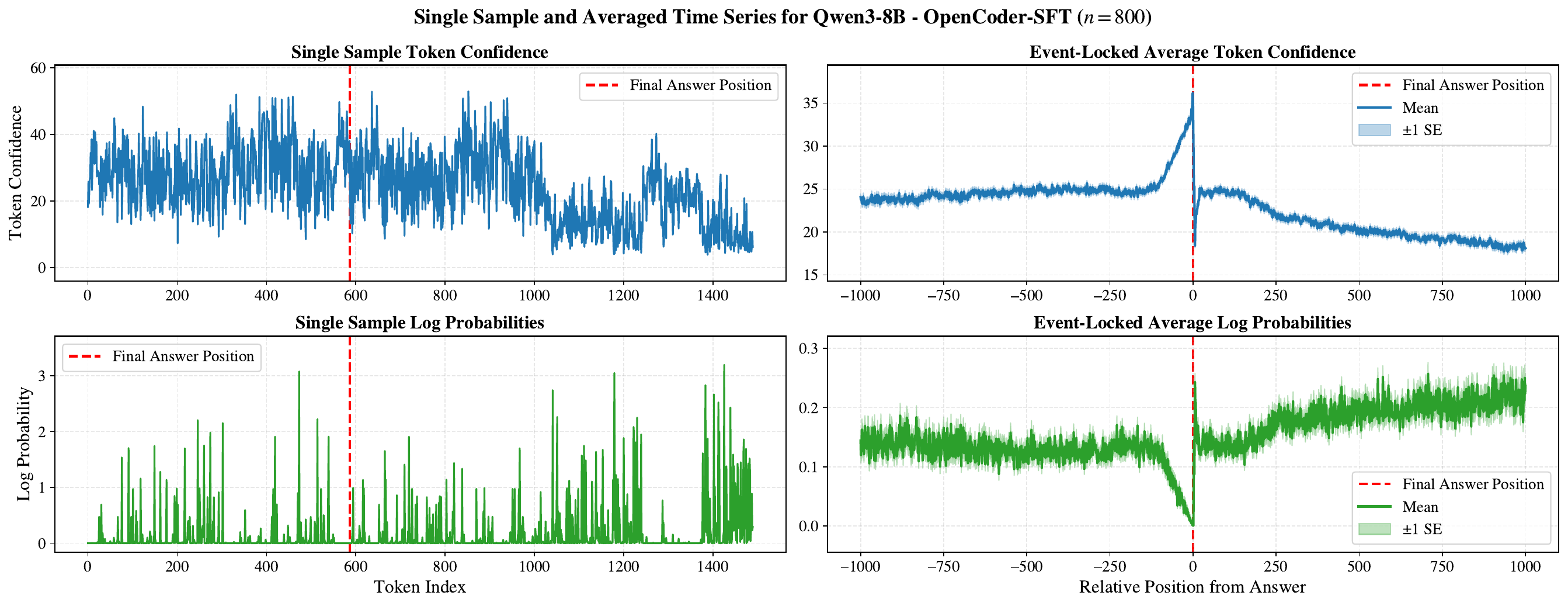}
    \caption{\textbf{Event-Locked Averaging of \tokconf for OpenCoder-SFT.} A reproduction of \cref{fig:aligned-timeseries-all}, but only using \rtx[s] from 800 randomly selected OpenCoder-SFT problems.}
    \label{fig:aligned-timeseries-opencoder}
\end{figure*}

\begin{figure*}[!t]
    \centering
    \includegraphics[width=1.0\textwidth]{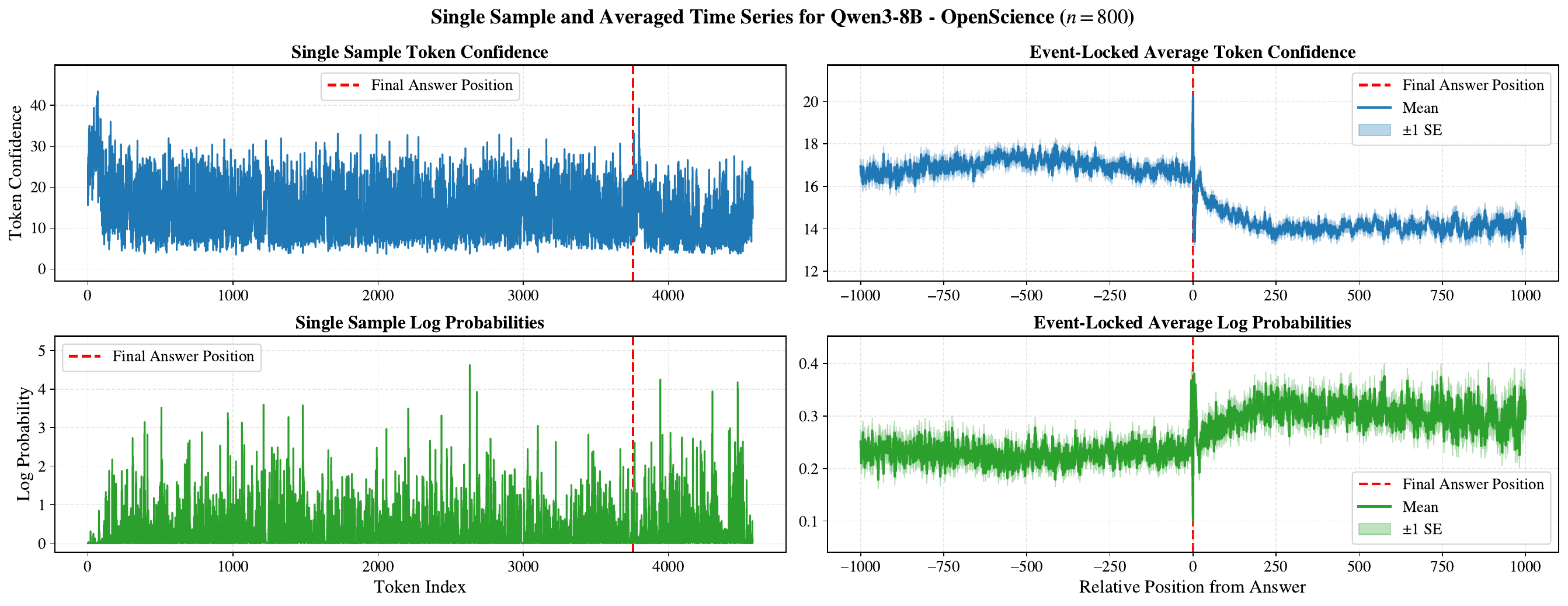}
    \caption{\textbf{Event-Locked Averaging of \tokconf for OpenScience.} A reproduction of \cref{fig:aligned-timeseries-all}, but only using \rtx[s] from 800 randomly selected OpenScience problems.}
    \label{fig:aligned-timeseries-openscience}
\end{figure*}

\cref{fig:aligned-timeseries-aime,fig:aligned-timeseries-math,fig:aligned-timeseries-opencoder,fig:aligned-timeseries-openscience} show the \tokconf and \logprob[s] for the single-sample and event-locked averaging case, similar to what's shown in \cref{fig:aligned-timeseries-all}, separately for each data source. While the exact contours of these two signal types vary for different data sources, the same idea applies to all: the LRM's \tokconf has a sharp spike at the position of the first occurrence of \ans, followed by a sharp decrease. In all cases, the \tokconf then has a quick recovery before plateauing or decaying.

\begin{figure*}[!t]
    \centering
    \includegraphics[width=1.0\textwidth]{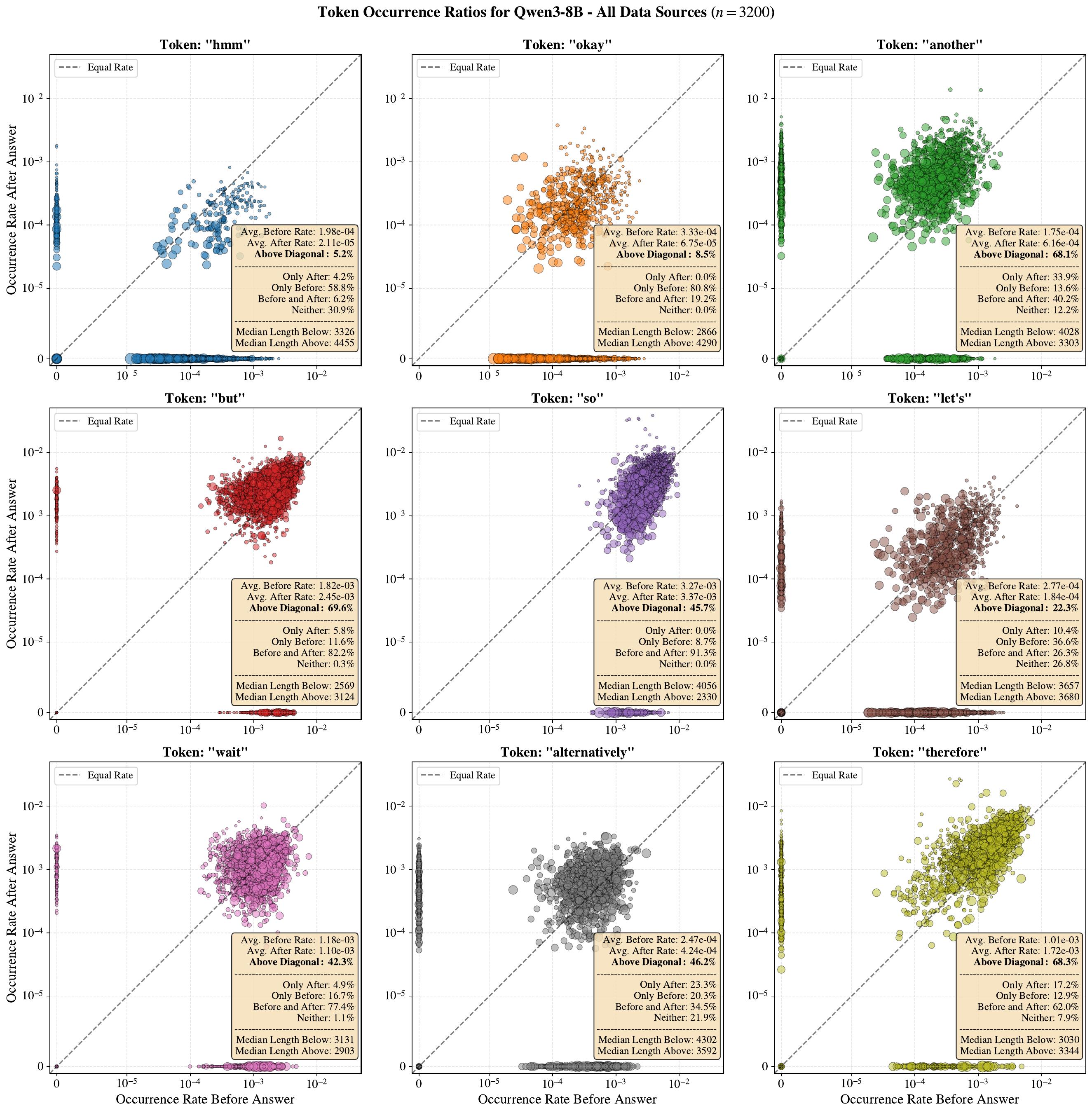}
    \caption{\textbf{Token Usage Frequency Shift.} An extension of the results shown in \cref{fig:token-scatter}, highlighting additional ``thinking tokens.'' While most ``thinking tokens'' shown here have some bias, often occurring before and after the first occurrence of the final answer as indicated by the \textbf{Above Diagonal} statistic, some tokens, like ``\texttt{so},'' are close to an equal rate on average.}
    \label{fig:token-scatter-all}
\end{figure*}

\begin{figure*}[!t]
    \centering
    \includegraphics[width=1.0\textwidth]{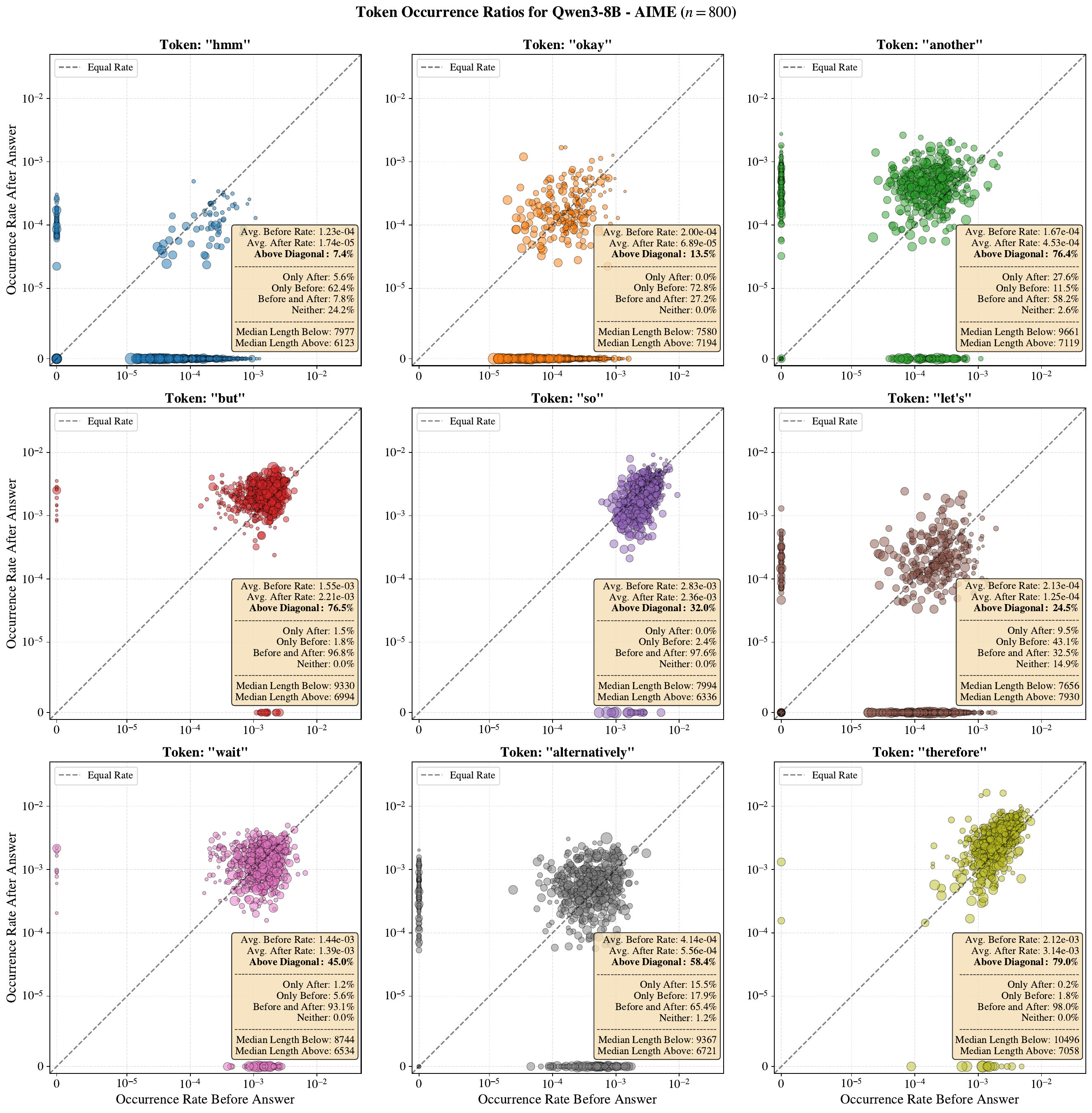}
    \caption{\textbf{Token Usage Frequency Shift for AIME (1983--2024.} A reproduction of \cref{fig:token-scatter-all}, but only using \rtx[s] from 800 randomly selected AIME (1983--2024) problems.}
    \label{fig:token-scatter-aime}
\end{figure*}

\begin{figure*}[!t]
    \centering
    \includegraphics[width=1.0\textwidth]{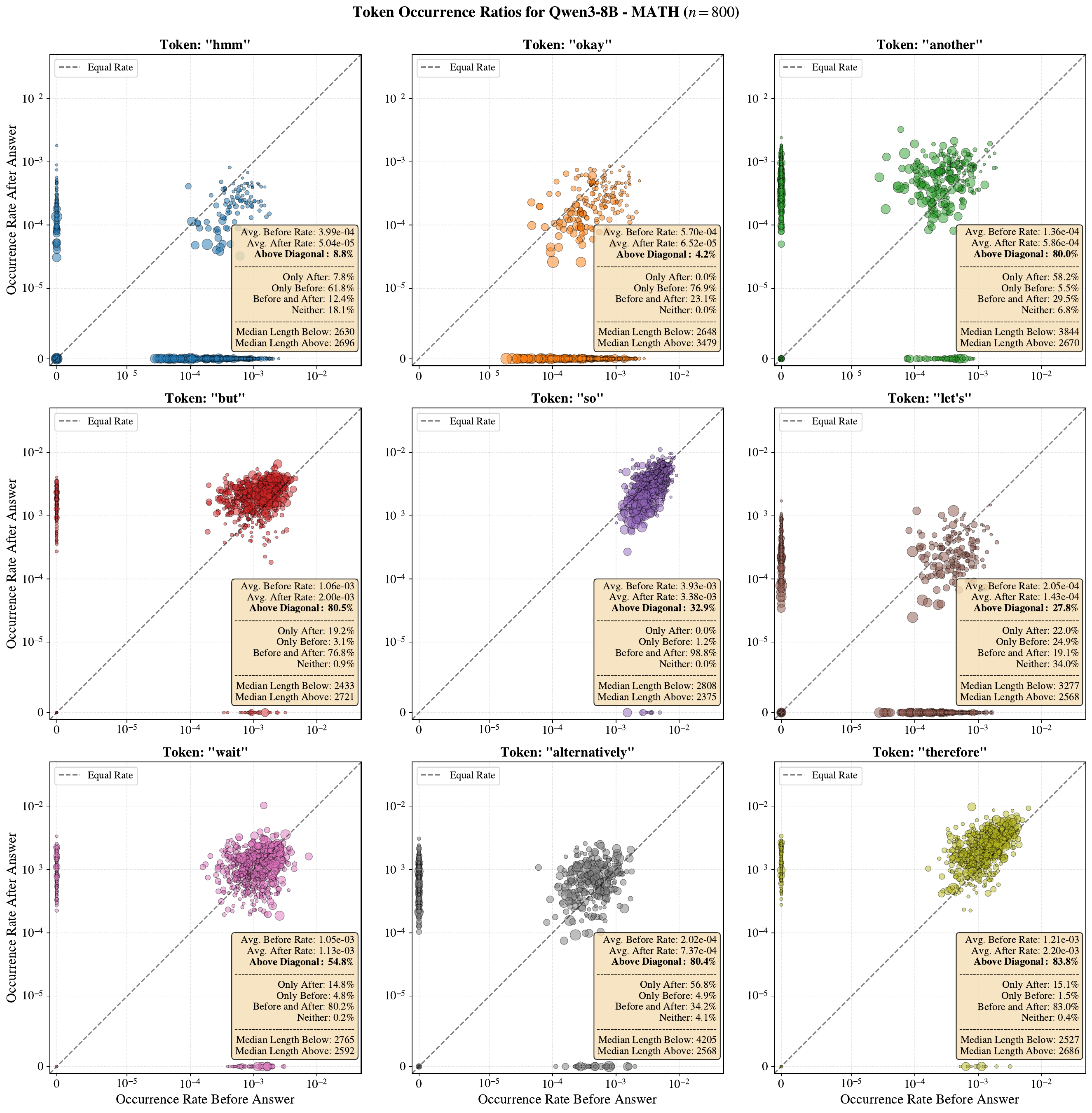}
    \caption{\textbf{Token Usage Frequency Shift for MATH.} A reproduction of \cref{fig:token-scatter-all}, but only using \rtx[s] from 800 randomly selected MATH problems.}
    \label{fig:token-scatter-math}
\end{figure*}

\begin{figure*}[!t]
    \centering
    \includegraphics[width=1.0\textwidth]{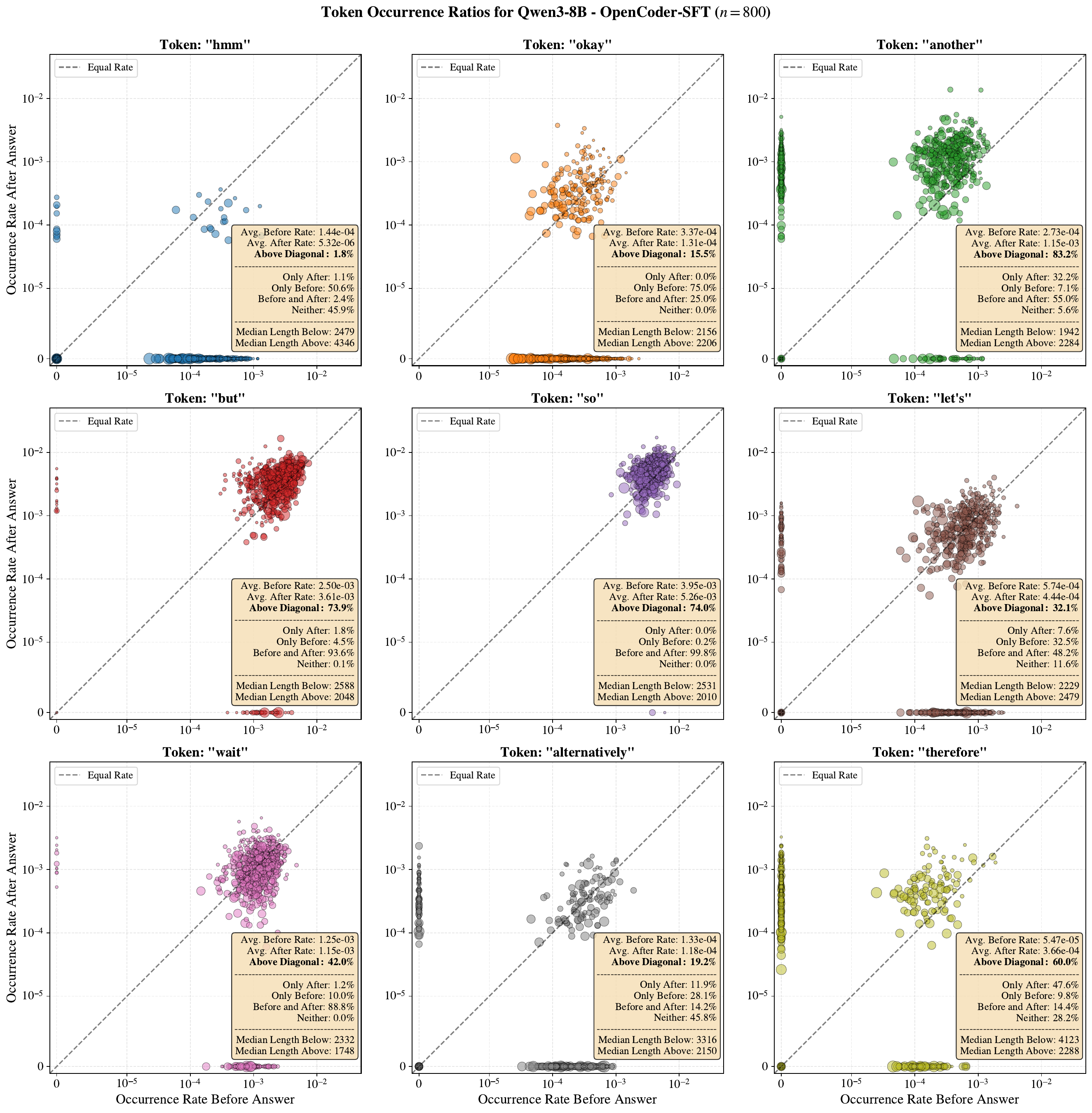}
    \caption{\textbf{Token Usage Frequency Shift for OpenCoder-SFT.} A reproduction of \cref{fig:token-scatter-all}, but only using \rtx[s] from 800 randomly selected OpenCoder-SFT problems.}
    \label{fig:token-scatter-opencoder}
\end{figure*}

\begin{figure*}[!t]
    \centering
    \includegraphics[width=1.0\textwidth]{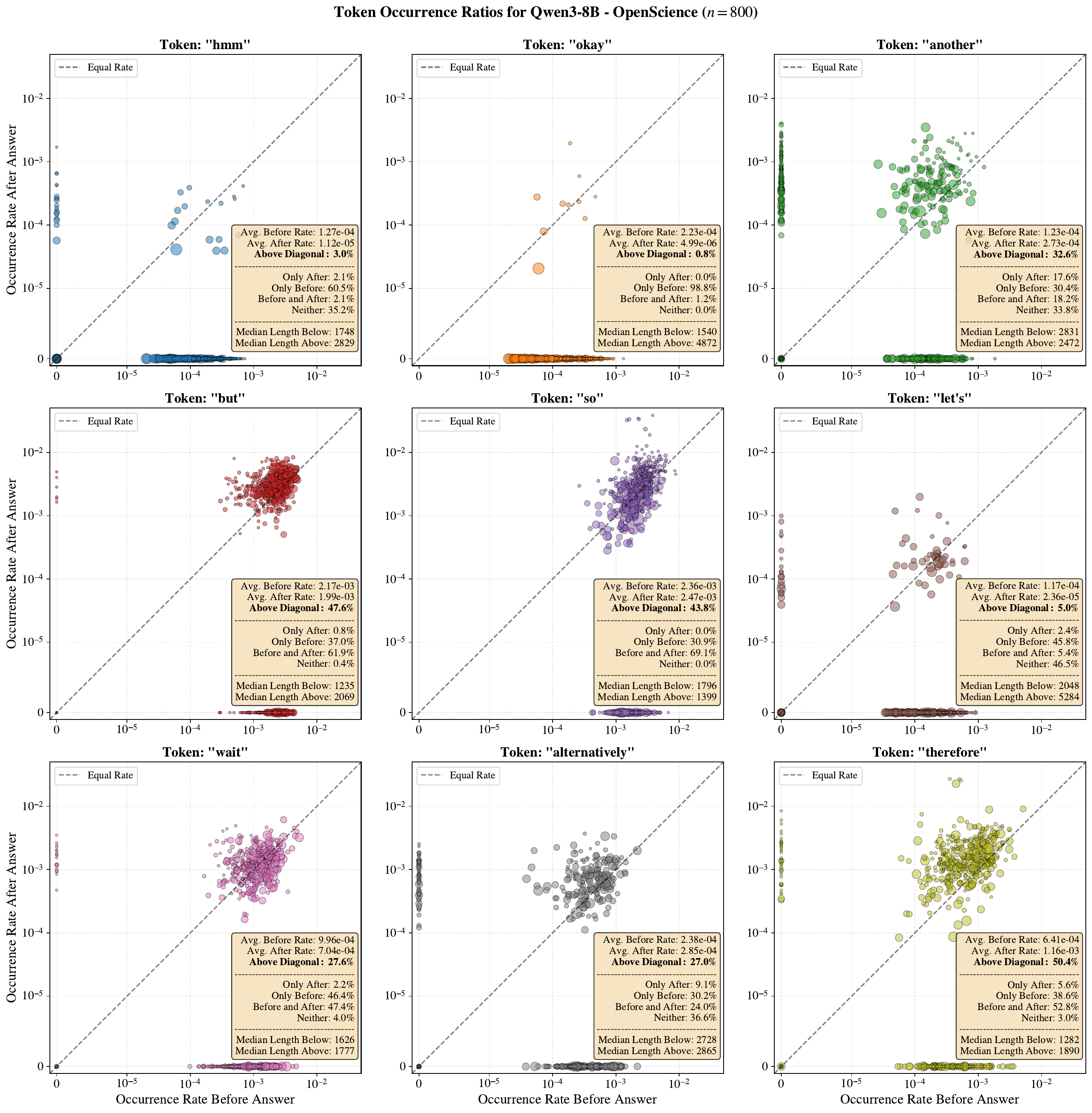}
    \caption{\textbf{Token Usage Frequency Shift for OpenScience.} A reproduction of \cref{fig:token-scatter-all}, but only using \rtx[s] from 800 randomly selected OpenScience problems.}
    \label{fig:token-scatter-openscience}
\end{figure*}

\textbf{Token Usage Frequency Shift. }
Beyond providing motivation for our method, the results of \cref{fig:token-scatter} offer some interesting insights on the usage of ``thinking tokens.'' These plots show the strong usage bias that can occur with respect to the first occurrence of \ans. For example, 63.9\% and 91.5\% of \rtx[s] contain the tokens \texttt{hmm} and \texttt{okay} more often before \ans than after it, respectively. Other tokens, like \texttt{another} are more frequently used after. Moreover, \cref{fig:token-scatter-all,fig:token-scatter-math,fig:token-scatter-aime,fig:token-scatter-opencoder,fig:token-scatter-openscience} show that the occurrence rates can differ drastically between data sources. For example, the token \texttt{alternatively} has an above-diagonal rate of 80.4\% for MATH, but only 19.2\% for OpenCoder-SFT.

\begin{figure*}[!t]
    \centering
    \includegraphics[width=1.0\textwidth]{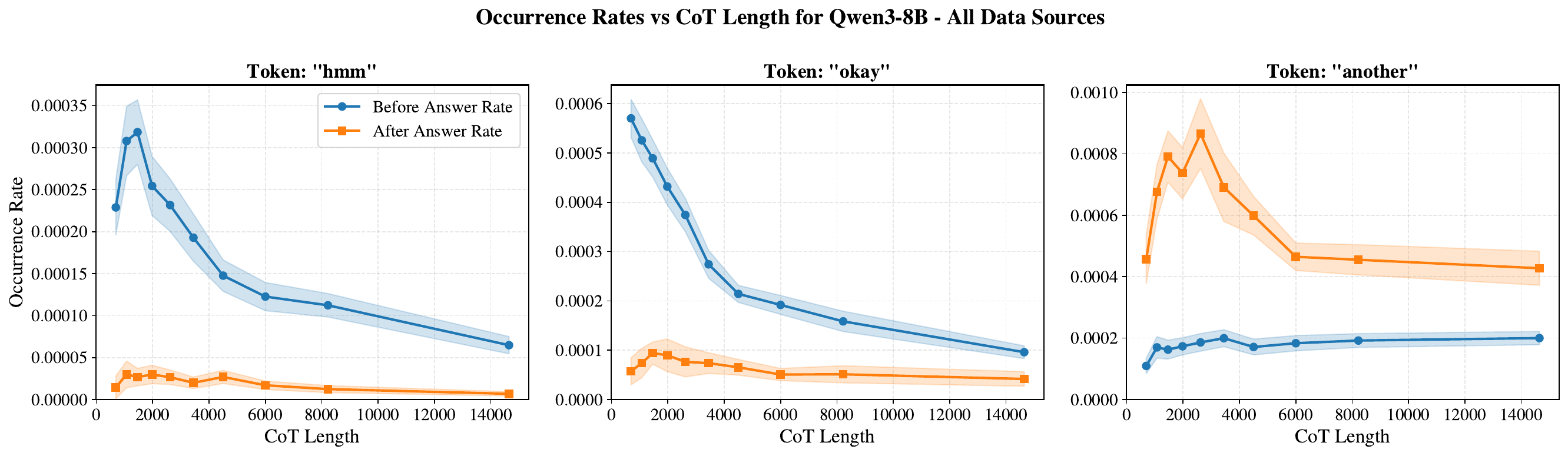}
    \caption{\textbf{Token Occurrence Rate vs \rtx Length.} For some tokens, such as these three ``thinking tokens,'' occurrence rates decrease rapidly as \rtx length increases. Interestingly, the side of the answer (either ``before'' or ``after'') with the highest rate is the one that decays the most (the ``before'' rate for \texttt{hmm} and \texttt{okay} and the ``after'' rate for \texttt{another}), while the side with the lower rate sees only a slight decrease or increase. For each plot, the lengths are placed into ten bins with percentile-based bin edges. In other words, each bin contains approximately 10\% of the samples. Shaded regions indicate the 95\% confidence interval.}
    \label{fig:rate-vs-length}
\end{figure*}

\cref{fig:token-scatter} also indicates the length of each \rtx by its dot size. Upon closer inspection, it appears that there is some correlation between the dot size and the occurrence rates. We plot the before and after occurrence rates with respect to \rtx length for these three tokens in \cref{fig:rate-vs-length}. Interestingly, shorter \rtx[s] do in fact correlate with higher token occurrences for these three ``thinking tokens.''

\cref{fig:token-scatter-all} shows an expanded view of \cref{fig:token-scatter}, including six additional ``thinking tokens.'' \cref{fig:token-scatter-aime,fig:token-scatter-math,fig:token-scatter-opencoder,fig:token-scatter-openscience} reproduces this expanded plot, but with the data sources separated. Interestingly, the occurrence rates can vary substantially across different data sources. For example, for the token \texttt{alternatively}, only 19.2\% of points lie above the diagonal (only 19.2\% of \rtx[s] have the token \texttt{alternatively} occurring more frequently after the answer than before), and 45.8\% (nearly half!) of the points are at the origin (the token \texttt{alternatively} never occurs in 45.8\% of \rtx[s]) for OpenCoder-SFT. However, for MATH, 80.4\% of points lie above the diagonal and only 4.1\% lie at the origin for the same token. Other tokens, like \texttt{therefore}, are strongly biased toward occurring after the answer.

\cref{fig:rate-vs-length} shows that the average occurrence rate across \rtx[s] from all data sources changes depending on how long the \rtx is. We normalize by the number of tokens occurring before and after the first occurrence of \ans for the before and after rates, respectively, so these plots suggest that the LRM uses \texttt{hmm}, \texttt{okay}, and \texttt{another} more frequently for shorter \rtx[s] than longer ones.

\begin{table}[t]
\centering
\caption{\textbf{Cross-Domain vs.\ Single-Domain Training.} Comparison of accuracy and compression ratio when \method is trained on a single-domain (based on \cref{fig:heatmaps}) and all domains (based on \cref{tab:performance}). Cross-domain training (the proposed, unified model) and single-domain training are both valid approaches; cross-domain achieves a 1\% increase in accuracy over single-domain, but at a 1.5\% worse compression rate.}
\label{tab:cross_vs_single}
\resizebox{0.9\textwidth}{!}{%
\begin{tabular}{@{}lcccccccccc@{}}
\toprule
 & \multicolumn{4}{c}{\textbf{Math}} & \multicolumn{2}{c}{\textbf{Coding}} & \multicolumn{2}{c}{\textbf{Science}} & \multicolumn{2}{c}{\textbf{}} \\
\cmidrule(l){2-5} \cmidrule(l){6-7} \cmidrule(l){8-9}
\textbf{Method}
  & \multicolumn{2}{c}{\textbf{MATH-500}}
  & \multicolumn{2}{c}{\textbf{AIME25}}
  & \multicolumn{2}{c}{\textbf{HumanEval}}
  & \multicolumn{2}{c}{\textbf{GPQA}}
  & \multicolumn{2}{c}{\textbf{Overall}} \\
 & {Acc$\uparrow$} & {CR$\downarrow$}
 & {Acc$\uparrow$} & {CR$\downarrow$}
 & {Acc$\uparrow$} & {CR$\downarrow$}
 & {Acc$\uparrow$} & {CR$\downarrow$}
 & {Acc$\uparrow$} & {CR$\downarrow$} \\
\midrule
Single-domain training (\cref{fig:heatmaps}) 
  & \textbf{91\%} & 47\%
  & 66\% & \textbf{70\%}
  & \textbf{96\%} & \textbf{67\%}
  & 55\% & \textbf{82\%}
  & 77\% & \textbf{66.5\%} \\

Cross-domain training (\cref{tab:performance})
  & \textbf{91\%} & \textbf{45\%}
  & \textbf{69\%} & 71\%
  & \textbf{96\%} & 70\%
  & \textbf{56\%} & 86\%
  & \textbf{78\%} & 68\% \\
\bottomrule
\end{tabular}%
}
\end{table}
\begin{figure*}[!t]
    \centering
    \includegraphics[width=0.7\textwidth]{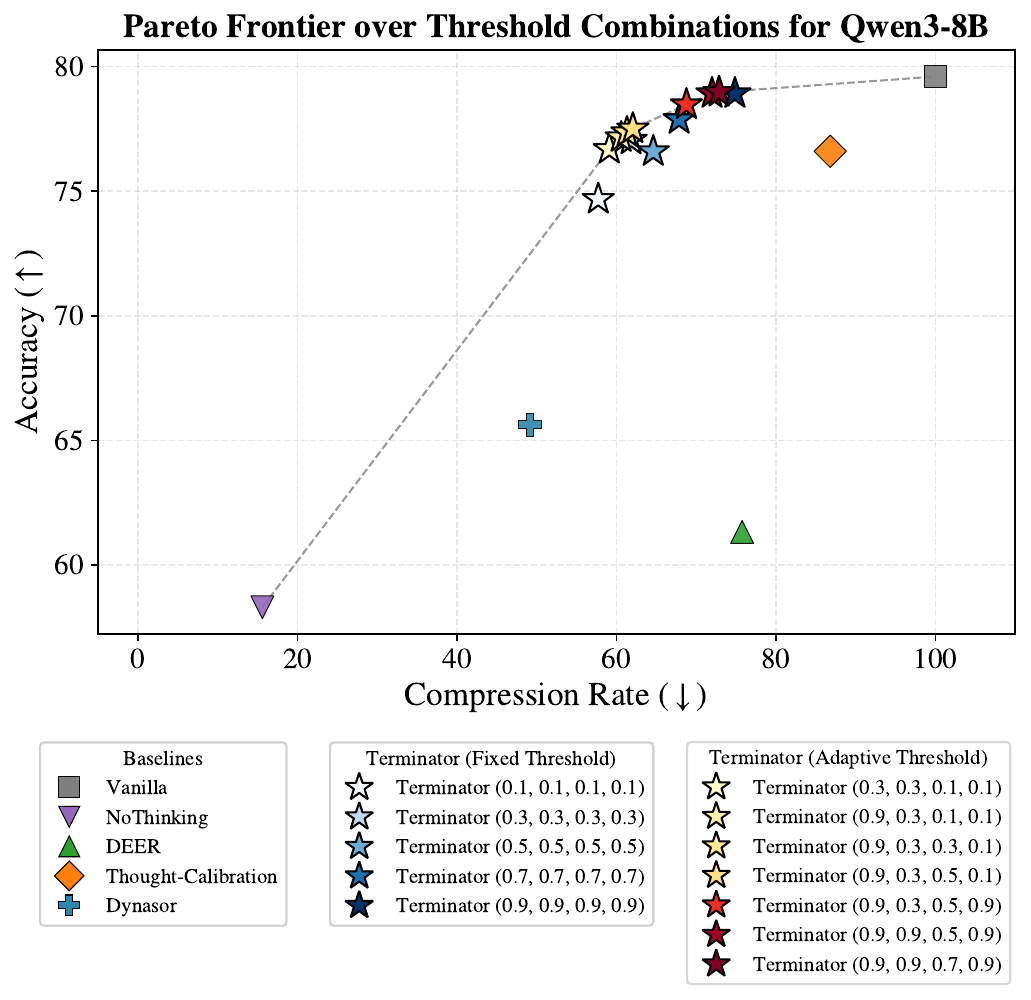}
    \caption{\textbf{Adaptive Thresholding for \method.} We plot \method at various thresholds (blue stars, fixed for all domains) and compare with a domain-adaptive threshold selection (red stars). The tuples correspond to the thresholds chosen for (MATH-500, AIME 25, HumanEval, GPQA) accordingly. For example, ``(0.9, 0.3, 0.1, 0.1)'' corresponds to choosing thresholds 0.9, 0.3, 0.1, 0.1 for MATH-500, AIME 25, HumanEval, and GPQA, respectively. We enumerate all possible threshold combinations and retain those that define the Pareto frontier. Domain adaptation provides incremental gains but requires domain knowledge at inference time.}
    \label{fig:pareto-frontier-thresh-combos}
\end{figure*}

\subsection{Adaptive Strategies for \method}
In principle, adaptive early-exiting strategies offer a promising way to tailor inference to the target domain \cite{wu2026efficiency}. There are two design choices for \method that are amenable to inference-time adaptation: (1) domain-based model selection, and (2) threshold selection. Domain-based model selection is concerned with training four separate \method models on each of the four training data sets separately and choosing the appropriate one at inference time. This contrasts with our proposed approach, which trains \method on all domains simultaneously. Similarly, threshold selection considers adaptivity at inference time by setting a separate threshold for each domain, rather than using a single threshold for all domains. We seek to understand how much performance \method can gain, if any, by enabling domain adaptation on these two design choices.

\begin{figure}[!t]
    \centering
    \includegraphics[width=0.8\columnwidth]{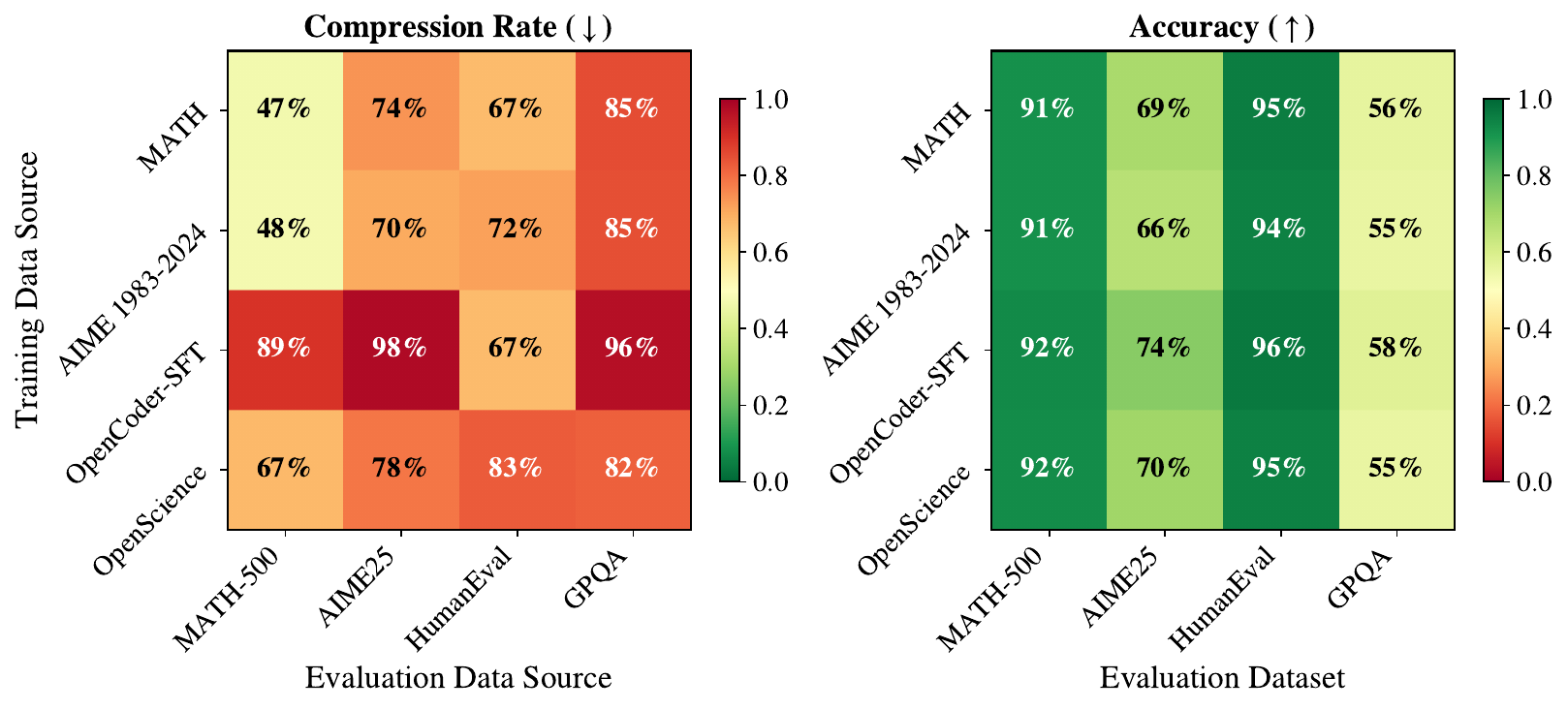}
    \caption{\textbf{OOD Performance of \method.} The best trade-off between accuracy and compression rate is achieved when the evaluation set is in-distribution with the training dataset. Here the out-of-distribution performance of \method with respect to the compression rate \textbf{(left)} and the accuracy \textbf{(right)} for Qwen3-8B is shown. Training datasets are listed along the row axis, and the evaluation sets are listed across the column axis. For example, training \method on MATH and evaluating on HumanEval yields a compression rate of $67\%$ and an accuracy of $95\%$. Every training dataset has an in-domain evaluation dataset, \ie MATH $\rightarrow$ MATH-500, AIME 1983--2024 $\rightarrow$ AIME25, OpenCoder-SFT $\rightarrow$ HumanEval, and OpenScience $\rightarrow$ GPQA.}
    \vspace{-14pt}
    \label{fig:heatmaps}
\end{figure}

\cref{tab:cross_vs_single} shows the outcome of comparing domain-based model selection with our proposed approach. The domain-based model selection results (``single-domain training'') come from the diagonals of \cref{fig:heatmaps}, and the results of our proposed method (``cross-domain training'') for \method come from \cref{tab:performance}. Overall, both approaches yield good performance, without a clear winner: cross-domain training achieves 1\% higher accuracy than single-domain training, but at a 1.5\% worse compression rate.

\cref{fig:pareto-frontier-thresh-combos} shows the best possible performance with a domain-adaptive thresholding approach for \method for five possible thresholds: $0.1, 0.3, 0.5, 0.7, and 0.9$. In the figure, the tuples correspond to the thresholds chosen for (MATH-500, AIME 25, HumanEval, GPQA) accordingly. For example, ``($0.9, 0.3, 0.1, 0.1$)'' corresponds to choosing thresholds $0.9, 0.3, 0.1, 0.1$ for MATH-500, AIME 25, HumanEval, and GPQA, respectively. Five thresholds across four datasets yield $5^4 = 625$ possible combinations, so we show only those that define the Pareto frontier. We use a threshold of $0.7$ (represented with ``($0.7, 0.7, 0.7, 0.7$)'' in \cref{fig:pareto-frontier-thresh-combos}) for \method. While domain-adaptive thresholding (red stars) generally lies above fixed thresholding (blue stars), the improvement in performance is marginal.

However, we note that the results for these domain-adaptive are an upper bound to the performance that can be realized in practice, since perfect knowledge of which domain a given problem belongs to is assumed. In practice, an inference-time strategy is required to appropriately choose the model or threshold, which may introduce error and additional computational overhead. Furthermore, the domain-based model selection approach requires four separately trained models rather than a single unified model, thereby incurring additional compute and memory overhead for model routing.

\subsection{\method Prediction Analysis}

\begin{figure*}[!t]
    \centering
    \includegraphics[width=1.0\textwidth]{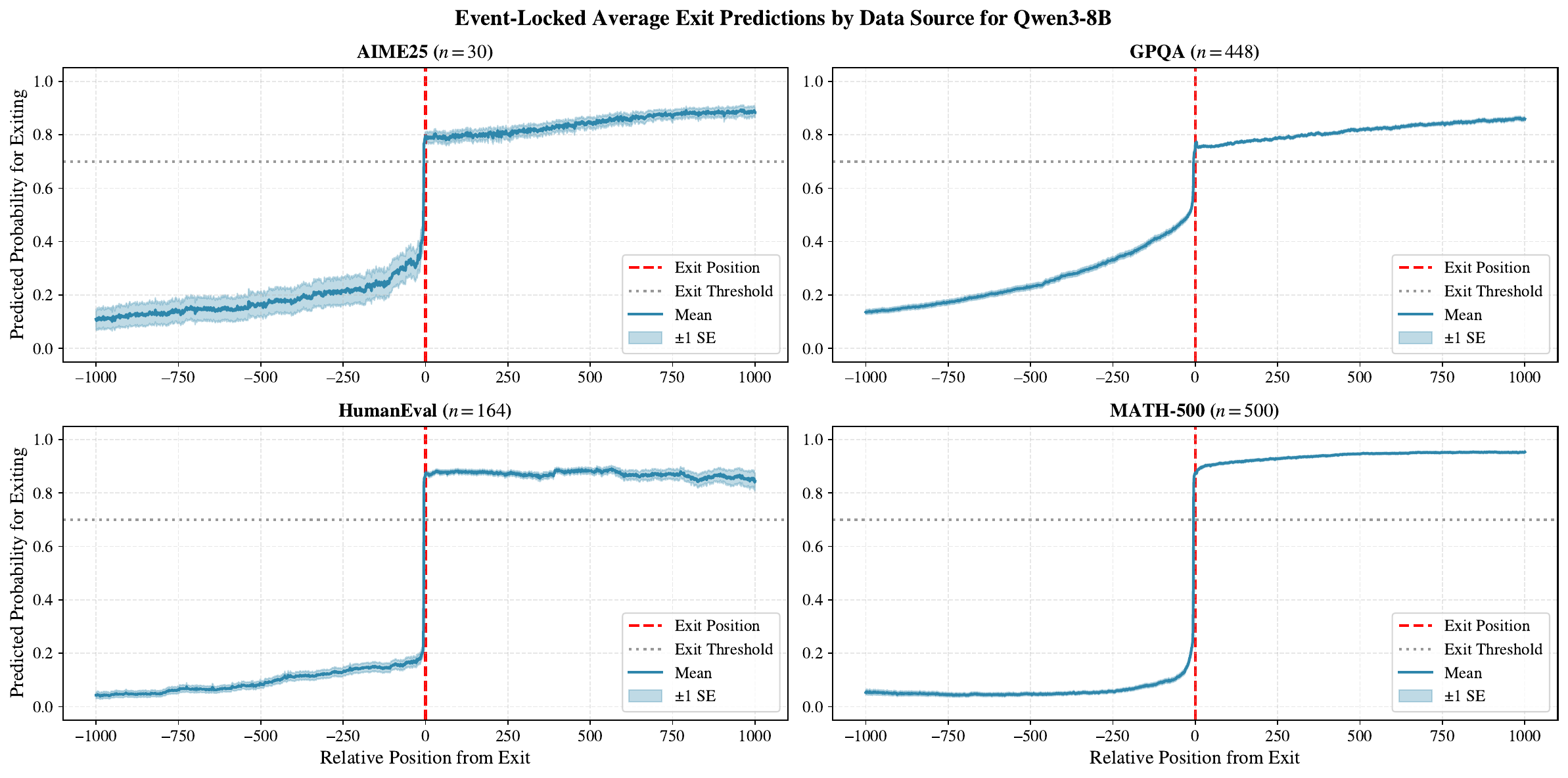}
    \caption{\textbf{Predicted Probabilities Event-Locked Averaging.} 
    The dashed vertical line shows where \method terminates the \rtx with a sliding window of 10 and an exit threshold of 0.7, as indicated by the horizontal dotted line. We show the average predicted probability stream across all test problems from MATH-500, AIME25, HumanEval, and GPQA. \cref{fig:predictions-aime,fig:predictions-math,fig:predictions-humaneval,fig:predictions-gpqa} show predictions streams from individual, randomly drawn samples from each data source.}
    \label{fig:aligned-predictions-all-app}
\end{figure*}

\textbf{\method Predictions During Reasoning. } \cref{fig:aligned-predictions-all-app} shows the event-locked average of the predicted probabilities from \method for each data source, and \cref{fig:predictions-aime}, \cref{fig:predictions-math}, \cref{fig:predictions-humaneval}, and \cref{fig:predictions-gpqa} show the predicted probabilities from four randomly chosen examples for AIME25, MATH-500, HumanEval, and GPQA, respectively. The event-locked average and the individual examples from MATH-500 and HumanEval show sharp transitions in predicted confidence at the exiting threshold, with good separation (dotted gray line). However, AIME25 and GPQA examples do not show such a sharp transition, suggesting that it is challenging for \method to identify a good exit position for very hard tasks.

\begin{figure}[!t]
    \centering
    \includegraphics[width=1.0\textwidth]{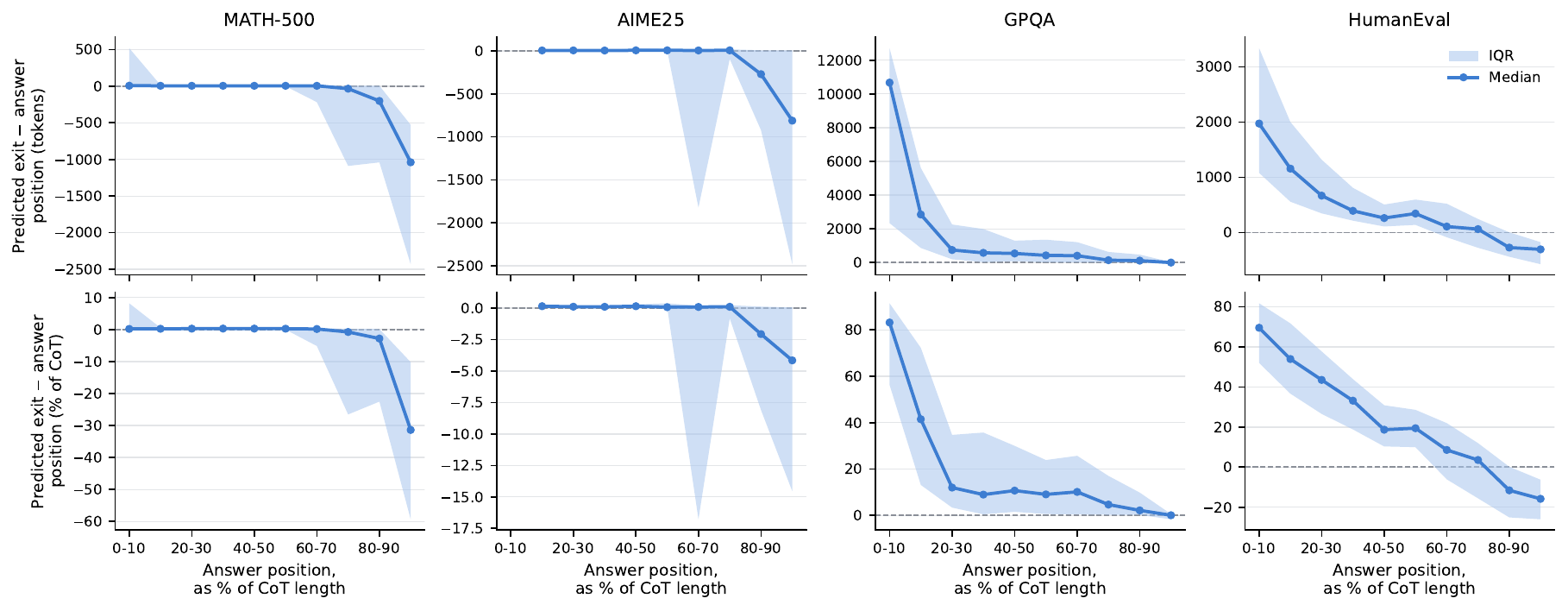}
    \caption{\textbf{Exit Timing Relative to Answer Arrival.} Each Qwen3-8B \rtx is binned by where the pipeline's answer position falls in the \rtx. We plot the median and IQR of the signed difference between \method's predicted exit position and the answer position, in tokens \textbf{(top)} and as a percentage of \rtx length \textbf{(bottom)}. Negative values indicate that \method exits earlier.}
    \label{fig:exit-timing}
\end{figure}

\textbf{Exit Timing Relative to Answer Arrival. }
On harder problems, reasoning after the first arrival of the answer may be needed for self-correction rather than being wasteful. This failure mode is absent from our training labels by construction: the hindsight-optimal length is the earliest arrival at the \emph{eventual final} answer, so any answer-changing self-correction falls before it. The LRM may also mention an answer-like value before it has completed its reasoning. Such superficial answer mentions do occur, but they primarily precede the position the pipeline selects. Isolating their causal effect would require detecting every candidate answer in a \rtx, and unlike the final answer, these appear without a reliable logical anchor, so the measurement is not directly tractable. In both cases, the open question is whether the trained probe exits early at inference. We bin each \rtx by where the pipeline's answer falls (an answer within the first 10\% of tokens goes in the 0--10\% bin) and plot the median and IQR of the signed difference between \method's predicted exit and the pipeline's position in \cref{fig:exit-timing}. Negative values mean that \method exits earlier. Bin index acts as a difficulty proxy: a late-arriving answer means most of the \rtx was spent deriving rather than verifying.

On MATH-500, the median \method exit falls 6--7 tokens after the pipeline's position whenever that position lies within the first 70\% of tokens, which covers 93\% of \rtx[s]; this matches the minimum delay of 6 tokens imposed by our strict majority vote (\cref{app:inference-details}). Premature exits appear only once the answer arrives after 70\% of the \rtx; those three bins hold only 7\% of \rtx[s], so the behavior is rare. The trend is dataset-specific. On GPQA, the harder benchmark, the median is positive in every bin, so \method does not exit early at the median. It exits later than the pipeline's position throughout, by a wide margin when the answer arrives early, but converges to a median of 1 token for the 1,251 \rtx[s] whose answer appears in the last 10\%. AIME25 resembles the shape of MATH-500, although its median turns negative only after 80\% of the \rtx, and those two bins hold 34\% of AIME25 \rtx[s]. HumanEval resembles the shape of GPQA over the first 80\% of the \rtx, but unlike GPQA, its median turns negative in the last two bins (10\% of \rtx[s]).

\begin{figure*}[!t]
    \centering
    \includegraphics[width=1.0\textwidth]{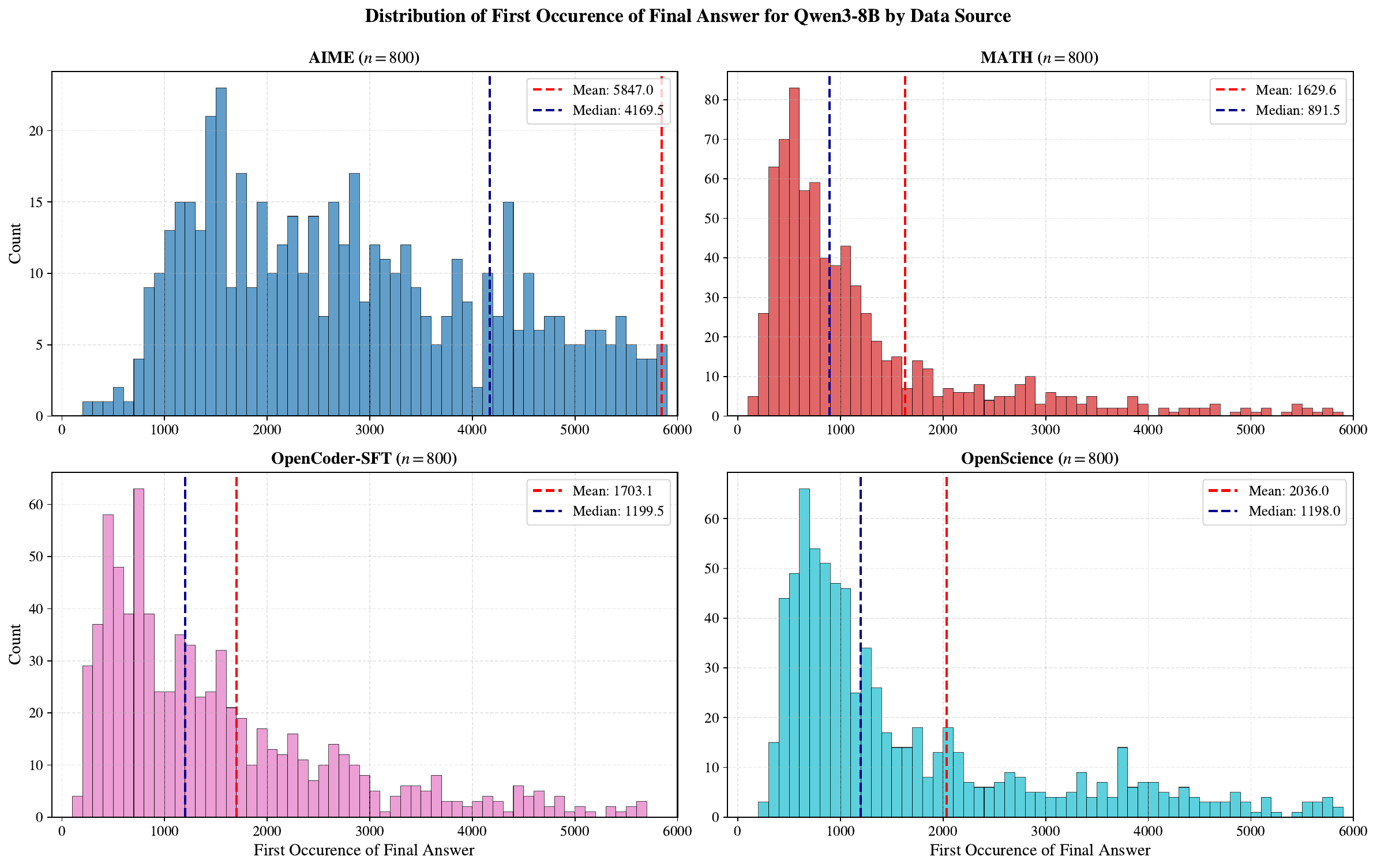}
    \caption{\textbf{First Answer Occurrence Histogram.} A histogram of the first occurrence of the final answer for each data source used in our training dataset is shown.}
    \label{fig:histogram-train-data}
\end{figure*}

\begin{figure*}[!t]
    \centering
    \includegraphics[width=1.0\textwidth]{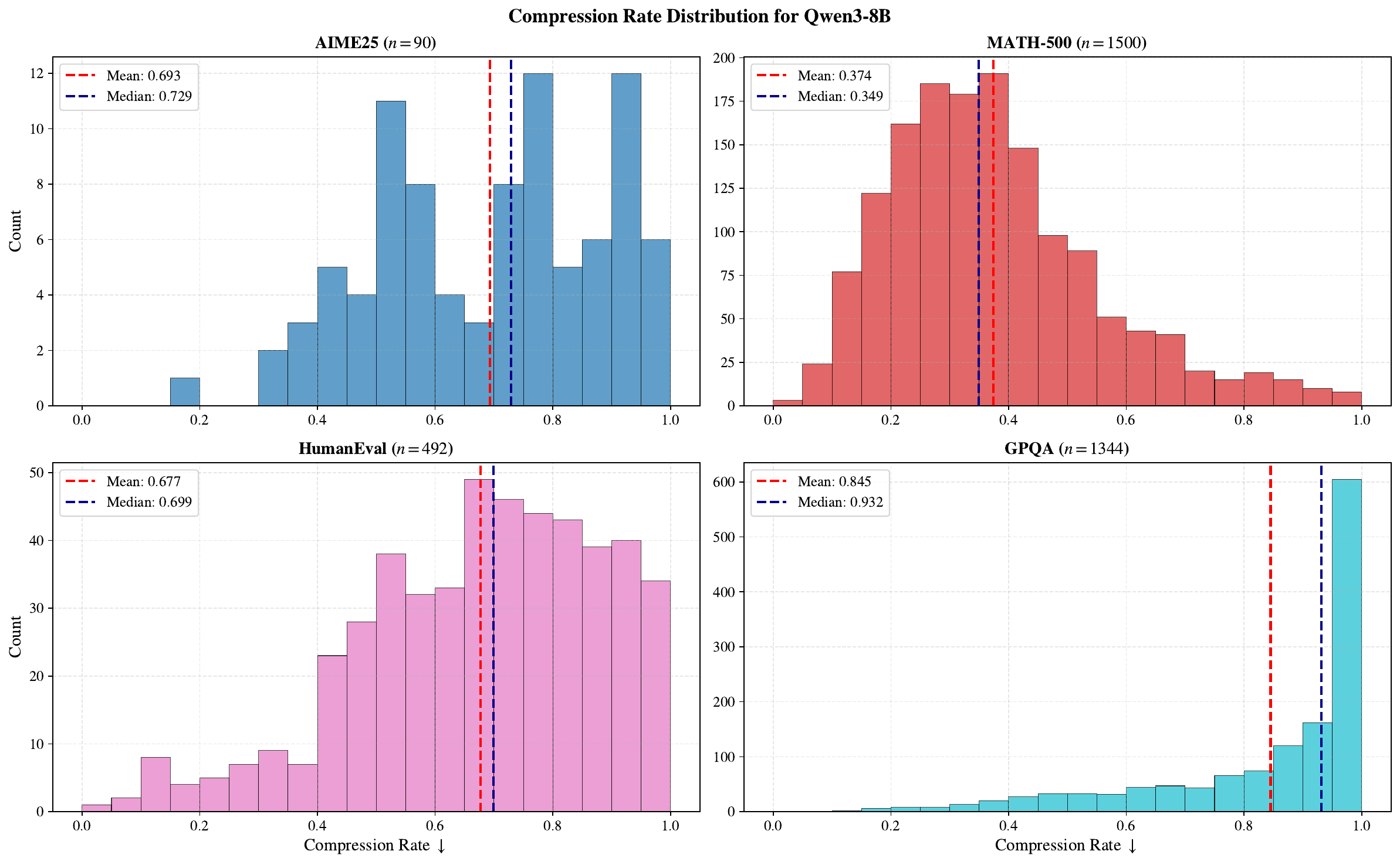}
    \caption{\textbf{Compression Ratio Histograms for \method.} Each histogram shows the frequency of an achieved compression rate when early-exiting with \method. $\downarrow$ indicates that a lower compression rate is better, as it results in more tokens saved with our method. Three \rtx[s] are sampled per data source.}
    \label{fig:histogram-comp-rate}
\end{figure*}

\textbf{Answer Prediction Histograms. } \cref{fig:histogram-train-data} gives an overview of the position of the first occurrence of \ans in the \rtx[s] for each data source we used for training. \cref{fig:histogram-comp-rate} shows the compression rate statistics of \method for each test dataset.

\begin{figure*}[!t]
    \centering
    \includegraphics[width=0.8\textwidth]{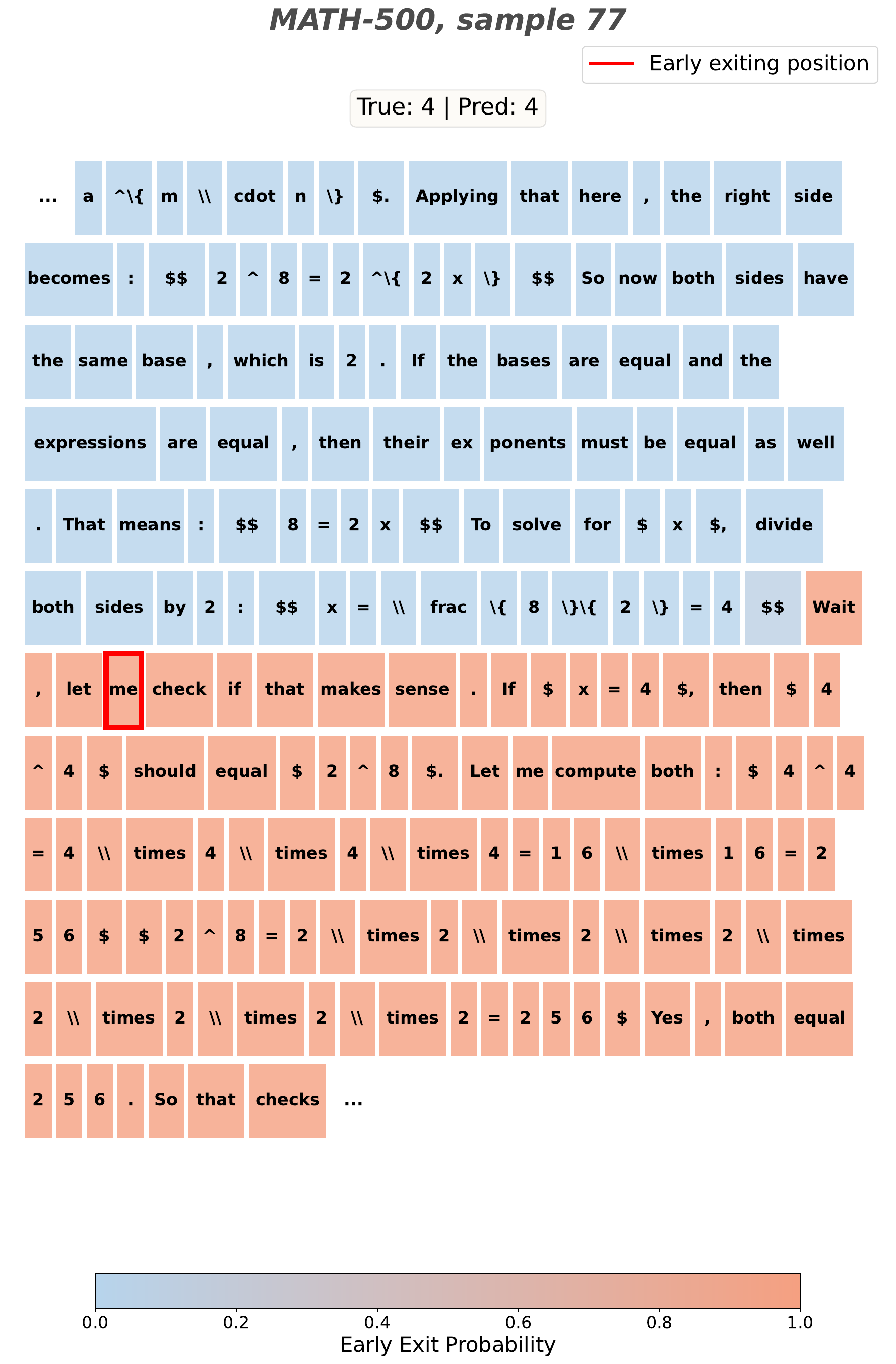}
    \caption{\textbf{Predicted Probabilities for MATH-500.} \method's predicted probabilities for early-exit on a randomly chosen sample from MATH-500. The beginning and the end are truncated for better visibility.}
    \label{fig:cot-math-77}
\end{figure*}

\begin{figure*}[!t]
    \centering
    \includegraphics[width=0.8\textwidth]{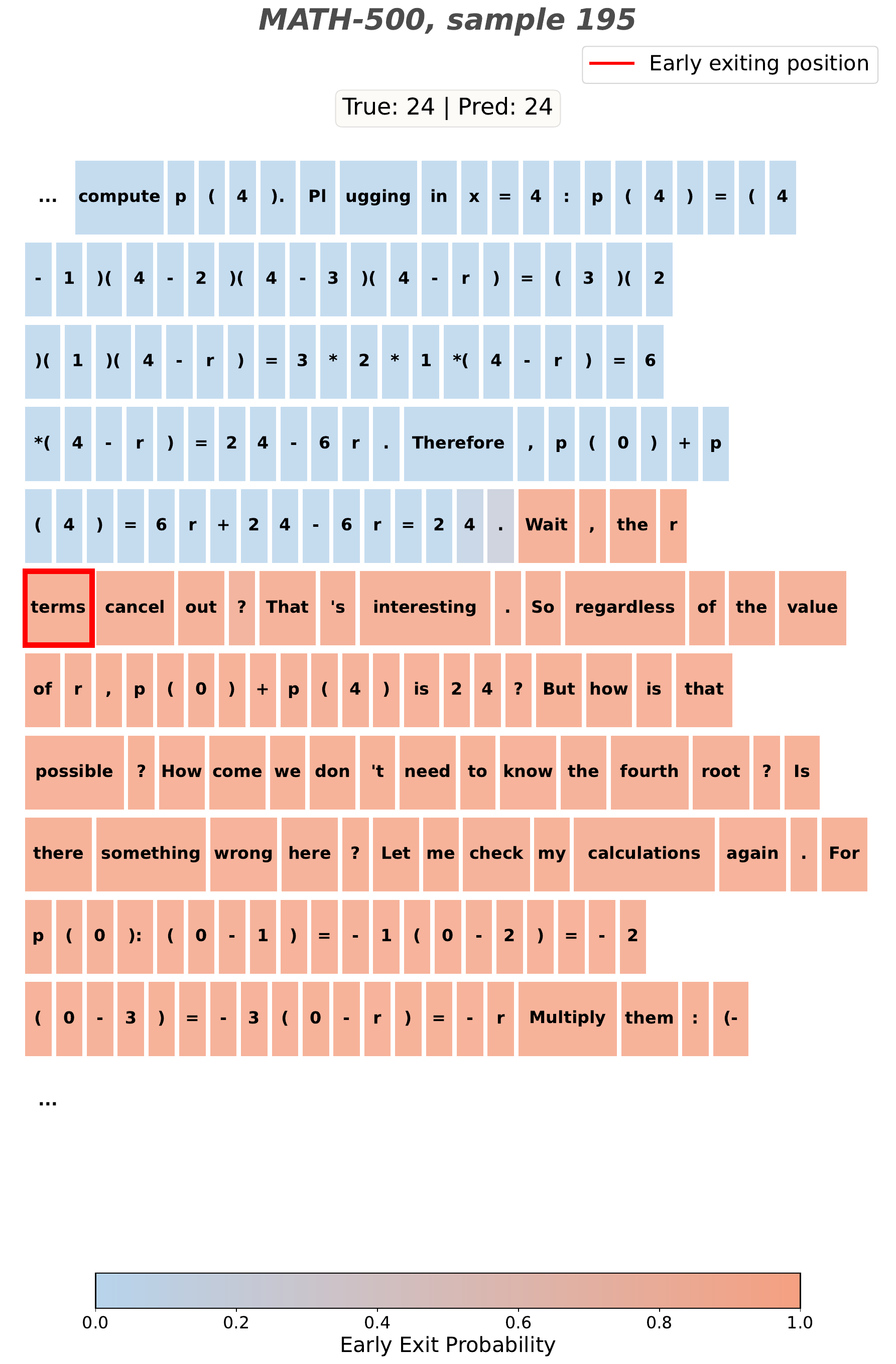}
    \caption{\textbf{Predicted Probabilities for MATH-500.} \method's predicted probabilities for early-exit on a randomly chosen sample from MATH-500. The beginning and the end are truncated for better visibility.}
    \label{fig:cot-math-195}
\end{figure*}

\begin{figure*}[!t]
    \centering
    \includegraphics[width=0.8\textwidth]{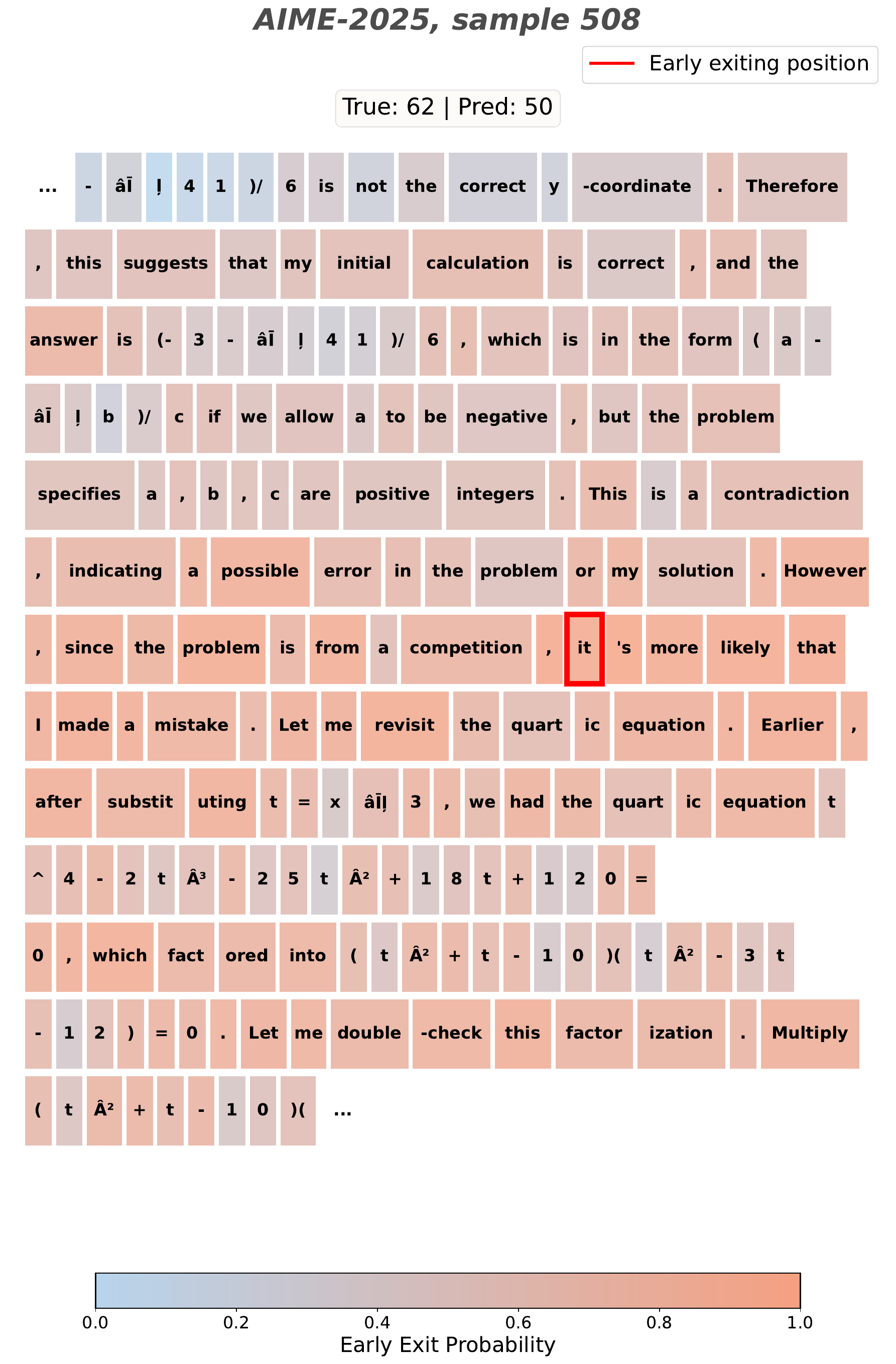}
    \caption{\textbf{Predicted Probabilities for AIME25.} \method's predicted probabilities for early-exit on a randomly chosen sample from AIME-2025. The beginning and the end are truncated for better visibility.}
    \label{fig:cot-aime-508}
\end{figure*}

\begin{figure*}[!t]
    \centering
    \includegraphics[width=0.8\textwidth]{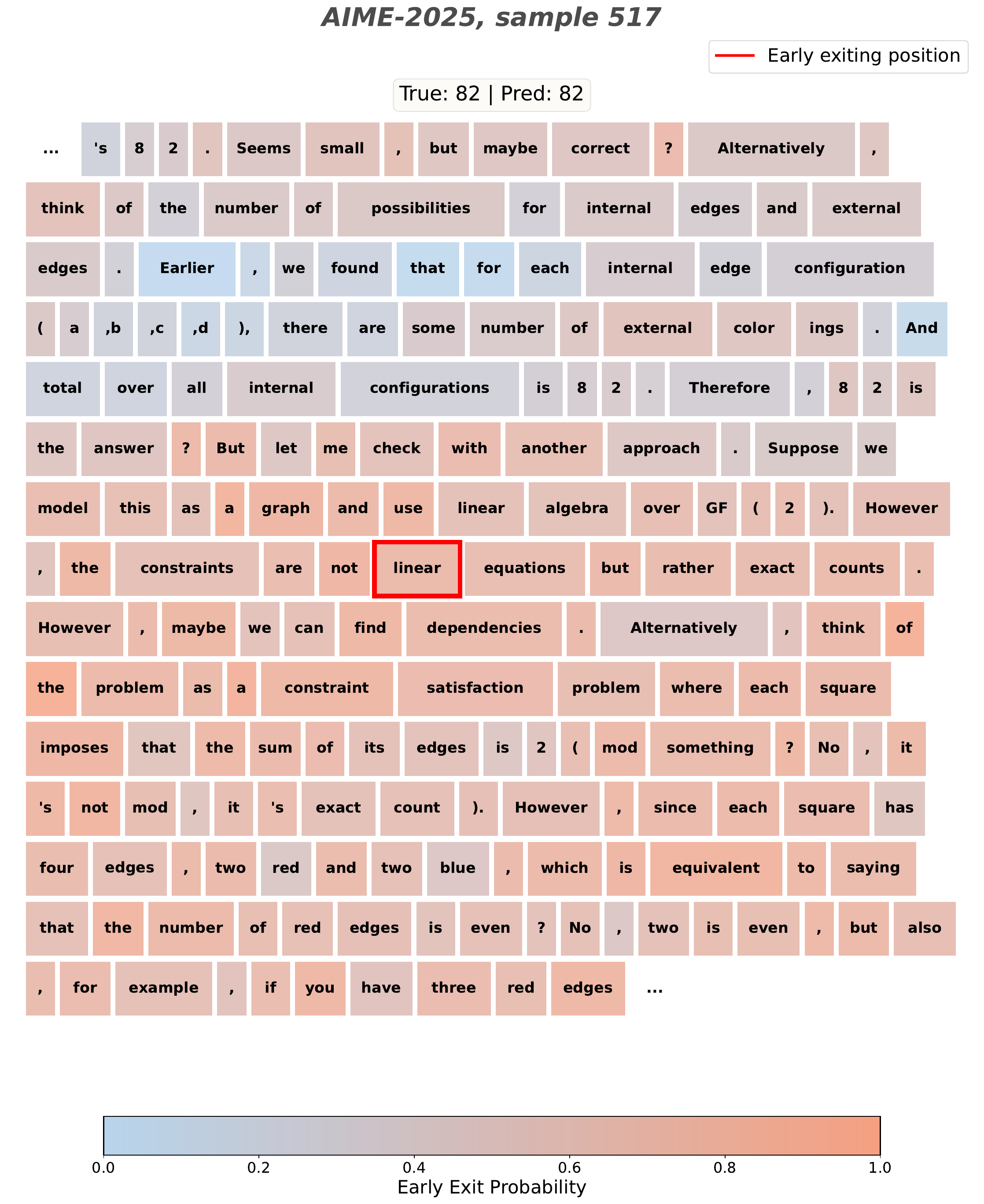}
    \caption{\textbf{Predicted Probabilities for AIME25.} \method's predicted probabilities for early-exit on a randomly chosen sample from AIME-2025. The beginning and the end are truncated for better visibility.}
    \label{fig:cot-aime-517}
\end{figure*}

\textbf{Case Study of \method.}
By manual inspection, we have seen that easy problems have a clearer transition to overthinking, which is well detected by \method. However, \method may not always detect the appropriate answer position cleanly on harder questions. \cref{fig:cot-math-195,fig:cot-math-77,fig:cot-aime-508,fig:cot-aime-517} demonstrate this behavior on Qwen3-14B \rtx[s].

\begin{figure}[!t]
    \centering
    \includegraphics[width=0.9\columnwidth]{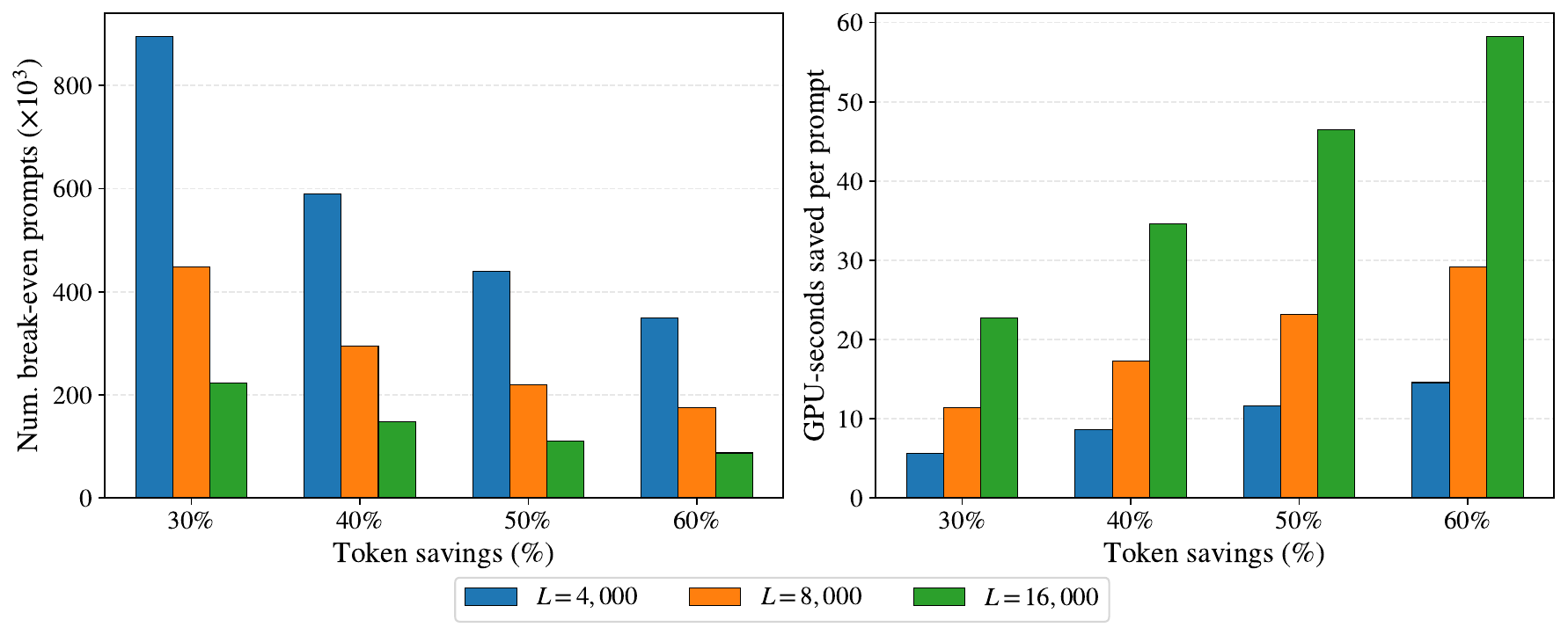}
    \caption{\textbf{Pipeline Cost-Benefit Analysis.} We report an estimate on the number of prompts to \method needed to break even with the compute cost associated with running our data curation pipeline \textbf{(left)} and on the number of GPU-seconds saved per prompt \textbf{(right)}. We use the Qwen3-8B throughput data from \cref{tab:latency} and calculate the compute cost across fixed token savings (\ie a reduction in the number of tokens by $30\%, 40\%, 50\%,$ and $ 60\%$) and fixed prompt lengths of $4,000, 8,000$ and $16,000$ tokens. At the time of writing, the number of \textit{monthly} Qwen3-8B downloads from Hugging Face is ${\sim}9.9$ million \cite{qwen3-8b-model-card}. If just a single prompt were run per download on Qwen3-8B with \method, the compute offset would be compensated for many times over.}
    \label{fig:cost-analysis}
\end{figure}

\subsection{Pipeline Cost-Benefit Analysis}
\label{app:cost-analysis}
Gathering $110,799$ \rtx[s] for training takes approximately $3$ days with a single GH200. The pipeline for obtaining the position of \ans on those \rtx[s] takes roughly one week on $8$ GH200s, totaling an estimated $1,416$ GPU-hours. \cref{fig:cost-analysis} provides a rough estimate of the number of prompts needed to offset the one-time data curation compute cost to train Terminator with Qwen3-8B (left), along with the number of seconds saved per prompt. The x-axis corresponds to the fixed token savings across all prompts, and $L$ represents the length of the CoTs. For example, for a $30\%$ savings, about ${\sim}200,000$ prompts with CoT lengths of $16,000$ are sufficient to break even with the data curation step. The number of prompts required to break even is small compared to the number of monthly downloads for Qwen3-8B on Hugging Face (${\sim}9.9$ million at the time of writing \cite{qwen3-8b-model-card}). If just a single prompt were run per download on Qwen3-8B with \method, the compute offset for the training data curation would be compensated for many times over.

\subsection{Memory Overhead}
\label{app:memory}
We profile \method's VRAM cost at inference on Qwen3-8B. \method adds a static overhead but reduces peak KV cache by exiting earlier; the net effect on memory is favorable. At fixed GPU memory utilization, \method reduces the available KV blocks by 3.2\% and the KV pool by 0.5\% (68.81 to 68.45~GiB; \cref{tab:memory-static}). Despite this, median peak KV cache per request falls 22\% on GPQA and 50\% on MATH-500 (\cref{tab:memory-peak-kv}).

Profiling used 20 random samples per dataset, a batch size of 1, and disabled prefix caching. We report a batch size of 1 because our vLLM integration is an out-of-tree prototype in which the termination decision runs outside CUDA graph capture; supporting continuous batching is a serving-systems task rather than a change to the method.

\begin{table}[t]
\centering
\caption{\textbf{Peak KV Cache per Request.} Peak KV cache per request for Qwen3-8B-based vanilla and \method models (batch size 1). Mean values are reported as mean $\pm$ 95\% CI.}
\label{tab:memory-peak-kv}
\begin{tabular}{@{}llcc@{}}
\toprule
\textbf{Dataset} & \textbf{Method} & \textbf{Mean Peak KV (MiB)} & \textbf{Median Peak KV (MiB)} \\
\midrule
\multirow{2}{*}{GPQA}     & \textit{Vanilla}    & $1424.1 \pm 306.5$ & 1396.1 \\
                          & \textit{\method}    & $1237.0 \pm 314.9$ & 1095.0 \\
\midrule
\multirow{2}{*}{MATH-500} & \textit{Vanilla}    & $508.9 \pm 174.2$  & 348.8 \\
                          & \textit{\method}    & $284.1 \pm 155.9$  & 175.8 \\
\bottomrule
\end{tabular}
\end{table}

\begin{table}[t]
\centering
\caption{\textbf{Static Memory Footprint.} Static memory footprint of Qwen3-8B-based vanilla and \method models at fixed GPU memory utilization. Head parameters are reported at BF16 serving precision (the FP32 checkpoint is 736.1~MiB on disk).}
\label{tab:memory-static}
\small
\begin{tabular}{@{}lcccc@{}}
\toprule
\textbf{Model} & \textbf{Head Params (MiB)} & \textbf{Extra Layers} & \textbf{KV Blocks} & \textbf{KV Pool (GiB)} \\
\midrule
Vanilla-Qwen3-8B    & --     & 0 & 31,317 & 68.81 \\
\method-Qwen3-8B    & 368.05 & 1 & 30,311 & 68.45 \\
\bottomrule
\end{tabular}
\end{table}

\subsection{Disentangling Hindsight-Optimal Labels and Token-Level Exit Prediction}
\label{app:hindsight_vs_token}
We analyze the relative roles of two design choices in \method: 
(1) hindsight-optimal exit labels and 
(2) token-level exit prediction. 
These two ingredients are closely coupled. Hindsight-optimal labels provide the supervision signal that makes token-level predictions meaningful, while token-level predictions are needed to fully exploit the fine-grained structure of those labels. To isolate their effects, we consider all four combinations of hindsight-optimal versus consistency-based labels and token-level versus chunk-level prediction.

We first describe the two alternatives considered in this analysis.

\textit{Consistency-based Labels.}
Consistency-based early-exit methods inject \texttt{</think>} at intermediate points in the \rtx and force the model to produce a final answer. One common labeling strategy is to assign a positive exit label to the first intermediate point whose forced answer matches the model's final answer $\hat{a}$. In contrast, our hindsight-optimal labeling assigns a positive label at the first point where $\hat{a}$ itself appears in the \rtx. Thus, rather than repeatedly probing whether the model can already reproduce the final answer, hindsight-optimal labeling directly identifies when the final answer has emerged in the reasoning trajectory.

\textit{Chunk-level Prediction.}
Prior early-exit approaches typically make exit decisions only at predefined checkpoints during generation. These checkpoints may occur at fixed intervals, after structural delimiters such as \texttt{\textbackslash n\textbackslash n}, or after heuristic ``thinking tokens'' such as \texttt{wait}, \texttt{alternatively}, or \texttt{therefore}. In contrast, \method predicts whether to exit at every generated token.

The four possible combinations are summarized in \cref{tab:label_granularity_summary}.
\begin{table}[h]
\centering
\caption{\textbf{Combinations of exit supervision and prediction granularity.} 
\method combines hindsight-optimal labels with token-level prediction. 
The remaining combinations either correspond to prior work, underperform our method, or are computationally infeasible.}
\label{tab:label_granularity_summary}
\begin{tabular}{lcc}
\toprule
 & \textbf{Token-level prediction} & \textbf{Chunk-level prediction} \\
\midrule
\textbf{Hindsight-optimal labels} 
& \method (proposed)
& Underperforms \method \\
\textbf{Consistency-based labels} 
& Computationally infeasible 
& Prior work (underperforms) \\
\bottomrule
\end{tabular}
\end{table}

\textit{Hindsight-optimal Labels with Token-level Prediction.}
This is our proposed method, \method. It uses hindsight-optimal labels to identify the earliest point at which the final answer appears in the \rtx, and trains a token-level predictor to decide whether generation can safely terminate at each position.

\textit{Hindsight-optimal Labels with Chunk-level Prediction.}
To test whether hindsight-optimal labels alone are sufficient, we also train a variant that only makes exit predictions at chunk boundaries, following the prediction granularity used in prior early-exit work. Specifically, we use the \texttt{\textbackslash n\textbackslash n} delimiter to define chunks and assign the target exit to the end of the chunk containing the hindsight-optimal exit position. At inference time, predictions are made only at chunk boundaries.

\cref{tab:chunks} shows that this variant substantially underperforms token-level \method. Although the chunk-level variant exits much less aggressively, achieving substantially lower compression rates, it also suffers large accuracy drops across evaluation datasets. For example, the chunk-level variant achieves an overall accuracy of $61.9\%$ and compression rate of $24.1\%$, compared to $77.9\%$ accuracy and $67.8\%$ compression for \method. The gap is particularly large on AIME25 and GPQA, where coarse exit checkpoints appear unable to reliably capture the precise point at which the model has completed the necessary reasoning. These results indicate that token-level granularity is necessary to fully leverage hindsight-optimal supervision.

\begin{table}[t]
\centering
\caption{\textbf{Effect of chunk-level exit prediction.} 
We train a chunk-level \method variant using the \texttt{\textbackslash n\textbackslash n} delimiter so that, for each \rtx, the positive exit label is assigned to the end of the chunk containing the hindsight-optimal exit position. At inference time, exit predictions are made only at chunk boundaries. All other hyperparameters, including the exit threshold, are kept the same as in \method.}
\label{tab:chunks}
\resizebox{1.0\textwidth}{!}{%
\begin{tabular}{@{}l|cc|cc|cc|cc|cc@{}}
\toprule
 & \multicolumn{4}{c}{\textbf{Math}} & \multicolumn{2}{c}{\textbf{Coding}} & \multicolumn{2}{c}{\textbf{Science}} & \multicolumn{2}{c}{} \\
\cmidrule(l){2-5} \cmidrule(l){6-7} \cmidrule(l){8-9} 
\textbf{Method}
  & \multicolumn{2}{c}{\textbf{MATH-500}}
  & \multicolumn{2}{c}{\textbf{AIME25}}
  & \multicolumn{2}{c}{\textbf{HumanEval}}
  & \multicolumn{2}{c}{\textbf{GPQA}}
  & \multicolumn{2}{c}{\textbf{Overall}} \\
 & {Acc$\uparrow$} & {CR$\downarrow$}
 & {Acc$\uparrow$} & {CR$\downarrow$}
 & {Acc$\uparrow$} & {CR$\downarrow$}
 & {Acc$\uparrow$} & {CR$\downarrow$}
 & {Acc$\uparrow$} & {CR$\downarrow$} \\
\midrule

\method (Proposed)
  & \textbf{90.7\%} & 45.1\%
  & \textbf{69.4\%} & 70.7\%
  & \textbf{95.7\%} & 69.9\%
  & \textbf{55.7\%} & 85.7\%
  & \textbf{77.9\%} & 67.8\% \\
\midrule

Chunk Variant
  & 80.8\% & \textbf{18.1\%}
  & 34.9\% & \textbf{43.0\%}
  & 86.6\% & \textbf{18.4\%}
  & 45.4\% & \textbf{16.9\%}
  & 61.9\% & \textbf{24.1\%} \\
\bottomrule
\end{tabular}%
}
\end{table}

\textit{Consistency-based Labels with Token-level Prediction.}
A token-level consistency-based method would require injecting \texttt{</think>} after every generated token and forcing the model to produce an intermediate final answer at each position. This is computationally prohibitive. For example, assuming a conservative overhead of 1 second per forced-answer generation, labeling a single \rtx with 8,000 tokens would require roughly 2.22 hours of compute. Using the same compute budget as for our training data ($1,416$ GPU hours, see \cref{app:cost-analysis}), this would label only a small fraction (about $700$) of the available \rtx[s] ($110,799$ in total), yielding too little data for effective training. Thus, while token-level consistency labels are conceptually possible, they are not practical at scale.

\textit{Consistency-based labels with chunk-level prediction.}
This setting corresponds to the dominant design used in prior early-exit work, where the model is probed at coarse checkpoints and exits when an intermediate forced answer is consistent with the final answer. Our main results show that \method outperforms these methods. Moreover, the hindsight-optimal chunk-level variant in \cref{tab:chunks} performs worse than token-level \method, despite using our stronger labeling signal. This demonstrates that hindsight-optimal labels alone are not sufficient: they define a better target, but token-level prediction is needed to realize their benefit.

Overall, these results suggest that hindsight-optimal labels set the achievable performance ceiling by identifying meaningful early-exit positions, while token-level prediction provides the granularity needed to reach that ceiling.

{
\sisetup{
  mode = text,
  group-separator = {,},
  group-minimum-digits = 4
}

\newcommand{\pct}[1]{#1\%}                          
\newcommand{\tkn}[1]{\text{\num{#1}}}               
\newcommand{\red}[1]{\textcolor{red}{#1}}
\newcommand{\lbar}[1]{\multicolumn{1}{|l}{#1}}

\newcommand{\trip}[3]{#1 & #2 & #3}        
\newcommand{\overall}[2]{#1 & #2}          

\setlength{\tabcolsep}{2.5pt}
\begin{sidewaystable}
\centering
\caption{\textbf{Performance of \method and Baselines.} $\uparrow$ Indicates that higher values are better, while $\downarrow$ indicates that lower values are better. CR is the compression rate, reported here as the mean per-sample compression rate. Tok is the mean number of tokens per sample. \textbf{Bold} and \underline{Underlined} values highlight the best and second-best performing early exit methods, respectively. \method demonstrates superior accuracy-efficiency trade-offs (best or second-best performance across 28 out of 32 metrics). \cref{fig:pareto-frontier} in \cref{sec:experiments} shows the results of this table on the Pareto frontier.}
\begin{tabular}{@{}lcccccccccccccc@{}} 
\toprule
 & \multicolumn{6}{c}{\textbf{Math}} & \multicolumn{3}{c}{\textbf{Coding}} & \multicolumn{3}{c}{\textbf{Science}}  \\ 
 \cmidrule(l){2-7} \cmidrule(l){8-10} \cmidrule(l){11-13}
\textbf{Method}
 & \multicolumn{3}{c}{\textbf{MATH-500}} & \multicolumn{3}{c}{\textbf{AIME25}} & \multicolumn{3}{c}{\textbf{HumanEval}} & \multicolumn{3}{c}{\textbf{GPQA}}  & \multicolumn{2}{c}{\textbf{Overall}} \\
   & {Acc$\uparrow$} & {Tok$\downarrow$} & {CR$\downarrow$} & Acc$\uparrow$ & Tok$\downarrow$ & {CR$\downarrow$} & {Acc}$\uparrow$  & {Tok$\downarrow$} & {CR$\downarrow$} & {Acc$\uparrow$} & {Tok$\downarrow$} & {CR$\downarrow$} & {Acc$\uparrow$} & {CR}$\downarrow$  \\ 
\hline


\multicolumn{15}{l}{\cellcolor[rgb]{0.9,0.9,0.9}\textit{\textbf{Qwen3-8B}}} \\

\textit{Vanilla}
  & \trip{\lbar{\pct{91.1}}}{\tkn{5037}}{\pct{100}}
  & \trip{\lbar{\pct{74.4}}}{\tkn{14499}}{\pct{100}}
  & \trip{\lbar{\pct{94.9}}}{\tkn{3792}}{\pct{100}}
  & \trip{\lbar{\pct{58.0}}}{\tkn{8594}}{\pct{100}}
  & \overall{\lbar{\pct{79.6}}}{\pct{100}} \\

\textit{NoThinking}
  & \trip{\lbar{\pct{80.7}}}{\tkn{809}}{\pct{16.1}}
  & \trip{\lbar{\pct{22.0}}}{\tkn{2355}}{\pct{18.6}}
  & \trip{\lbar{\pct{84.6}}}{\tkn{353}}{\pct{11.8}}
  & \trip{\lbar{\pct{46.0}}}{\tkn{1204}}{\pct{15.8}}
  & \overall{\lbar{\pct{58.3}}}{\pct{15.6}} \\

\textit{DEER}
  & \trip{\lbar{\pct{79.9}}}{\tkn{2602}}{\pct{52.0}}
  & \trip{\lbar{\pct{21.4}}}{\tkn{10349}}{\underline{\pct{67.8}}}
  & \trip{\lbar{\pct{93.7}}}{\tkn{3275}}{\pct{83.6}}
  & \trip{\lbar{\pct{50.3}}}{\tkn{8553}}{\pct{99.6}}
  & \overall{\lbar{\pct{61.3}}}{\pct{75.8}} \\

\textit{Thought-Calib}
  & \trip{\lbar{\underline{\pct{90.1}}}}{\tkn{4372}}{\pct{93.9}}
  & \trip{\lbar{\underline{\pct{65.8}}}}{\tkn{11014}}{\pct{81.5}}
  & \trip{\lbar{\pct{93.9}}}{\tkn{3267}}{\pct{92.9}}
  & \trip{\lbar{\textbf{\pct{56.6}}}}{\tkn{6240}}{\underline{\pct{78.9}}}
  & \overall{\lbar{\underline{\pct{76.6}}}}{\pct{86.8}} \\

\textit{Dynasor}
  & \trip{\lbar{\pct{78.3}}}{\tkn{1850}}{\textbf{\pct{41.0}}}
  & \trip{\lbar{\pct{48.0}}}{\tkn{7479}}{\textbf{\pct{48.8}}}
  & \trip{\lbar{\underline{\pct{94.5}}}}{\tkn{2883}}{\underline{\pct{78.4}}}
  & \trip{\lbar{\pct{41.7}}}{\tkn{2455}}{\textbf{\pct{28.4}}}
  & \overall{\lbar{\pct{65.6}}}{\textbf{\pct{49.2}}} \\

\rowcolor[rgb]{0.87,0.94,1}
\textit{\method}
  & \trip{\lbar{\pct{\textbf{90.7}}}}{\tkn{2425}}{\underline{\pct{45.1}}}
  & \trip{\lbar{\pct{\textbf{69.4}}}}{\tkn{10970}}{\pct{70.7}}
  & \trip{\lbar{\textbf{\pct{95.7}}}}{\tkn{2716}}{\textbf{\pct{69.9}}}
  & \trip{\lbar{\underline{\pct{55.7}}}}{\tkn{7543}}{\pct{85.7}}
  & \overall{\lbar{\textbf{\pct{77.9}}}}{\underline{\pct{67.8}}} \\

\hline


\multicolumn{15}{l}{\cellcolor[rgb]{0.9,0.9,0.9}\textit{\textbf{Qwen3-14B}}} \\

\textit{Vanilla}
  & \trip{\lbar{\pct{92.0}}}{\tkn{4598}}{\pct{100}}
  & \trip{\lbar{\pct{79.9}}}{\tkn{14255}}{\pct{100}}
  & \trip{\lbar{\pct{96.9}}}{\tkn{3296}}{\pct{100}}
  & \trip{\lbar{\pct{60.2}}}{\tkn{7628}}{\pct{100}}
  & \overall{\lbar{\pct{82.3}}}{\pct{100}} \\

\textit{NoThinking}
  & \trip{\lbar{\pct{84.1}}}{\tkn{786}}{\pct{17.5}}
  & \trip{\lbar{\pct{26.3}}}{\tkn{2472}}{\pct{19.9}}
  & \trip{\lbar{\pct{83.7}}}{\tkn{317}}{\pct{12.2}}
  & \trip{\lbar{\pct{49.8}}}{\tkn{1265}}{\pct{18.8}}
  & \overall{\lbar{\pct{61.0}}}{\pct{17.1}} \\

\textit{DEER}
  & \trip{\lbar{\pct{80.9}}}{\tkn{2501}}{\pct{56.2}}
  & \trip{\lbar{\pct{27.6}}}{\tkn{10497}}{\underline{\pct{71.0}}}
  & \trip{\lbar{\underline{\pct{96.9}}}}{\tkn{2961}}{\pct{87.3}}
  & \trip{\lbar{\pct{52.0}}}{\tkn{7451}}{\pct{97.4}}
  & \overall{\lbar{\pct{64.5}}}{\pct{78.0}} \\

\textit{Thought-Calib}
  & \trip{\lbar{\underline{\pct{89.8}}}}{\tkn{3778}}{\pct{92.0}}
  & \trip{\lbar{\underline{\pct{63.3}}}}{\tkn{9429}}{\pct{71.3}}
  & \trip{\lbar{\pct{94.3}}}{\tkn{2582}}{\pct{87.1}}
  & \trip{\lbar{\underline{\pct{57.3}}}}{\tkn{5757}}{\underline{\pct{81.9}}}
  & \overall{\lbar{\underline{\pct{76.2}}}}{\pct{83.1}} \\

\textit{Dynasor}
  & \trip{\lbar{\pct{79.6}}}{\tkn{1702}}{\textbf{\pct{42.4}}}
  & \trip{\lbar{\pct{61.8}}}{\tkn{7937}}{\textbf{\pct{52.8}}}
  & \trip{\lbar{\pct{96.5}}}{\tkn{2611}}{\underline{\pct{82.2}}}
  & \trip{\lbar{\pct{45.7}}}{\tkn{2101}}{\textbf{\pct{29.1}}}
  & \overall{\lbar{\pct{70.9}}}{\textbf{\pct{51.6}}} \\

\rowcolor[rgb]{0.87,0.94,1}
\textit{\method}
  & \trip{\lbar{\textbf{\pct{90.7}}}}{\tkn{2261}}{\underline{\pct{46.8}}}
  & \trip{\lbar{\textbf{\pct{74.2}}}}{\tkn{10787}}{\underline{\pct{71.0}}}
  & \trip{\lbar{\textbf{\pct{97.1}}}}{\tkn{2358}}{\textbf{\pct{70.9}}}
  & \trip{\lbar{\textbf{\pct{59.6}}}}{\tkn{6798}}{\pct{87.1}}
  & \overall{\lbar{\textbf{\pct{80.4}}}}{\underline{\pct{68.9}}} \\

\hline


\multicolumn{15}{l}{\cellcolor[rgb]{0.9,0.9,0.9}\textit{\textbf{Ministral-3-8B-Reasoning-2512}}} \\

\textit{Vanilla}
  & \trip{\lbar{\pct{93.5}}}{\tkn{6212}}{\pct{100}}
  & \trip{\lbar{\pct{92.6}}}{\tkn{22124}}{\pct{100}}
  & \trip{\lbar{\pct{97.1}}}{\tkn{4367}}{\pct{100}}
  & \trip{\lbar{\pct{63.4}}}{\tkn{11765}}{\pct{100}}
  & \overall{\lbar{\pct{86.6}}}{\pct{100}} \\

\textit{NoThinking}
  & \trip{\lbar{\pct{83.2}}}{\tkn{1908}}{\pct{28.1}}
  & \trip{\lbar{\pct{43.6}}}{\tkn{7711}}{\pct{36.5}}
  & \trip{\lbar{\pct{87.6}}}{\tkn{727}}{\pct{16.4}}
  & \trip{\lbar{\pct{42.4}}}{\tkn{2106}}{\pct{16.0}}
  & \overall{\lbar{\pct{64.2}}}{\pct{24.3}} \\

\textit{DEER}
  & \trip{\lbar{\pct{71.0}}}{\tkn{3791}}{\pct{60.3}}
  & \trip{\lbar{\pct{67.1}}}{\tkn{17481}}{\pct{77.0}}
  & \trip{\lbar{\pct{80.1}}}{\tkn{3606}}{\underline{\pct{84.0}}}
  & \trip{\lbar{\textbf{\pct{61.9}}}}{\tkn{11312}}{\pct{94.1}}
  & \overall{\lbar{\pct{70.0}}}{\pct{78.9}} \\

\textit{Thought-Calib}
  & \trip{\lbar{\pct{87.7}}}{\tkn{5695}}{\pct{87.8}}
  & \trip{\lbar{\underline{\pct{83.7}}}}{\tkn{20358}}{\pct{91.2}}
  & \trip{\lbar{\pct{57.9}}}{\tkn{3536}}{\pct{87.2}}
  & \trip{\lbar{\pct{50.6}}}{\tkn{7406}}{\textbf{\pct{71.8}}}
  & \overall{\lbar{\pct{70.0}}}{\pct{84.5}} \\

\textit{Dynasor}
  & \trip{\lbar{\underline{\pct{88.1}}}}{\tkn{2967}}{\underline{\pct{56.8}}}
  & \trip{\lbar{\textbf{\pct{87.6}}}}{\tkn{15407}}{\textbf{\pct{66.3}}}
  & \trip{\lbar{\textbf{\pct{96.9}}}}{\tkn{3931}}{\pct{88.6}}
  & \trip{\lbar{\pct{57.7}}}{\tkn{9766}}{\pct{83.7}}
  & \overall{\lbar{\textbf{\pct{82.6}}}}{\underline{\pct{73.9}}} \\

\rowcolor[rgb]{0.87,0.94,1}
\textit{\method}
  & \trip{\lbar{\textbf{\pct{89.1}}}}{\tkn{2863}}{\textbf{\pct{47.8}}}
  & \trip{\lbar{\pct{79.1}}}{\tkn{15748}}{\underline{\pct{67.8}}}
  & \trip{\lbar{\underline{\pct{96.5}}}}{\tkn{2960}}{\textbf{\pct{66.6}}}
  & \trip{\lbar{\underline{\pct{58.2}}}}{\tkn{9588}}{\underline{\pct{77.4}}}
  & \overall{\lbar{\underline{\pct{80.7}}}}{\textbf{\pct{64.9}}} \\

\hline


\multicolumn{15}{l}{\cellcolor[rgb]{0.9,0.9,0.9}\textit{\textbf{Ministral-3-14B-Reasoning-2512}}} \\

\textit{Vanilla}
  & \trip{\lbar{\pct{93.0}}}{\tkn{6385}}{\pct{100}}
  & \trip{\lbar{\pct{88.1}}}{\tkn{23694}}{\pct{100}}
  & \trip{\lbar{\pct{97.5}}}{\tkn{3918}}{\pct{100}}
  & \trip{\lbar{\pct{62.9}}}{\tkn{9539}}{\pct{100}}
  & \overall{\lbar{\pct{85.4}}}{\pct{100}} \\

\textit{NoThinking}
  & \trip{\lbar{\pct{79.1}}}{\tkn{535}}{\pct{11.7}}
  & \trip{\lbar{\pct{20.5}}}{\tkn{2413}}{\pct{13.8}}
  & \trip{\lbar{\pct{88.5}}}{\tkn{528}}{\pct{14.1}}
  & \trip{\lbar{\pct{42.8}}}{\tkn{570}}{\pct{6.6}}
  & \overall{\lbar{\pct{57.75}}}{\pct{11.5}} \\

\textit{DEER}
  & \trip{\lbar{\pct{69.8}}}{\tkn{4279}}{\pct{74.6}}
  & \trip{\lbar{\pct{55.9}}}{\tkn{20049}}{\pct{84.9}}
  & \trip{\lbar{\pct{20.5}}}{\tkn{1684}}{\textbf{\pct{46.5}}}
  & \trip{\lbar{\underline{\pct{58.1}}}}{\tkn{9185}}{\pct{95.0}}
  & \overall{\lbar{\pct{51.1}}}{\pct{75.3}} \\

\textit{Thought-Calib}
  & \trip{\lbar{\underline{\pct{87.3}}}}{\tkn{5860}}{\pct{95.9}}
  & \trip{\lbar{\pct{59.4}}}{\tkn{17763}}{\pct{79.8}}
  & \trip{\lbar{\pct{45.3}}}{\tkn{3465}}{\pct{96.3}}
  & \trip{\lbar{\pct{56.2}}}{\tkn{6028}}{\textbf{\pct{73.8}}}
  & \overall{\lbar{\pct{62.1}}}{\pct{86.5}} \\

\textit{Dynasor}
  & \trip{\lbar{\pct{86.3}}}{\tkn{3240}}{\underline{\pct{55.5}}}
  & \trip{\lbar{\underline{\pct{83.2}}}}{\tkn{17920}}{\underline{\pct{70.6}}}
  & \trip{\lbar{\underline{\pct{94.7}}}}{\tkn{3538}}{\pct{88.9}}
  & \trip{\lbar{\pct{55.9}}}{\tkn{7917}}{\pct{85.2}}
  & \overall{\lbar{\underline{\pct{80.0}}}}{\underline{\pct{75.1}}} \\

\rowcolor[rgb]{0.87,0.94,1}
\textit{\method}
  & \trip{\lbar{\textbf{\pct{90.2}}}}{\tkn{2946}}{\textbf{\pct{43.9}}}
  & \trip{\lbar{\textbf{\pct{84.2}}}}{\tkn{15518}}{\textbf{\pct{63.5}}}
  & \trip{\lbar{\textbf{\pct{97.8}}}}{\tkn{2903}}{\underline{\pct{71.0}}}
  & \trip{\lbar{\textbf{\pct{61.2}}}}{\tkn{7727}}{\underline{\pct{76.5}}}
  & \overall{\lbar{\textbf{{\pct{83.4}}}}}{\textbf{\pct{63.7}}} \\

\bottomrule
\end{tabular}
\label{tab:performance}
\end{sidewaystable}
}




\clearpage
\section*{NeurIPS Paper Checklist}

\begin{enumerate}

\item {\bf Claims}
    \item[] Question: Do the main claims made in the abstract and introduction accurately reflect the paper's contributions and scope?
    \item[] Answer: \answerYes{} 
    \item[] Justification: We provide a boxed area which outlines the main contributions of our paper in the introduction \cref{sec:intro}.

\item {\bf Limitations}
    \item[] Question: Does the paper discuss the limitations of the work performed by the authors?
    \item[] Answer: \answerYes{} 
    \item[] Justification: Please see \cref{app:limitations}.

\item {\bf Theory assumptions and proofs}
    \item[] Question: For each theoretical result, does the paper provide the full set of assumptions and a complete (and correct) proof?
    \item[] Answer: \answerNA{} 
    \item[] Justification: Our paper does not include theoretical results.

    \item {\bf Experimental result reproducibility}
    \item[] Question: Does the paper fully disclose all the information needed to reproduce the main experimental results of the paper to the extent that it affects the main claims and/or conclusions of the paper (regardless of whether the code and data are provided or not)?
    \item[] Answer: \answerYes{} 
    \item[] Justification: Our approach is fully outlined in \cref{sec:methods}. Additional details on training data, model training, and model inference is provided in \cref{app:dataset,app:training-details,app:inference-details}.

\item {\bf Open access to data and code}
    \item[] Question: Does the paper provide open access to the data and code, with sufficient instructions to faithfully reproduce the main experimental results, as described in supplemental material?
    \item[] Answer: \answerYes{} 
    \item[] Justification: We include our code and a sample of our training data as supplementary material, and we will fully release our code and data.

\item {\bf Experimental setting/details}
    \item[] Question: Does the paper specify all the training and test details (e.g., data splits, hyperparameters, how they were chosen, type of optimizer) necessary to understand the results?
    \item[] Answer: \answerYes{} 
    \item[] Justification: See \cref{app:dataset,app:training-details,app:inference-details}.

\item {\bf Experiment statistical significance}
    \item[] Question: Does the paper report error bars suitably and correctly defined or other appropriate information about the statistical significance of the experiments?
    \item[] Answer: \answerYes{} 
    \item[] Justification: Statistical significance information is explicitly reported in \cref{fig:aligned-timeseries-all,fig:aligned-timeseries-aime,fig:aligned-timeseries-math,fig:aligned-timeseries-opencoder,fig:aligned-timeseries-openscience,fig:timeseries-comparison,fig:token-scatter,fig:token-scatter-all,fig:token-scatter-aime,fig:token-scatter-math,fig:token-scatter-opencoder,fig:token-scatter-openscience,fig:scatter-overlay,fig:timeseries-comparison,fig:predictions-aime,fig:predictions-math,fig:predictions-gpqa,fig:predictions-humaneval,tab:latency}.

\item {\bf Experiments compute resources}
    \item[] Question: For each experiment, does the paper provide sufficient information on the computer resources (type of compute workers, memory, time of execution) needed to reproduce the experiments?
    \item[] Answer: \answerYes{} 
    \item[] Justification: The details are given in \cref{app:cost-analysis}.
    
\item {\bf Code of ethics}
    \item[] Question: Does the research conducted in the paper conform, in every respect, with the NeurIPS Code of Ethics \url{https://neurips.cc/public/EthicsGuidelines}?
    \item[] Answer: \answerYes{} 
    \item[] Justification: Our research conforms to the NeurIPS Code of Ethics.

\item {\bf Broader impacts}
    \item[] Question: Does the paper discuss both potential positive societal impacts and negative societal impacts of the work performed?
    \item[] Answer: \answerYes{} 
    \item[] Justification: Please see \cref{app:broader-impacts}.
    
\item {\bf Safeguards}
    \item[] Question: Does the paper describe safeguards that have been put in place for responsible release of data or models that have a high risk for misuse (e.g., pre-trained language models, image generators, or scraped datasets)?
    \item[] Answer: \answerNA{} 
    \item[] Justification: Our work builds on top of existing pre-trained language models, which have safeguards in place already. Our work does not further increase the risk of misuse.

\item {\bf Licenses for existing assets}
    \item[] Question: Are the creators or original owners of assets (e.g., code, data, models), used in the paper, properly credited and are the license and terms of use explicitly mentioned and properly respected?
    \item[] Answer: \answerYes{} 
    \item[] Justification: We properly cite all owners of the assets from which we derive our work.

\item {\bf New assets}
    \item[] Question: Are new assets introduced in the paper well documented and is the documentation provided alongside the assets?
    \item[] Answer: \answerYes{} 
    \item[] Justification: Our original assets are well-documented throughout the paper and in our code.

\item {\bf Crowdsourcing and research with human subjects}
    \item[] Question: For crowdsourcing experiments and research with human subjects, does the paper include the full text of instructions given to participants and screenshots, if applicable, as well as details about compensation (if any)? 
    \item[] Answer: \answerNA{} 
    \item[] Justification: We do not use crowdsourcing nor involve human subjects.

\item {\bf Institutional review board (IRB) approvals or equivalent for research with human subjects}
    \item[] Question: Does the paper describe potential risks incurred by study participants, whether such risks were disclosed to the subjects, and whether Institutional Review Board (IRB) approvals (or an equivalent approval/review based on the requirements of your country or institution) were obtained?
    \item[] Answer: \answerNA{} 
\item[] Justification: We do not use crowdsourcing nor involve human subjects.

\item {\bf Declaration of LLM usage}
    \item[] Question: Does the paper describe the usage of LLMs if it is an important, original, or non-standard component of the core methods in this research? Note that if the LLM is used only for writing, editing, or formatting purposes and does \emph{not} impact the core methodology, scientific rigor, or originality of the research, declaration is not required.
    \item[] Answer: \answerYes{} 
    \item[] Justification: Our pipeline for training dataset curation relies heavily on LRM usage, which is fully detailed in \cref{sec:methods}.

\end{enumerate}

\end{document}